%% file: main.tex
\documentclass[lettersize,journal,compsoc]{IEEEtran}
\usepackage{amsmath,amsfonts}
\usepackage{algorithmic}
\usepackage{algorithm}
\usepackage{array}
\usepackage[caption=false,font=normalsize,labelfont=sf,textfont=sf]{subfig}
\usepackage{textcomp}
\usepackage{stfloats}
\usepackage{url}
\usepackage{verbatim}
\usepackage{graphicx}
\usepackage{wrapfig}
\usepackage{amsmath}
\usepackage{multirow}
\usepackage{color}
\usepackage{xcolor}
 \usepackage{subfig}
\usepackage{booktabs}       
\usepackage{enumitem}
\usepackage[figuresright]{rotating}
\usepackage[normalem]{ulem}
\useunder{\uline}{\ul}{}
\PassOptionsToPackage{numbers, square}{natbib}
\usepackage{cite}
\begin{document}


\title{Kernelized Sparse Fine-Tuning with Bi-level Parameter Competition for Vision Models}

\author{Shufan~Shen,
Junshu~Sun,
Shuhui~Wang,~\IEEEmembership{Member,~IEEE},  and~Qingming~Huang,~\IEEEmembership{Fellow,~IEEE}
\thanks{Corresponding author: Shuhui Wang.}
\thanks{S. Shen, J. Sun and S. Wang are with the Key Laboratory of Intelligent Information Processing, Institute of Computing Technology, Chinese Academy of Sciences, Beijing 100190, China. \protect\\E-mail: \{shenshufan22z, sunjunshu21s, wangshuhui\}@ict.ac.cn.

\noindent S. Shen, J. Sun and Q. Huang are with the School of Computer Science and Technology, University of Chinese Academy of Sciences, Beijing 101408, China.
\protect\\E-mail: qmhuang@ucas.ac.cn.
}
}


\markboth{Journal of \LaTeX\ Class Files,~Vol.~14, No.~8, August~2021}%
{Shell \MakeLowercase{\textit{et al.}}: A Sample Article Using IEEEtran.cls for IEEE Journals}


\maketitle

\input{sec/0_abstract}

\input{sec/1_introduction}
\input{sec/2_relatedwork}
\input{sec/3_method}
\input{sec/4_experiments}
\input{sec/5_conclusion}

{
\bibliographystyle{IEEEtran}
\bibliography{main}
}

\newpage
\begin{IEEEbiography}[{\includegraphics[width=1in,height=1.25in,keepaspectratio]{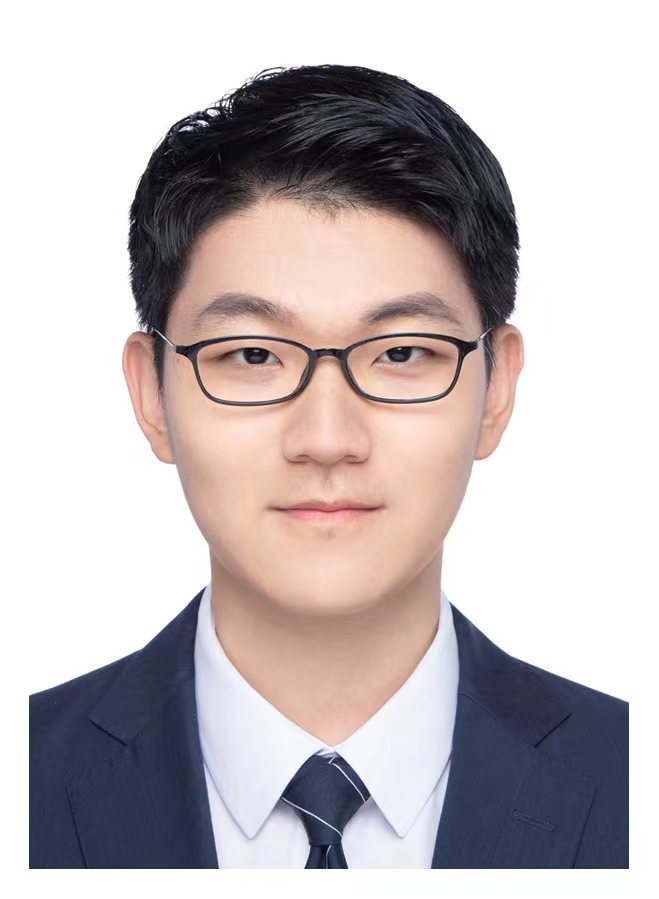}}]{Shufan Shen} 
Ph.D. student at the Institute of Computing Technology, Chinese Academy of Sciences. He received the B.S. degree in Data Science from Tongji University in 2022. His current research interests include interpretable machine learning
and parameter-efficient fine-tuning.
\end{IEEEbiography}

\begin{IEEEbiography}[{\includegraphics[width=1in,height=1.25in,keepaspectratio]{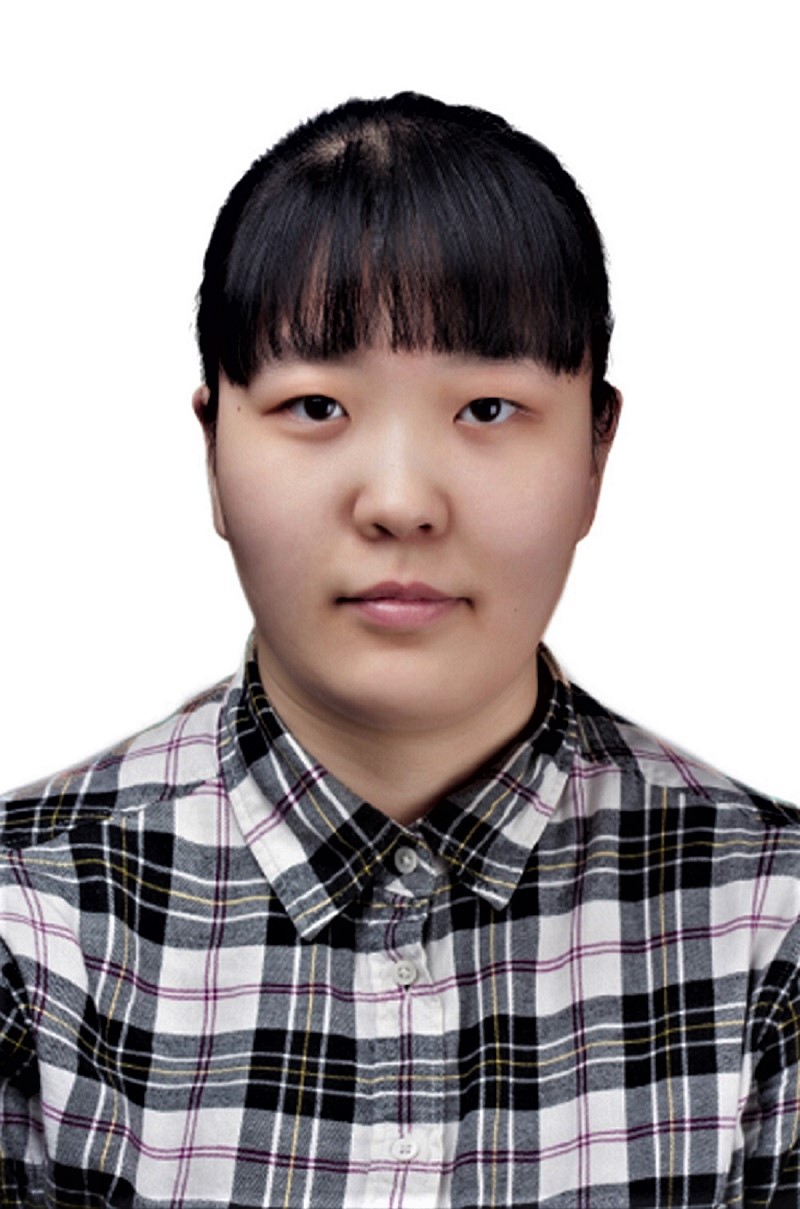}}]{Junshu Sun} 
Ph.D. student at the Institute of Computing Technology, Chinese Academy of Sciences. She received the B.S. degree in Biomedical Engineering from the University of Electronic Science and Technology in 2021. Her current research interests include graph representation learning
and geometric deep learning.
\end{IEEEbiography}

\begin{IEEEbiography}[{\includegraphics[width=1in,height=1.25in,clip,keepaspectratio]{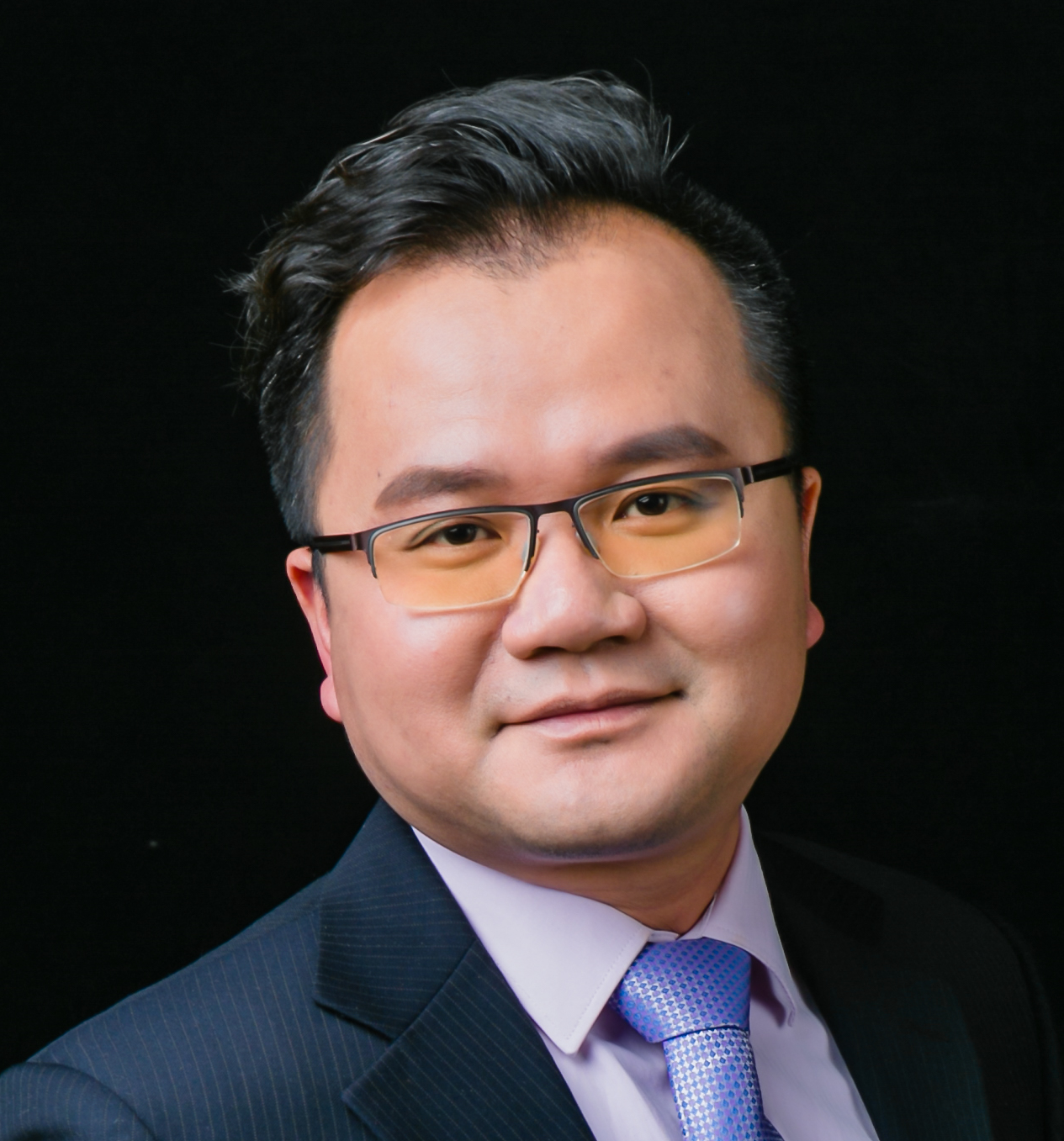}}]{Shuhui Wang}
	Full professor with the Key Laboratory of Intelligent Information Processing (CAS), Institute of Computing Technology, Chinese Academy of Sciences. He received the B.S. degree in electronics engineering from Tsinghua University, in 2006, and the Ph.D. degree from the Institute of Computing Technology, Chinese Academy of Sciences, in 2012. He is also with Pengcheng Laboratory, Shenzhen. His research interests include image/video understanding/retrieval, cross-media analysis, and visual-textual knowledge extraction.
\end{IEEEbiography}

\begin{IEEEbiography}[{\includegraphics[width=1in,height=1.25in,clip,keepaspectratio]{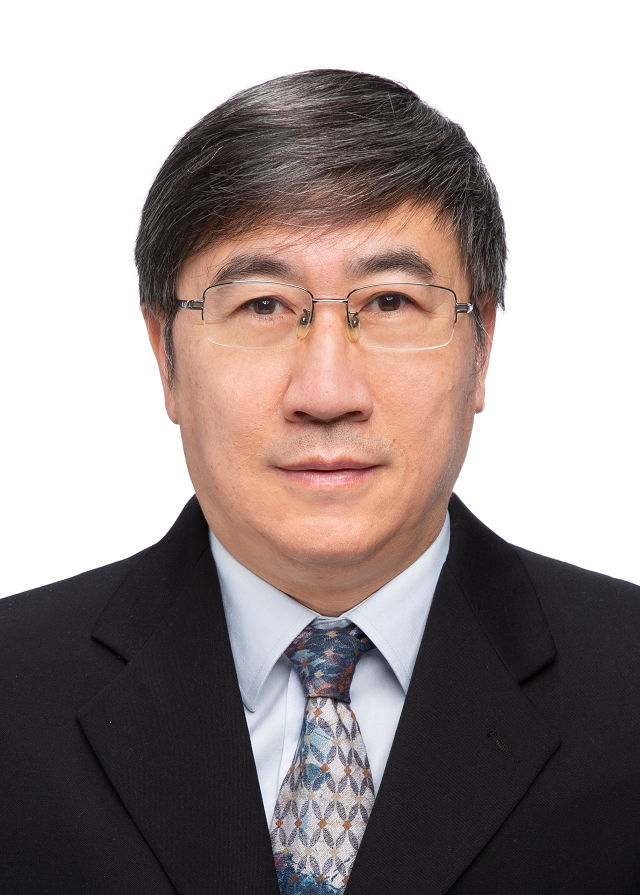}}]{Qingming Huang}
	 Chair professor with the School of Computer Science and Technology, University of Chinese Academy of Sciences. He received the B.S. degree in computer science and the Ph.D. degree in computer engineering from the Harbin Institute of Technology, China, in 1988 and 1994, respectively. He has published more than 500 academic papers in international journals, such as IEEE Transactions on Pattern Analysis and Machine Intelligence, IEEE Transactions on Image Processing, IEEE Transactions on Multimedia, IEEE Transactions on Circuits and Systems for Video Technology, and top-level international conferences, including the ACM Multimedia, ICCV, CVPR, ECCV, VLDB, and IJCAI. He was the associate editor for IEEE Transactions on Circuits and Systems for Video Technology and the associate editor for Acta Automatica Sinica. His research interests include multimedia computing, image/video processing, pattern recognition, and computer vision.
\end{IEEEbiography}

\newpage

\appendices
\input{sec/6_appendix}
 




\vfill

\end{document}

%% file: sec/0_abstract.tex
\begin{abstract}
Parameter-efficient fine-tuning~(PEFT) aims to adapt pre-trained vision models to downstream tasks.
Among PEFT paradigms, sparse tuning achieves remarkable performance by adjusting only the weights most relevant to downstream tasks, rather than densely tuning the entire weight matrix. Current sparse tuning methods follow a two-stage paradigm. First, it locates task-relevant weights by gradient information, which overlooks the parameter adjustments during fine-tuning and limits the performance. Second, it updates only the located weights by applying a sparse mask to the gradient of the weight matrix, which results in high memory usage due to the storage of all weight matrices in the optimizer. 
In this paper, we propose a one-stage method named SNELLA~(\textbf{S}parse tuning with ker\textbf{NEL}ized \textbf{L}oRA and \textbf{A}daptive bi-level sparsity allocation) to overcome the above limitations. 
For memory usage, SNELLA selectively updates the weight matrix by adding it to another sparse matrix that is merged by two low-rank learnable matrices. 
We extend the low-rank decomposition by introducing nonlinear kernel functions, thereby increasing the rank of the resulting merged matrix to prevent the interdependency among weight updates, enabling better adaptation to downstream tasks.
For locating task-relevant weights, we propose an adaptive bi-level sparsity allocation mechanism that encourages weights to compete across and inside layers based on their importance scores in an end-to-end manner under a predefined overall budget of weight updating, ensuring task-relevant weights to attain higher scores and be updated more likely.
Extensive experiments are conducted on classification, segmentation, and generation tasks using pre-trained vision models with different parameter scales, architectures, and pre-training strategies. The results show that SNELLA achieves state-of-the-art performance with low memory usage. Notably, SNELLA obtains 1.8\% (91.9\% v.s. 90.1\%) higher Top-1 accuracy on the FGVC benchmark compared to SPT-LoRA. Compared to previous sparse tuning methods, SNELLA achieves a memory reduction of 31.1\%-39.9\% across models with parameter scales from 86M to 632M.
Our source codes are available at https://github.com/ssfgunner/SNELL.
\end{abstract}


%% file: sec/1_introduction.tex
\section{Introduction}
\begin{figure}
  \centering
  \includegraphics[width=0.92\linewidth]{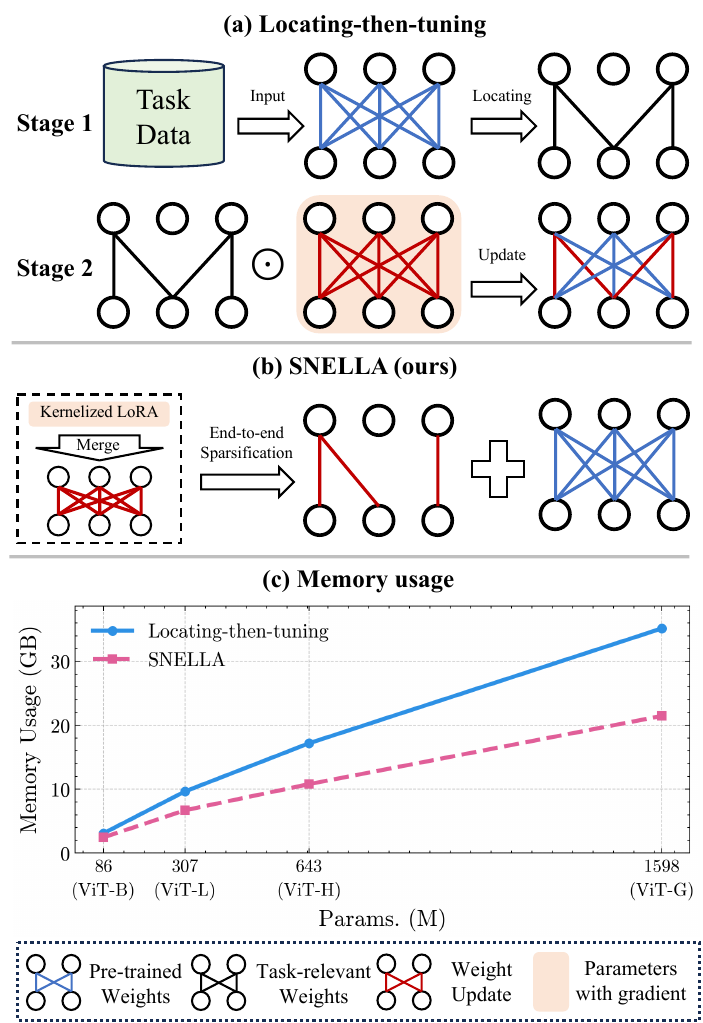}
  \vspace{-0.2cm}
  \caption{(a) The two-stage paradigm first locates task-relevant weights based on gradients and then directly updates the located weights.
  (b) SNELLA updates the pre-trained weights via a sparse matrix merged by low-rank matrices. 
  (c) Our method enables locating and updating task-relevant weights in an end-to-end manner with low memory usage.}
  \label{fig: introduction}
  \vspace{-0.5cm}
\end{figure}
\IEEEPARstart{F}{ine-tuning} has become a predominant way for adapting pre-trained vision models to a wide spectrum of downstream tasks~\cite{chen2020simple,he2020momentum,he2022masked, liang2025parameter, kirillov2023segment, esser2024scaling}.
Nevertheless, it is known that fine-tuning all model parameters requires substantial memory usage and is susceptible to over-fitting, making it costly and infeasible on large-scale pre-trained models given limited resources~\cite{zhai2022scaling,bai2023sequential,dai2021coatnet}.
Inspired by fine-tuning methods developed in the study of large language models, researchers leverage the parameter-efficient fine-tuning~(PEFT)~\cite{zhao2020masking,hu2021lora,zhang2024neural, jia2022visual, chen2022adaptformer, he2023sensitivity} to address these limitations by tuning a small subset of model parameters while keeping others frozen.
Current PEFT methods can be categorized into addition-based and reparameterization-based methods.
The former attaches additional trainable parameters to a frozen pre-trained vision backbone, while the latter adjusts the original parameters in the pre-trained vision backbone.

Addition-based methods~\cite{tu2023visual, zhang2024neural, jia2022visual} have achieved remarkable performance on vision tasks. 
However, the inclusion of additional parameters incurs extra inference cost. Reparameterization-based methods~\cite{zaken2021bitfit, caelles2017one, hu2021lora} select and adjust specific pre-trained parameters at the matrix level or weight level, involving reduced memory usage compared to full-parameter fine-tuning. 
Matrix-level methods focus on the whole weight matrix. For example, Bitfit~\cite{zaken2021bitfit} adjusts bias to reduce the volume of tunable parameters, while Partial-\textit{k}~\cite{jia2022visual} fine-tunes the weight matrices only in the last few layers. 
To fine-tune arbitrary layers with low memory usage, LoRA~\cite{hu2021lora} optimizes each layer using two additional low-rank matrices.
Despite reduced memory usage, these methods usually gain inferior accuracy compared to addition-based ones~\cite{jia2022visual}.
Given that vision downstream tasks often involve limited samples~\cite{zhai2019large, jha2020medico, ruiz2023dreambooth}, sparse tuning enables adjustment on individual weights, mitigating the over-fitting risks~\cite{fu2023effectiveness}. For example, SPT~\cite{he2023sensitivity}, a weight-level strategy that only adjusts the most task-relevant weights in a matrix, shows a promising ability of sparse tuning.

\begin{figure}
  \centering
  \includegraphics[width=0.93\linewidth]{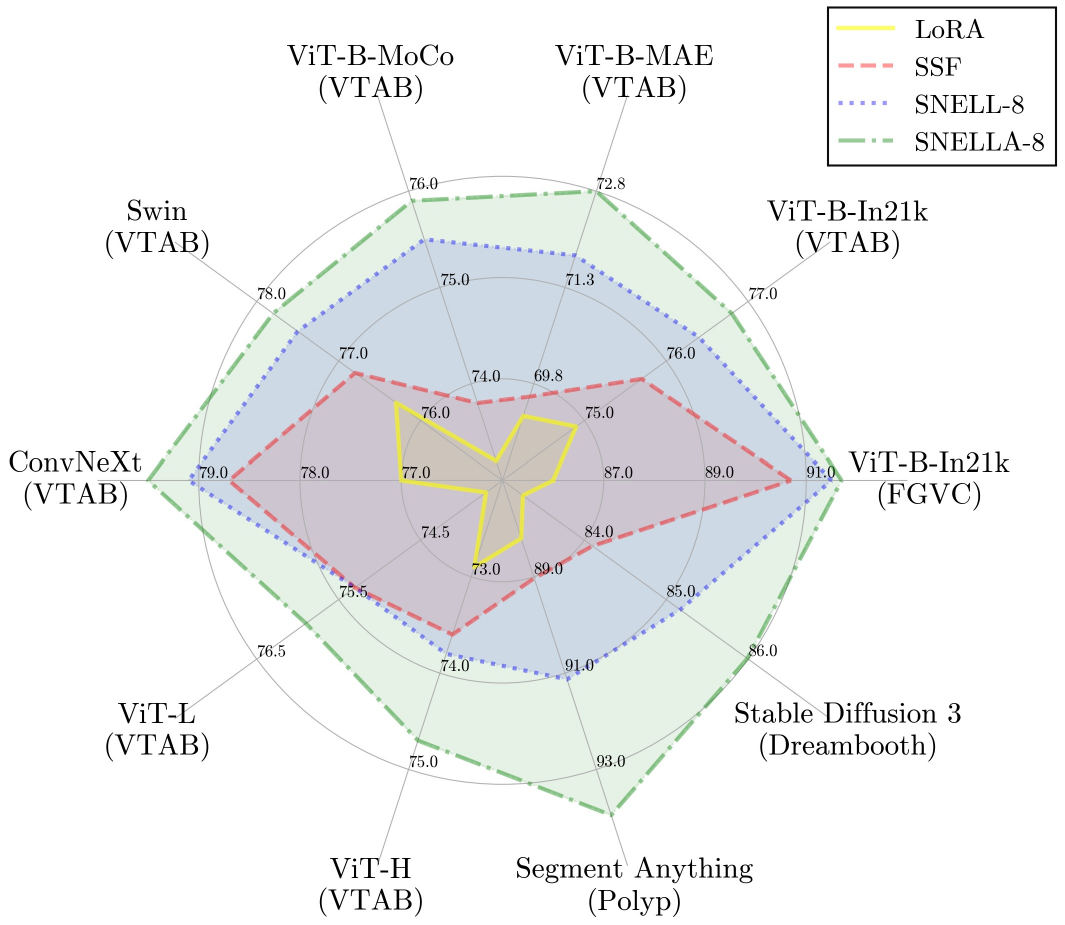}
  \vspace{-0.4cm}
  \caption{Performance comparisons between LoRA~\cite{hu2021lora}, SSF~\cite{lian2022scaling}, SNELL-8~\cite{SNELL} and SNELLA-8 across different pre-trained models and benchmarks. SNELLA demonstrates superior performance over others.}
  \label{fig: radar}
  \vspace{-0.5cm}
\end{figure}

Current sparse tuning methods generally adopt a locating-then-tuning paradigm as shown in Figure~\ref{fig: introduction}(a), which first locates task-relevant weights based on gradients and then fine-tunes them, facing issues of increased memory usage and distracted weight locating result.
For memory usage, it selectively updates weights by masking out the gradients of task-irrelevant weights. Despite only updating part of the weights in the pre-trained weight matrix, the whole matrix still needs to be stored as learnable parameters in the optimizer for gradient computation. 
In other words, this paradigm offers no advantage over full fine-tuning regarding memory usage, especially when the parameter scales of pre-trained models increase~\cite{zhai2022scaling, bai2023sequential}. 
Furthermore, this scheme may fail to locate the tunable weights relevant to downstream tasks, as it separates the locating and fine-tuning stages, neglecting the dynamic evolution of weight values along the adaptation process. 

In this paper, 
to facilitate sparse fine-tuning of pre-trained vision models with low memory usage,
we propose a one-stage \textbf{S}parse tuning method with ker\textbf{NEL}ized \textbf{L}oRA and \textbf{A}daptive bi-level sparsity allocation~(SNELLA), as shown in Figure~\ref{fig: introduction}(b).
Specifically, for \textit{memory usage}, we add the pre-trained weight matrix with another matrix merged by two low-rank matrices. The sparse adjustment is achieved by sparsifying this merged matrix. Compared to the locating-then-tuning paradigm that stores the whole weight matrix, only the low-rank matrices need to be stored in the optimizer, resulting in significantly reduced memory usage. Nevertheless, directly merging low-rank matrices through inner product~\cite{hu2021lora} preserves the low-rank structure, making the weight updates highly coupled with each other, especially when the matrix rank is low. This coupling nature restricts the independent adjustment of task-relevant weights, hindering their expressivity for adapting to downstream tasks.
To address this limitation, we propose the kernelized LoRA that constructs a high-rank matrix by merging low-rank matrices with nonlinear kernels instead of the inner product. For the kernel function, we observe a tradeoff between expressivity and optimization stability of existing kernels in fine-tuning scenarios. To break this tradeoff, we introduce a mixture of piecewise linear and normalized RBF kernels~(Mix-K) inspired by multi-kernel learning~\cite{gonen2011multiple}. Mix-K achieves high expressivity and optimization stability, enabling better adaptability to downstream tasks.

For \textit{locating task-relevant weights}, we propose an adaptive bi-level sparsity allocation mechanism that encourages end-to-end parameter competition at both the layer and weight levels. 
At the layer level, given a fixed number of tunable weights as a budget, layers 
compete based on their sensitivity scores~\cite{zhang2023adalora}, which quantify their relevance to downstream tasks. Layers with higher scores are allocated a larger number of tunable weights.
At the weight level, competition occurs among weights within each layer based on the magnitude of their updates. Those task-relevant weights are more likely to achieve significant updates and survive through the sparsification process. 
Unlike existing end-to-end sparsification methods~\cite{louizos2017learning} that impose additional penalty terms on loss functions, the competition-based mechanism preserves the original optimization objective, ensuring a stronger ability for locating and updating task-relevant weights.

We apply SNELLA to a wide range of downstream tasks, \textit{i.e.,} image classification~\cite{jia2022visual, zhai2019large}, medical image segmentation~\cite{jha2020medico}, and text-to-image generation~\cite{ruiz2023dreambooth}, with various pre-trained vision models~(\textit{i.e.,} ViT-B/L/H~\cite{dosovitskiy2020image}, Swin-B~\cite{liu2021swin}, ConvNeXt-B~\cite{liu2022convnet}, SAM~\cite{kirillov2023segment} and SD3~\cite{esser2024scaling}) with supervised and self-supervised pre-training strategies, see Figure~\ref{fig: radar}. SNELLA achieves advanced performance in classification and segmentation tasks with low memory usage. For generation tasks, SNELLA can adapt the pre-trained model to new concepts while preserving high visual generation ability.
Our main contributions are summarized as follows:
\begin{itemize}
    \item We propose SNELLA, an end-to-end sparse tuning framework that locates and updates the task-relevant weights during fine-tuning with low-memory usage.
    \item We propose kernelized LoRA that merges low-rank matrices with nonlinear kernel functions to largely reduce the memory usage of sparse tuning while preserving the representation ability remarkably.
    \item We present an end-to-end sparsification mechanism to locate task-relevant weights, improving the effectiveness of sparse tuning to downstream tasks.
    \item We evaluate SNELLA on various downstream tasks using multiple pre-trained models. The results demonstrate that SNELLA consistently outperforms strong competitors across a wide spectrum of vision tasks while achieving remarkably low memory usage.
\end{itemize}

This paper provides a more general sparse tuning framework compared to our preliminary study SNELL~\cite{SNELL}. 
\textbf{(1)}~More comprehensive analysis of kernels in fine-tuning scenarios. SNELL uses the piecewise linear kernel function, sacrificing expressivity for optimization stability. The limited expressivity restricts the rank of the merged matrix. In contrast, this paper presents a more comprehensive analysis of the applicability of existing kernels to fine-tuning tasks, revealing that the trade-off between optimization stability and expressivity in existing kernels stems from the gradient vanishing of the exponential kernels. Then, we construct a kernel that exhibits strong expressivity and optimization stability in fine-tuning scenarios through column-wise normalization and a mixture of piecewise linear and exponential kernels.
\textbf{(2)}~A bi-level sparsity allocation mechanism. SNELL simply assigns an identical number of tunable weights to all layers, promoting weight competition within each layer. This strategy leads to sub-optimal locating results as different layers contribute variably to distinct downstream tasks~\cite{he2023sensitivity}. 
In contrast, we propose a competition mechanism across layers for SNELLA. By designing a sensitivity-based importance score, different layers compete to obtain more tunable weights. This mechanism enables more flexible and globally optimal adjustment of tunable weights across layers according to downstream tasks, \textit{i.e.}, more task-relevant layers acquire weaker sparsity constraints.
\textbf{(3)} Improved performance on more vision tasks. SNELL is only evaluated on classification benchmarks including FGVC~\cite{jia2022visual} and VTAB-1k~\cite{zhai2019large}, while this paper extends to segmentation~\cite{jha2020medico} and text-to-image generation~\cite{ruiz2023dreambooth} with more pre-trained models~\cite{kirillov2023segment, esser2024scaling}. As shown in Figure~\ref{fig: radar}, SNELLA consistently outperforms SNELL across various pre-trained models and downstream tasks.

%% file: sec/2_relatedwork.tex
\section{Related Work}

\subsection{Parameter-efficient Fine-tuning}
Parameter-efficient fine-tuning~\cite{chen2022adaptformer, jia2022visual, zhang2024neural, yao2025bi, qiao2025gradient, kim2024prompt, chen2025graph} can efficiently adapt pre-trained models to downstream tasks by tuning only a tiny portion of parameters.
Current methods can be categorized into addition-based~\cite{bapna2019simple,houlsby2019parameter,pfeiffer2020adapterfusion,sung2022vl,ding2021openprompt,ju2022prompting,liu2021p,zhang2024neural} and \textit{reparameterization-based}~\cite{zaken2021bitfit,guo2020parameter,hu2021lora} methods. 

\textit{Addition-based} methods attach additional trainable parameters to a frozen pre-trained backbone.
Adapters~\cite{bapna2019simple,houlsby2019parameter,pfeiffer2020adapterfusion,sung2022vl,zhang2021tip} adopt a residual pathway and learn a bottleneck layer with two linear projections and a non-linear activation.
Prompt-tuning~\cite{ding2021openprompt,ju2022prompting,liu2021p,li2021prefix} adds trainable parameters to the input and keeps the entire pre-trained model unchanged during training.
Recent work~\cite{zhang2024neural} attempts to find the optimal configurations to combine multiple addition-based methods.
Despite the effectiveness of addition-based methods, the additional trainable parameters incur excess computational costs during the inference process~\cite{bapna2019simple,li2022cross}.

Reparametization-based methods adjust the inherent parameters in pre-trained models to avoid excess computational costs during inference.
Early work directly selects parameters with low memory usage for fine-tuning, such as the bias terms~\cite{zaken2021bitfit} and the final few layers~\cite{caelles2017one}.
To further reduce the memory usage of tuning the selected matrices, LoRA~\cite{hu2021lora} optimizes low-rank matrices that can be reparameterized into the pre-trained weight matrices.
Exploring finer-grained parameter selection, researchers propose sparse tuning~\cite{guo2020parameter,zhao2020masking}, which involves selecting and tuning individual weights sparsely within the weight matrices.
SPT~\cite{he2023sensitivity} combines sparse tuning and LoRA in a hybrid framework. 
SPT has revealed that optimizing the weights most relevant to the downstream task through sparse tuning can significantly enhance the performance, which is also supported by GPS~\cite{DBLP:conf/cvpr/ZhangZGZSZZ24}.
However, these methods follow a two-stage paradigm that faces the challenge of increased memory usage and distracted weight locating results.
In contrast, our SNELLA introduces a one-stage framework that achieves both high performance and low memory usage by updating pre-trained weight matrices with an additional matrix, which is merged by low-rank matrices and sparsified in an end-to-end manner.

\subsection{Low-Rank Matrix Factorization}
Low-rank matrix factorization, which approximates high-dimensional matrices using low-dimensional counterparts~\cite{davenport2016overview}, is widely used in model fine-tuning~\cite{hu2021lora} and compression~\cite{wang2024svd}. This approach significantly reduces memory usage by storing only the low-rank matrices as learnable parameters. However, the constrained parameter space leads to lower expressivity than full-parameter fine-tuning on downstream tasks~\cite{zhao2024galore}. 
To approximate full fine-tuning with low-rank matrices, PISSA~\cite{meng2024pissa} designs initialization strategies to approximate the learning behavior of full-parameter optimization.
ReLoRA~\cite{lialin2023relora} improves expressivity by merging multiple sets of low-rank matrices. 
GaLore~\cite{zhao2024galore} represents the gradients with a low-rank structure to approximate full fine-tuning.
These methods still rely on linear approaches that inherently limit expressivity. It has been shown that introducing nonlinearity into matrix factorization may result in a better approximation and representation ability~\cite{gonen2013kernelized, rendle2008online}.
Nonlinear approaches such as the Fourier transform~\cite{gao2024parameter} and softmax-based gating~\cite{tian2024hydralora} have been introduced to enhance the expressivity of low-rank matrices.
Unlike existing methods that decompose nonlinear kernel matrices with linear formulation or introduce nonlinearity via heuristic designs, we extend the low-rank factorization from the kernel perspective, inspired by neuron dynamics~\cite{pei2023dynamics}, offering a more general nonlinear matrix computation framework for model fine-tuning, and introduce a mixture of nonlinear kernels that exhibits both high expressivity and training stability in fine-tuning scenarios.

\subsection{Weight Sparsity}
The weight sparsity~\cite{wang2021learning, peng2022exact, liu2022learning, niu2021grim, zhang2023lottery, shen2025enhancing, shen2025vlsae, liu2025edit} is regarded as an important optimization objective in model pruning towards highly efficient model computing~\cite{han2016eie,molchanov2017variational,liu2021discrimination}.
To minimize the performance degradation during pruning, these methods first determine the sparsity level for each layer and then remove the task-irrelevant weights within layers. 
For sparsity allocation across layers, GRIFFIN~\cite{dong2024prompt} selects weights based on their high activation magnitudes in response to input prompts. FLAP~\cite{an2024fluctuation} computes the sample variance of each input feature to assess layer importance and allocate sparsity accordingly. RL-Pruner~\cite{wang2024rl} determines the layer-wise sparsity distribution through reinforcement learning. 
Furthermore, the task relevance of individual weights can be estimated with activations~\cite{hu2016network}, redundancy~\cite{srinivas2015data}, second derivatives~\cite{dong2017learning}, and energy efficiency~\cite{yang2017designing}. 
In parallel to the post-training pruning, sparse neural networks~\cite{bellec2017deep,louizos2017learning,frankle2018lottery} introduce sparsity into the pre-training stage, removing redundant weights more precisely~\cite{frankle2018lottery}.
Rather than sparsifying the weight matrices as an objective, our study focuses on the end-to-end locating and updating of task-relevant weights via competitions at layer and weight levels, relying solely on low-rank learnable matrices.

%% file: sec/3_method.tex
\section{Method}

\begin{figure*}
  \centering
  \includegraphics[width=0.95\linewidth]{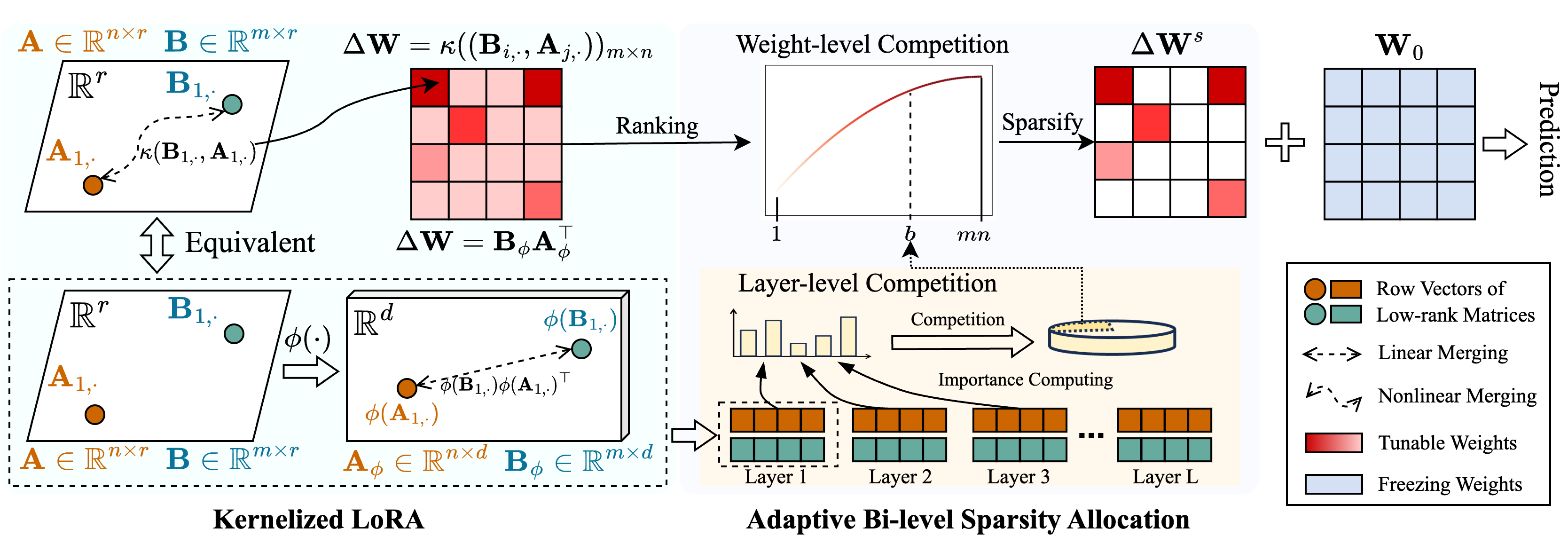}
  \vspace{-0.3cm}
  \caption{Overview of our SNELLA strategy. Given two learnable low-rank matrices, we merge them using a non-linear kernel function~(\textit{left}). This merging process is equivalent to mapping the matrices into higher-rank matrices and then performing matrix multiplication. Then we sparsify this merged matrix using an adaptive sparsity allocation mechanism (right). First, the layers compete with each other to determine their number of tunable weights $b$. Then, competition within layers is conducted by preserving the top-$b$  weight updates and setting the remaining updates to zero.}
  \label{fig: framework}
  \vspace{-0.3cm}
\end{figure*}

\subsection{Preliminaries}
\label{subsec: preliminaries}
\noindent\textbf{Sparse Tuning}.
Given a downstream training set $\mathcal{D}=\{x^{(n)}, y^{(n)}\}_{n=1}^N$, the objective of sparse tuning is to minimize the model's empirical risk on downstream task, with the sparsity constraints on the number of tunable weights in weight matrix $\mathbf{W}\in\mathbb{R}^{m\times n}$.
The sparsification is usually achieved by multiplying the gradient of weights with a pre-defined binary mask $\mathbf{M}\in\{0,1\}^{m\times n}$ as follows
\begin{equation}
    \min_{\mathbf{W}\odot\mathbf{M}} \frac{1}{N} \sum_{n=1}^N \mathcal{L}\left(f(x^{(n)} ; \mathbf{W}), y^{(n)}\right),
\label{eq: training_objective}
\end{equation}
where $f(\cdot; \cdot)$ is a parameterized function over the input (\textit{e.g.}, a neural network), $\mathcal{L}(\cdot, \cdot)$ is a loss function (\textit{e.g.}, cross-entropy), and $\odot$ denotes element-wise multiplication.
The binary mask $\mathbf{M}$ determines the tunable weights and is typically pre-computed with heuristics such as gradients~\cite{he2023sensitivity}.
Despite having fewer learnable parameters, $\mathbf{W}\odot\mathbf{M}$ occupies the same amount of memory as the weight matrix $\mathbf{W}$ in practice. 
As a result, the memory usage of current sparse tuning methods is even higher than that of full fine-tuning.
Moreover, the pre-defined gradient masking $\mathbf{M}$ fails to assess the task relevance of weights, as it neglects dynamic changes in weight values during fine-tuning. Treating $\mathbf{M}$ as parameters leads to the unstable training process~\cite{zhao2020masking} and incurs additional memory usage due to the storage of $\mathbf{M}$.

\noindent\textbf{Low-Rank Adaptation~(LoRA)}.
Given a pre-trained weight matrix $\mathbf{W}_0$, LoRA~\cite{hu2021lora} optimizes two low-rank matrices $\mathbf{A}\in\mathbb{R}^{n\times r}$ and
$\mathbf{B}\in\mathbb{R}^{m\times r}$ 
to reduce the memory usage during fine-tuning. 
The low-rank matrices can be reparameterized into the pre-trained weight $\mathbf{W}_0$ as,
\begin{equation}
    \mathbf{W} = \mathbf{W}_0 + \Delta \mathbf{W} = \mathbf{W}_0 + \mathbf{BA}^\top.
\end{equation}
For $r\ll \min(m,n)$, LoRA achieves high training efficiency and low memory usage through the low-rank matrices.

\noindent\textbf{Kernel Trick}~\cite{koutroumbas2008pattern}.
Mapping the vectors into higher dimensions is frequently used to achieve linear separability in many machine learning tasks~\cite{suykens1999chaos}.
However, the explicit mapping process incurs significant computational costs. To address this problem, the kernel trick is proposed to model the data relationships in high-dimensional spaces, without the need to formulate the space explicitly. According to Mercer's theorem~\cite{berlinet2011reproducing}, a kernel function $\kappa: \mathbb{R}^r\times \mathbb{R}^r \rightarrow \mathbb{R}$ can express an inner product in some space as $\kappa(\mathbf{x}, \mathbf{x'}) = \phi(\mathbf{x})^\top \phi(\mathbf{x'})$, if and only if $\kappa$ is positive semi-definite~(Appendix A.3). $\mathbf{x}, \mathbf{x'} \in \mathbb{R}^r$, and $\phi: \mathbb{R}^r \rightarrow \mathbb{R}^d$ is an implicit feature map.
By selecting an appropriate kernel function $\kappa$, we can obtain the inner product of two vectors in higher-dimensional space $\mathbb{R}^d~(d\ge r)$ without explicitly formulating the feature mapping $\phi$.

\begin{figure*}
  \centering
  \includegraphics[width=0.97\linewidth]{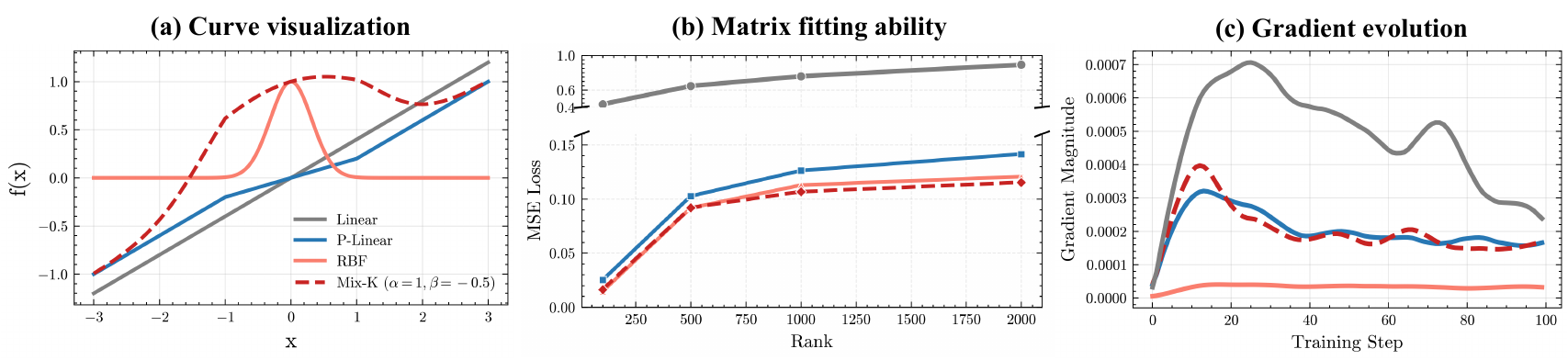}
  \vspace{-0.5cm}
  \caption{(a) Visualization examples of multiple kernel functions in the one-dimensional form. 
  (b) To evaluate the expressivity of different kernels, we fit random sparse matrices with varying ranks by merging two learnable low-rank matrices with these kernels and compute the MSE loss. 
  (c) Gradient evolution of different kernels during fine-tuning. Experiments are conducted on the Stanford-Cars dataset using pre-trained ViT-B/16.}
  \label{fig: kernel_design}
  \vspace{-0.6cm}
\end{figure*}

\subsection{Expanding LoRA in the Kernel Perspective}
\label{subsec: k_lora}
\noindent\textbf{Applying LoRA to Sparse Tuning.}
We leverage low-rank matrices to reduce the memory usage of sparse tuning.
An intuitive solution is to sparsify the adaptation matrix $\Delta \mathbf{W}$ merged by the two low-rank matrices like LoRA~\cite{hu2021lora}.
Under this setting, Equation~\ref{eq: training_objective} becomes:
\begin{equation}
    \min_{\mathbf{A}, \mathbf{B}} \frac{1}{N} \sum_{n=1}^N \mathcal{L}\left(f(x^{(n)} ; \mathbf{W_0}+\Delta \mathbf{W}\odot\mathbf{M}), y^{(n)}\right),
\label{eq: training_objective_lora}
\end{equation}
Nevertheless, in practice, directly using LoRA with limited expressivity can lead to the performance degradation of sparse tuning.
For the original sparse tuning, the weight matrix $\mathbf{W}$ is free of the rank constraint, and weight updates are independent of each other. 
The independence of weight updates enables task-relevant weights to adapt flexibly to the specific requirements of downstream tasks.
For sparse tuning with LoRA, the low-rank structure imposes strong smoothness and coupling effect among weight updates~(\textit{i.e.}, elements in $\Delta \mathbf{W}$), limiting the ability of task-relevant weights to be independently adjusted according to downstream tasks. 
Therefore, we introduce the kernel trick to LoRA to
ensure the high rank of $\Delta \mathbf{W}$ based on low-rank learnable matrices $\mathbf{A}$ and $\mathbf{B}$.

\noindent\textbf{Kernelized LoRA.}
We extend LoRA from a kernel perspective and propose constructing a high-rank matrix by low-rank matrices with nonlinear kernels, inspired by DyN~\cite{pei2023dynamics}.
Given two vectors $\boldsymbol{x}, \boldsymbol{x'}\in\mathbb{R}^r$, the kernel function $\kappa(\boldsymbol{x}, \boldsymbol{x'})$ can be formulated as an inner product $\phi(\boldsymbol{x})^\top\phi(\boldsymbol{x'})$ with an implicit feature map $\phi: \mathbb{R}^r \rightarrow \mathbb{R}^d$.
The merging process of LoRA can be interpreted as applying a linear kernel function $\kappa_l(\cdot,\cdot)$ on the rows of the learnable parameters $\mathbf{A}$ and $\mathbf{B}$,
\begin{equation}
    \Delta \mathbf{W}_{ij} = \kappa_l(\mathbf{A}_{j,\cdot}, \mathbf{B}_{i,\cdot}) = \phi_l(\mathbf{B}_{i,\cdot})\phi_l(\mathbf{A}_{j,\cdot})^\top=\mathbf{B}_{i,\cdot}\mathbf{A}_{j,\cdot}^\top,
\end{equation}
where $\mathbf{A}_{j,\cdot}, \mathbf{B}_{i,\cdot}\in\mathbb{R}^r$, $\phi_l: \mathbb{R}^r \rightarrow \mathbb{R}^r$ denotes the identity mapping.
By replacing $\kappa_l(\cdot,\cdot)$ with more complex non-linear kernel functions, we can approximate inner production in higher-dimensional spaces $\mathbb{R}^d$ and obtain matrices with rank larger than $r$.
The merged adaptation matrix $\Delta \mathbf{W}$ in SNELLA can be represented by 
\begin{equation}\label{eq: klora}
\begin{split}
\Delta \mathbf{W} &= (\kappa(\mathbf{A}_{i,\cdot}, \mathbf{B}_{j,\cdot}))_{m\times n} \\ 
&=[\phi(\mathbf{B}_{1,\cdot})^\top,...,\phi(\mathbf{B}_{m,\cdot})^\top]^\top[\phi(\mathbf{A}_{1,\cdot})^\top,...,\phi(\mathbf{A}_{n,\cdot})^\top] \\ 
&= \mathbf{B}_\phi \mathbf{A}_\phi^\top.
\end{split}
\end{equation}
Note that in practice, explicit computation of high-rank matrices $\mathbf{A}_\phi\in\mathbb{R}^{n\times d}$ and $\mathbf{B}_\phi\in\mathbb{R}^{m\times d}$ is unnecessary. $\Delta \mathbf{W}$ can be directly derived based on low-rank matrices $\mathbf{A}$ and $\mathbf{B}$ with the non-linear kernel function $\kappa$.
By extending LoRA in a kernel perspective, one can build high-rank adaptation matrices based on low-rank learnable matrices, empowering strong sparse tuning with low memory usage.

\begin{figure}
  \centering
  \includegraphics[width=0.95\linewidth]{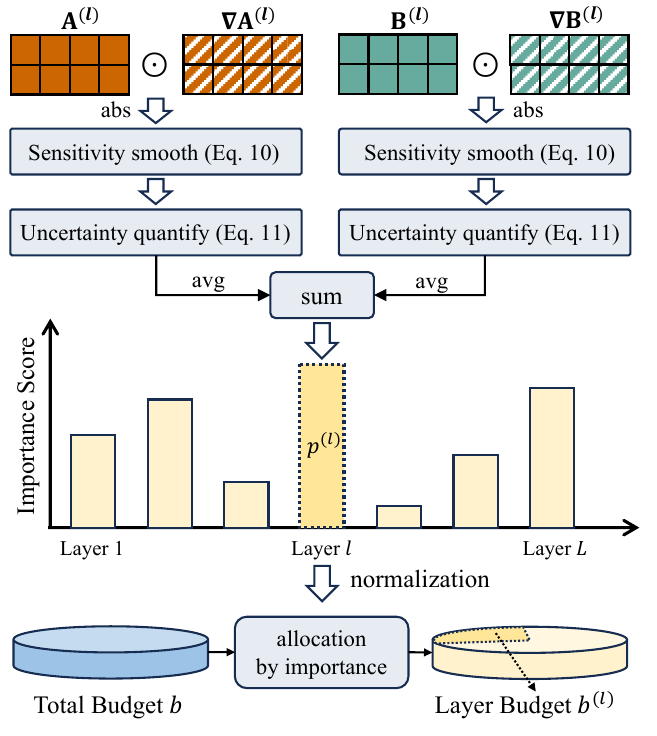}
  \vspace{-0.4cm}
  \caption{Layer-level competition mechanism. We integrate both sensitivity and uncertainty to compute an importance score for each layer. These scores then serve as the basis for competition among layers, enabling more important layers to gain a larger number of tunable parameters.}
  \label{fig: sparsity_allocation}
  \vspace{-0.2cm}
\end{figure}

\noindent\textbf{Kernel Function Analysis.}
According to Equation~\ref{eq: klora}, the nonlinearity of the kernel function determines the upper bound of the expressivity of kernelized LoRA~(\textit{i.e.,} the rank of $\mathbf{A}_\phi$ and $\mathbf{B}_\phi$).
To select a kernel function that performs well in model fine-tuning scenarios with strong expressivity, we analyze the curve, fitting ability, and gradient evolution of several kernel functions in Figure~\ref{fig: kernel_design}. 
For the linear kernel, its expressivity is constrained by the low-rank structure of its merged matrix. This is reflected by the significantly higher MSE loss compared to that of other nonlinear kernels, as shown in Figure~\ref{fig: kernel_design}(b).
For the piecewise linear kernel~(P-Linear), its expressivity is better than that of the linear kernel while remaining inferior to the RBF kernel, which exhibits the strongest expressivity by incorporating exponential families. Nevertheless, as shown in Figure~\ref{fig: kernel_design}(a), the gradient of the RBF kernel tends to be zero as the magnitude of the input increases, which leads to gradient vanishing in Figure~\ref{fig: kernel_design}(c) and then hinders the end-to-end optimization.

To leverage the expressivity of the RBF kernel with training efficiency, we first normalize each column of the merged matrix. The normalization scales the matrix values into the region with significant gradients of the exponential kernel, thereby preventing the gradient vanishing. 
However, the normalization constrains the range of kernel function values and introduces additional dependencies among them, which can reduce expressivity. To address this limitation, we propose a mixture of piecewise linear and normalized RBF kernels~(Mix-K), extending the kernel's value range while preserving optimization stability.
Specifically, the $(i,j)$-th element of the merged matrix $\Delta \mathbf{W}_{ij}$ is calculated as follows,
\begin{equation}
\begin{aligned}
    &\Delta \mathbf{W}_{ij} = \kappa_p(\mathbf{A}_{j,\cdot}, \mathbf{B}_{i,\cdot}) + \alpha \frac{\exp(\kappa_p(\mathbf{A}_{j,\cdot}, \mathbf{B}_{i,\cdot}))}{\sum_{k=1}^{m}\exp(\kappa_p(\mathbf{A}_{k,\cdot}, \mathbf{B}_{i,\cdot}))}+\beta,\\
    &\kappa_p(\mathbf{a}, \mathbf{b})=\sum\limits_{p=1}^P\alpha_p\Vert \mathbf{a}_{\lceil rp/P\rceil :\lceil r(p+1)/P\rceil}-\mathbf{b}_{\lceil rp/P\rceil :\lceil r(p+1)/P\rceil} \Vert_2,
\end{aligned}
\end{equation}
where $\alpha, \beta\in\mathbb{R}$ are learnable parameters, $P$ denotes the number of pieces. $\kappa_p$ denotes the piecewise linear kernel function~\cite{SNELL} that partitions the input vector pairs into several pieces of equal length, computes the $l_2$ distance between corresponding pieces, and then sums these distance values. By combining both kernel functions at different scales, the expressivity of the mixed kernel can be effectively enhanced. 
As shown in Figure~\ref{fig: kernel_design}, with strong nonlinearity, the mixed kernel demonstrates superior expressivity compared to existing kernel functions, while effectively avoiding the gradient vanishing during training.

\begin{figure}
  \centering
  \includegraphics[width=1.0\linewidth]{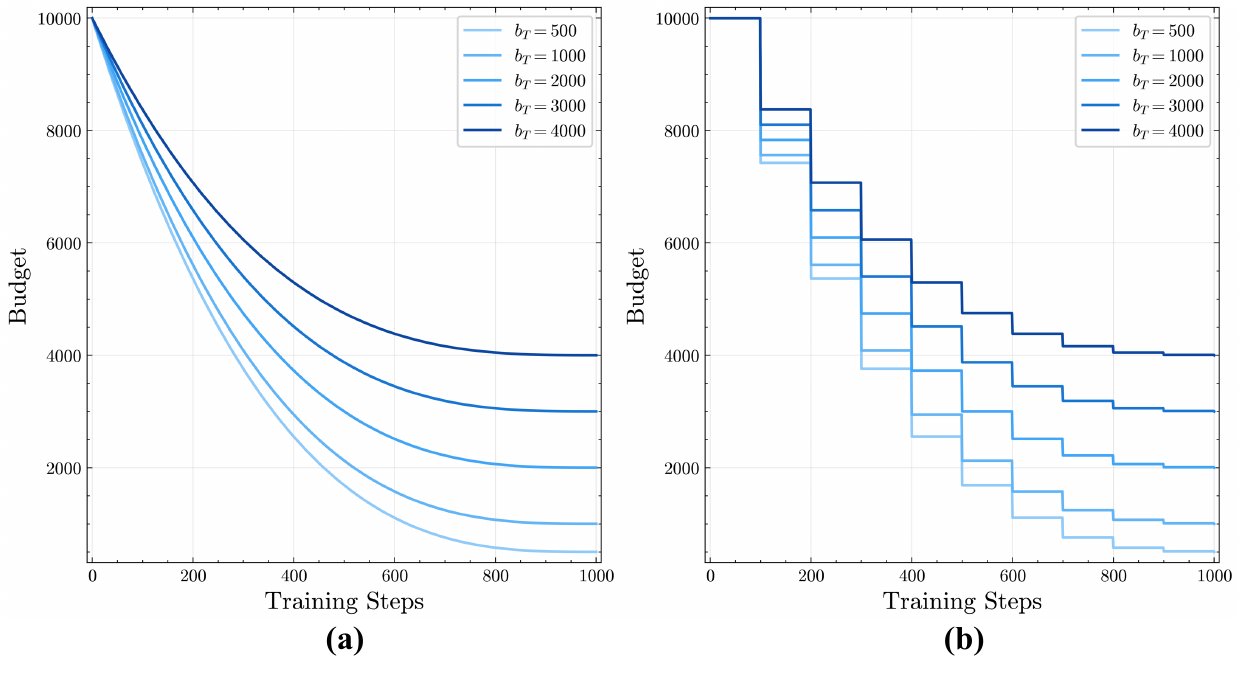}
  \vspace{-0.6cm}
  \caption{(a) The evolution of the number of tunable weights $b_t$ during the fine-tuning process under varying final values $b_T$. (b) In practice, we adjust $b_t$ at the end of each epoch rather than at every training step.}
  \label{fig: budget_evolution}
  \vspace{-0.5cm}
\end{figure}

\subsection{Adaptive Bi-level Sparsity Allocation}\label{subsec: sparsification_mechanism}

After merging matrices by kernelized LoRA at each layer, they are sparsified to determine which pre-trained weights should be tuned. First, we define $b_t$ as the number of tunable weights within the model at the $t$-th step, which serves as a total budget. Based on this budget, an adaptive bi-level sparsity allocation mechanism is proposed to locate and update the most task-relevant weights during fine-tuning, avoiding the weakness of gradient masking $\mathbf{M}$ in Equation~\ref{eq: training_objective_lora}. It first allocates the budget across layers through layer-level competition, and then determines the specific tunable weights of each layer via weight-level competition.

\begin{algorithm}[htp]
\caption{Sparsity allocation across layers.}\label{alg:alg1}
\begin{algorithmic}
\REQUIRE Importance score $P=\{p^{(l)}\}_{l=1}^L$, the allocable number of weights $\overline{B}=\{\overline{b}^{(l)}\}_{l=1}^L$ for $L$ layers, the number of tunable weights~(total budget) $b$. 
\ENSURE Number of tunable weights for $L$ layers $\{b^{(l)}\}_{l=1}^L$.

\STATE \hspace{-0.35cm}{\textbf{ALLOC}}($P, \overline{B}, b$):
\STATE $b_r \gets b$ \textcolor[HTML]{006400}{\# remaining budget}
\STATE $b^{(l)} \gets 0$ for $l$ in $\{1,...,L\}$ \textcolor[HTML]{006400}{\# Initialization}
\WHILE{$b_r\neq 0$}
\STATE $p_{sum}=\sum_{l=1}^Lp^{(l)}$
\FOR{$ l\in\{1,2,...,L\} $}
\STATE $ p^{(l)} \gets p^{(l)}/p_{sum}$ \textcolor[HTML]{006400}{\# normalize importance scores}
\STATE $ b^{(l)} \gets \min( b^{(l)}+p^{(l)}\times b_r, \overline{b}^{(l)})$ \textcolor[HTML]{006400}{\# allocate budget}
\IF{$b^{(l)}=\overline{b}^{(l)}$} 
\STATE $p^{(l)} \gets 0$ \textcolor[HTML]{006400}{\# remove non-allocable layers}
\ENDIF
\ENDFOR
\STATE $b_r\gets b-\sum_{l=1}^Lb^{(l)}$ \textcolor[HTML]{006400}{\# update remaining budget}
\ENDWHILE
\RETURN $\{b^{(l)}\}_{l=1}^L$

\end{algorithmic}
\end{algorithm}

\input{tab/tab_fgvc_vtab1k}

\noindent\textbf{Layer-level Competition}. 
As Figure~\ref{fig: sparsity_allocation} shows, given the budget $b_t$ and learnable matrices of $L$ layers $\{(\mathbf{A}^{(l)}, \mathbf{B}^{(l)})\}_{l=1}^L$ at the $t$-th optimization step, we distribute the budget to layers through competition based on their task relevance, \textit{i.e.}, more task relevant layers have more tunable weights.

First, we utilize a sensitivity-based metric to measure the task relevance of layers inspired by AdaLoRA~\cite{zhang2023adalora}. 
With the learnable matrices $(\mathbf{A}^{(l)}, \mathbf{B}^{(l)})$ at the $l$-th layer, the matrix sensitivity $I(\cdot)$ is defined as the magnitude of the element-wise multiplication between this matrix and its gradient,
\begin{equation}\label{eq: sensitivity}
    I(\mathbf{A}^{(l)}) = \vert\nabla \mathbf{A}^{(l)} \odot \mathbf{A}^{(l)} \vert\in\mathbb{R}^{n\times r}
\end{equation}
However, the sensitivity score is not yet a reliable importance metric, because it is estimated on the sampled mini-batch. The stochastic sampling and complicated training dynamics incur high variability and large uncertainty for estimating the sensitivity during fine-tuning. We solve this issue by sensitivity smoothing and uncertainty quantification following~\cite{zhang2023adalora}. Taking $\mathbf{A}^{(l)}_t$ as an example, at $t=0$, the smoothed sensitivity $\overline{I}_A^{(l)}(0)$ is initialized to the sensitivity value, while the smoothed uncertainty $\overline{U}_A^{(l)}(0)$ is set to zero.
\begin{equation}
\overline{I}_A^{(l)}(0)=I(\mathbf{A}^{(l)}),~~~\overline{U}^{(l)}_A(0)=0.
\end{equation}
When $t>0$, the smoothed sensitivity $\overline{I}_A^{(l)}(t)$ and uncertainty $\overline{U}_A^{(l)}(t)$ are defined as follows:
\begin{equation}
\overline{I}_A^{(l)}(t)=\beta_1\overline{I}_A^{(l)}(t-1) + (1-\beta_1)I(\mathbf{A}^{(l)}),
\end{equation}
\begin{equation}
\overline{U}^{(l)}_A(t)=\beta_2\overline{U}^{(l)}_A(t-1) + (1-\beta_2)\vert \overline{I}_A^{(l)}(t) - I(\mathbf{A}^{(l)})\vert
\end{equation}
where $\beta_1,\beta_2\in[0,1]$ are hyper-parameters used for smoothing.  The uncertainty term is quantified by the local variation between the sensitivity and its smoothed counterpart.
As the training step $t$ increases, the sensitivity and uncertainty are smoothed using the exponential moving average.
The importance of the $l$-th layer at the $t$-th optimization step $p^{(l)}_t$ is defined as the average of the product between sensitivity and uncertainty within the two low-rank matrices,
\begin{equation}
    p^{(l)}_t = \mathrm{avg}(\overline{I}^{(l)}_A(t)\cdot\overline{U}^{(l)}_A(t))) + \mathrm{avg}(\overline{I}^{(l)}_B(t)\cdot\overline{U}^{(l)}_B(t)).
\end{equation}

Next, we allocate the budget $b_t$ across layers in proportion according to their importance scores $\{p^{(l)}_t\}^L_{l=1}$. The budget allocated to each layer $\{b^{(l)}_t\}_{l=1}^L$ controls the intensity of weight-level competition, {\it i.e.}, less budget results in tougher competition among weights.
However, since the number of weights in a layer is determined beforehand, directly allocating the number of tunable weights proportionally to the importance score may assign highly important layers with tunable weights more than their total number of weights. To address this issue, we design a recursive algorithm that reallocates the remaining layer budget, {\it i.e.}, the overflowed number of tunable weights, to other layers that need more budget. Detailed implementation is presented in Algorithm~\ref{alg:alg1}. The optimization step $t$ is omitted for brevity.

Finally, the way to set the overall budget $b_t$ during the fine-tuning process is crucial to determine the overall performance.  Directly adopting a fixed value would randomly zero out a large number of weight updates at the early stage, hindering their importance computation. Therefore, $b_t$ is initialized to the total number of model weights $b_0$ and then gradually decreases to a predefined value $b_T$ as $t$ increases from $0$ to $T$,
\begin{equation}\label{eq: budget_scheduling}
    b_t = b_T+(1-\frac{t}{T})^3(b_0-b_T).
\end{equation}
We adopt the cubic scheduling strategy as it demonstrates a more aggressive decay trend and achieves the best performance in the ablation studies of Table~\ref{tab: abl_schedule}. The scheduling of $b_t$ is displayed in Figure~\ref{fig: budget_evolution}(a). In practice, to enhance training stability, we compute the importance score at each optimization step while reducing the budget and executing Algorithm~\ref{alg:alg1} at every epoch, as shown in Figure~\ref{fig: budget_evolution}(b).

\noindent\textbf{Weight-level Competition}. 
After allocating the budget to each layer via layer-level competition, weight-level competition is conducted to locate the most task-relevant weights within each layer.
Within the same layer, the importance of weights is reflected in the magnitude of their updates during the end-to-end optimization. 
Weights contributing more to the loss reduction are encouraged to have more significant updates during optimization, while weights contributing less approach zero. 
Specifically, given a merged matrix $\Delta \mathbf{W}\in\mathbb{R}^{m\times n}$ and the allocated number of tunable weights $b\in[0, mn]$~(with the optimization step $t$ and layer index $l$ omitted for brevity), we sparsify weights with a soft-threshold function. 
To induce weight competition during end-to-end fine-tuning, we propose a dynamic threshold $\Delta w_{b}$, \textit{i.e.}, the weight having the $b$-th largest magnitude in $\Delta \mathbf{W}$. This threshold ensures that only a fixed number of weights remain non-zero. Therefore, the weights have to compete with each other to be selected instead of just having a larger magnitude than a fixed threshold. 
\begin{equation}
    \Delta \mathbf{W}_{ij}^b = \Delta \mathbf{W}_{ij}\max(\vert\Delta \mathbf{W}_{ij}\vert-\vert\Delta w_b\vert, 0),
\end{equation}
where $\Delta \mathbf{W}^b=(\Delta \mathbf{W}_{ij}^{b})_{m\times n}$ denotes the sparse matrix with $b$ tunable weights.
Given a number of tunable weights $b$, the training objective in Equation~\ref{eq: training_objective_lora} can be reformulated as
\begin{equation}
\label{eq: snella_objective}
    \min_{\mathbf{A}, \mathbf{B}} \frac{1}{N} \sum_{n=1}^N \mathcal{L}\left(f(x^{(n)} ; \mathbf{W}_0+\Delta \mathbf{W}^b), y^{(n)}\right).
\end{equation}
This objective encourages weights that are most relevant to the downstream task to achieve more significant values for survival in the fine-tuning process.

%% file: tab/tab_fgvc_vtab1k.tex
\begin{table*}[]
\centering
\caption{Top-1 accuracy (\%) on FGVC and VTAB-1k benchmarks using ViT-B/16 pre-trained on ImageNet-21k supervisedly. The best result is in \textbf{bold}, and the second-best result is {\ul underlined}. $^*$For methods with different experimental settings, we utilize the re-implemented results under the settings of SNELLA to ensure a fair comparison.}
\setlength\tabcolsep{4pt}
\scalebox{0.95}{\begin{tabular}{@{}ccccccccccc@{}}
\toprule
\multicolumn{1}{l|}{\multirow{2}{*}{Method}} & \multicolumn{6}{c|}{FGVC}                                                                           & \multicolumn{4}{c}{VTAB-1k}                                   \\ \cmidrule(l){2-11} 
\multicolumn{1}{c|}{}                        & CUB-200 & NABirds & \begin{tabular}[c]{@{}c@{}}Oxford\\ Flowers\end{tabular} & \begin{tabular}[c]{@{}c@{}}Stanford\\ Dogs\end{tabular} & \begin{tabular}[c]{@{}c@{}}Stanford\\ Cars\end{tabular} & \multicolumn{1}{c|}{Mean Acc.} & Natural       & Specialized   & Structured    & Mean Acc.     \\ \midrule
\multicolumn{1}{l|}{Full}                    & 87.3    & 82.7    & 98.8           & 89.4          & 84.5          & \multicolumn{1}{c|}{88.5}      & 75.9          & 83.4          & 47.6          & 69.0          \\ \midrule
\multicolumn{11}{c}{Additional-based methods}                                                                                                                                                                      \\ \midrule
\multicolumn{1}{l|}{MLP-3~\cite{jia2022visual}}                   & 85.1    & 77.3    & 97.9           & 84.9          & 53.8          & \multicolumn{1}{c|}{79.8}      & 67.8          & 72.8          & 30.6          & 57.1          \\
\multicolumn{1}{l|}{VPT-Shallow~\cite{jia2022visual}}             & 86.7    & 78.8    & 98.4           & 90.7          & 68.7          & \multicolumn{1}{c|}{84.6}      & 76.8          & 79.7          & 47.0          & 67.8          \\
\multicolumn{1}{l|}{VPT-Deep~\cite{jia2022visual}}                & 88.5    & 84.2    & 99.0           & 90.2          & 83.6          & \multicolumn{1}{c|}{89.1}      & 78.5          & 82.4          & 55.0          & 72.0          \\
\multicolumn{1}{l|}{Adapter-8~\cite{houlsby2019parameter}}               & 87.3    & 84.3    & 98.4           & 88.8          & 68.4          & \multicolumn{1}{c|}{85.5}      & 79.0          & 84.1          & 58.5          & 73.9          \\
\multicolumn{1}{l|}{Adapter-32~\cite{houlsby2019parameter}}              & 87.2    & 84.3    & 98.5           & 89.6          & 68.4          & \multicolumn{1}{c|}{85.6}      & 79.6          & 84.0          & 58.3          & 74.0          \\
\multicolumn{1}{l|}{SPT-Adapter~\cite{he2023sensitivity}}             & 89.1    & 83.3    & 99.2           & 91.1          & 86.2          & \multicolumn{1}{c|}{89.8}      & 82.0          & 85.8          & 61.4          & 76.4\\
\multicolumn{1}{l|}{MoSA~\cite{zhang2023mosa}}             & 89.3    & 85.7    & 99.2           & 91.9          & 83.4          & \multicolumn{1}{c|}{89.9}      & 79.9          & 84.0          & 50.3         & 71.4\\ \midrule
\multicolumn{11}{c}{Reparameter-based methods}                                                                                                                                                                     \\ \midrule
\multicolumn{1}{l|}{Linear~\cite{jia2022visual}}                  & 85.3    & 75.9    & 97.9           & 86.2          & 51.3          & \multicolumn{1}{c|}{79.3}      & 68.9          & 77.2          & 26.8          & 57.6          \\
\multicolumn{1}{l|}{Partial-1~\cite{jia2022visual}}               & 85.6    & 77.8    & 98.2           & 85.5          & 66.2          & \multicolumn{1}{c|}{82.6}      & 69.4          & 78.5          & 34.2          & 60.7          \\
\multicolumn{1}{l|}{Bias~\cite{zaken2021bitfit}}                    & 88.4    & 84.2    & 98.8           & 91.2          & 79.4          & \multicolumn{1}{c|}{88.4}      & 73.3          & 78.3          & 44.1          & 65.2          \\
\multicolumn{1}{l|}{LoRA-8~\cite{hu2021lora}}                  & 84.9    & 79.0    & 98.1           & 88.1          & 79.8          & \multicolumn{1}{c|}{86.0}      & 79.5          & 84.6          & 60.5          & 74.9          \\
\multicolumn{1}{l|}{LoRA-16~\cite{hu2021lora}}                 & 85.6    & 79.8    & 98.9           & 87.6          & 72.0          & \multicolumn{1}{c|}{84.8}      & 79.8          & 84.9          & 60.2          & 75.0          \\
\multicolumn{1}{l|}{SPT-LoRA~\cite{he2023sensitivity}}                & 88.6    & 83.4    & {\ul 99.5}           & 91.4          & 87.3          & \multicolumn{1}{c|}{90.1}      & 81.9          & 85.9    & 61.3          & 76.4          \\
\multicolumn{1}{l|}{SSF~\cite{lian2022scaling}}                & 89.5    & 85.7    & \textbf{99.6}           & 89.6          & 89.2          & \multicolumn{1}{c|}{90.7}      & 81.6          & {\bf 86.6}    & 59.0          & 75.7          \\ 
\multicolumn{1}{l|}{GPS$^*$~\cite{DBLP:conf/cvpr/ZhangZGZSZZ24}}                & 89.7    & 86.4    & \textbf{99.7}           & 91.9          & 90.4          & \multicolumn{1}{c|}{91.6}      & 82.1          & {86.3}    & 61.3          & 76.6          \\ 
\midrule
\multicolumn{1}{l|}{SNELL-8~\cite{SNELL}}                 & 89.6    & {86.8}    & {99.3}           & {\ul 92.1}          & 89.9          & \multicolumn{1}{c|}{91.5}      & 82.0          & 85.7          & 61.6          & 76.4          \\
\multicolumn{1}{l|}{SNELL-16~\cite{SNELL}}                & {\ul 89.9}    & {\ul 87.0}    & 99.3           & {\bf 92.2}          & {90.3}          & \multicolumn{1}{c|}{{91.7}}      & {82.4}    & {86.1} & {61.7}    & {76.7}    \\
\multicolumn{1}{l|}{SNELL-32~\cite{SNELL}}                & {\ul 89.9}    & {\ul 87.0}    & 99.4           & 92.0          & 90.5          & \multicolumn{1}{c|}{{\ul 91.8}}          & {\ul 82.7} & 86.1 & 61.8 & 76.9 \\ \midrule
\multicolumn{1}{l|}{SNELLA-8~(\textbf{ours})}                & 89.8    & 86.9    & 99.4           & \textbf{92.2}          & 90.3          & \multicolumn{1}{c|}{91.7}          & 82.3 & 85.9 & 62.2 & 76.8 \\
\multicolumn{1}{l|}{SNELLA-16~(\textbf{ours})}                & {\ul 89.9}    & \textbf{87.2}    & 99.4           & \textbf{92.2}          & {\ul 90.6}          & \multicolumn{1}{c|}{\textbf{91.9}}          & {82.5} & 86.4 & \textbf{62.5} & {\ul 77.1} \\
\multicolumn{1}{l|}{SNELLA-32~(\textbf{ours})}                & \textbf{90.1}    & \textbf{87.2}    & \underline{99.5}           & \underline{92.1}          & \textbf{90.8}          & \multicolumn{1}{c|}{\textbf{91.9}}          & \textbf{82.8} & \underline{86.5} & \underline{62.3} & \textbf{77.2} \\
\bottomrule
\end{tabular}}
\label{tab: fgvc_vtab1k}
\vspace{-0.3cm}
\end{table*}

%% file: sec/4_experiments.tex
\section{Experiments}
\label{sec: experiments}

\input{tab/tab_mae_moco}
\input{tab/tab_swin_convnext}

\subsection{Implementation Details}

We evaluate SNELLA on multiple tasks, including image classification, segmentation, and generation.
\label{sec: implementation}

\noindent\textbf{Datasets and Metrics.} 
For \textit{classification}, we utilize 24 downstream tasks categorized into two groups.
(i) FGVC~\cite{jia2022visual} is a benchmark for fine-grained image classification. This benchmark includes 5 downstream tasks, which are CUB-200-2011~\cite{wah2011caltech}, NABirds~\cite{van2015building}, Oxford Flowers~\cite{nilsback2008automated}, Stanford Dogs~\cite{gebru2017fine} and Stanford Cars~\cite{dataset2011novel}. 
We follow the validation splits in \cite{he2023sensitivity} if the official validation set is unavailable.
(ii) VTAB-1k~\cite{zhai2019large} is a large-scale transfer learning benchmark consisting of 19 visual classification tasks. VTAB-1k can be further divided into three groups, {\it i.e.}, natural tasks with natural images, specialized tasks with images captured by specialized equipment, and structured tasks with images mostly generated from synthetic environments. 
We use top-1 accuracy averaged within each group as our main metric for evaluation following SPT~\cite{he2023sensitivity}.
For \textit{segmentation}, we adopt the polyp segmentation task from the Kvasir dataset~\cite{jha2020medico}. The evaluation is conducted using mDice and mIoU as metrics following GPS~\cite{DBLP:conf/cvpr/ZhangZGZSZZ24}.
For \textit{generation}, we employ concept customization tasks based on text-to-image generation models. Specifically, we fine-tune the model to associate a special token with a concept image from the Dreambooth~\cite{ruiz2023dreambooth} dataset. We select 30 distinct concepts and fine-tune the model using 4–6 images per concept. Following~\cite{ruiz2023dreambooth}, we use 25 diverse text prompts and generate 4 images for each prompt. All generated images are evaluated on the fidelity, diversity, and text-to-image alignment.

\noindent\textbf{Pre-trained Backbones.}
For classification tasks, we conduct experiments on the plain vision backbone ViT-B/16~\cite{dosovitskiy2020image} with different pre-training strategies following~\cite{he2023sensitivity}, including supervised pre-training on ImageNet-21k~\cite{ridnik2021imagenet} and self-supervised pre-training with MAE~\cite{he2022masked} and MoCo v3~\cite{chen2021empirical} on ImageNet-1k~\cite{russakovsky2015imagenet}. We also apply SNELLA on the representative hierarchical vision backbone Swin-B~\cite{liu2021swin} and CNN backbone ConvNeXt-Base~\cite{liu2022convnet} by supervised pre-training.
In addition, we fine-tune the supervised pre-trained large-scale models~(ViT-L/16~\cite{dosovitskiy2020image}, ViT-H/14~\cite{dosovitskiy2020image}) using SNELLA.
For segmentation tasks, we utilize the Segment Anything Model~(SAM)~\cite{kirillov2023segment} with ViT-B/16 as the vision backbone.
For text-to-image generation tasks, we use the medium version of Stable Diffusion 3~(SD3)~\cite{esser2024scaling}.

\noindent\textbf{Competitors.}
We compare SNELLA with addition-based methods including MLP-3, VPT-Shallow~\cite{jia2022visual}, VPT-Deep~\cite{jia2022visual}, Adapter-$r$~\cite{houlsby2019parameter}, MoSA~\cite{zhang2023mosa}, and SPT-Adapter~\cite{he2023sensitivity}.
For reparameterization-based methods, we compare with Linear, Partial-1, Bias~\cite{zaken2021bitfit}, LoRA-$r$~\cite{hu2021lora}, SSF~\cite{lian2022scaling}, SPT-LoRA~\cite{he2023sensitivity}, GPS~\cite{DBLP:conf/cvpr/ZhangZGZSZZ24}, and our previous work SNELL~\cite{SNELL}. 
Here $r$ represents the number of bottleneck dimensions in Adapter-$r$ and the value of rank in LoRA-$r$, SNELL-$r$, and our proposed SNELLA-$r$. 
Details of the competitors are presented in Appendix A.1. 

\begin{figure*}
  \centering
  \includegraphics[width=1.0\linewidth]{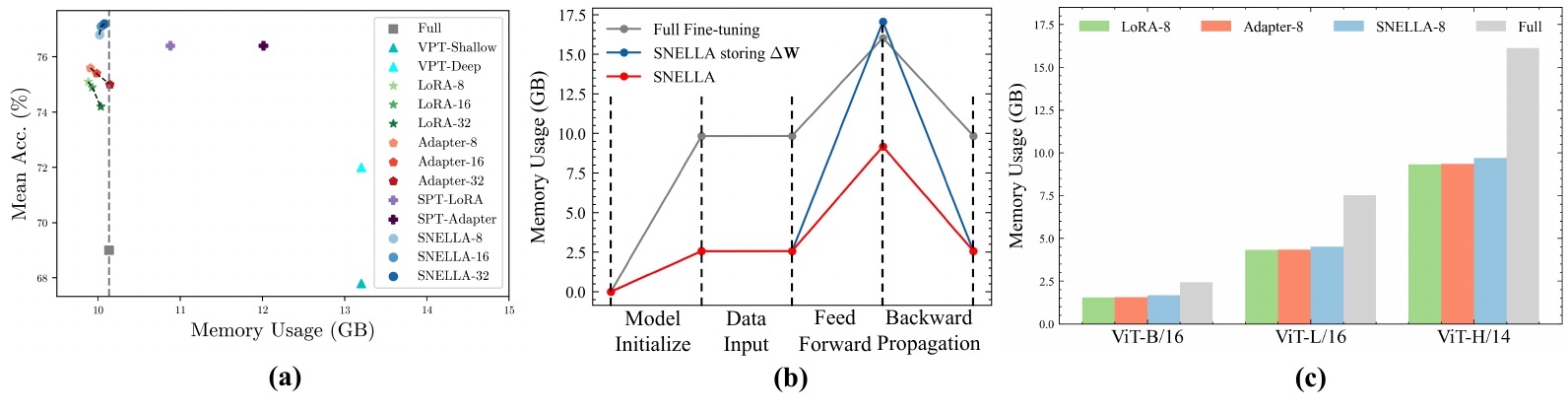}
  \vspace{-0.6cm}
  \caption{(a) Accuracy \textit{vs.} memory usage~(batchsize=64) with supervised pre-trained ViT-B/16 on VTAB-1k. 
  (b) Memory usage evolutions of full fine-tuning, SNELLA, and SNELLA storing the merged adaptation matrix~(SNELLA storing $\Delta \mathbf{W}$) on ViT-H/14 during the fine-tuning process~(batchsize=8).
  (c) Model parameter volumes \textit{vs.} memory usage~(batchsize=8). As the parameter scale of the pre-trained model increases, SNELLA's advantage of low memory usage over full fine-tuning becomes more obvious.}
  \label{fig: memory_analysis}
  \vspace{-0.3cm}
\end{figure*}
\input{tab/tab_vit_l_h}

\subsection{Performance on Classification Tasks}
\noindent\textbf{Performance on Different Benchmarks.}
Experiments on FGVC and VTAB-1k indicate that SNELLA achieves the best performance with a supervised pre-trained ViT-B/16 backbone, as shown in Table~\ref{tab: fgvc_vtab1k}.
SNELLA significantly outperforms LoRA variants, {\it e.g.}, SNELLA-8 surpasses LoRA-8 significantly by 5.7\% in terms of mean accuracy on the FGVC benchmark. Moreover, SNELLA outperforms our previous work SNELL by a clear margin of 0.3\%-0.4\% in terms of mean top-1 accuracy on the VTAB-1k benchmark.
This improvement can be attributed to the more expressive kernel function for merging low-rank matrices and the mechanism of locating task-relevant weights.

\noindent\textbf{Performance on Different Pre-training Strategies.}
Experiments on models pre-trained using different strategies are presented in Table~\ref{tab: mae_moco}. SNELLA outperforms our previous work SNELL on ViT-B/16 pre-trained with MAE~(72.8\% \textit{vs.} 71.8\%) and MoCo v3~(75.9\% \textit{vs.} 75.5\%) strategies.
SNELLA consistently outperforms other PEFT methods on every group of downstream datasets~(accuracy gains of 0.2\% to 1.3\%).
This demonstrates the general effectiveness of SNELLA under different pre-training strategies.

\noindent\textbf{Performance on Different Model Architectures.}
We apply SNELLA to the hierarchical vision transformer Swin-Base and the CNN-based architecture ConvNeXt-Base. Experimental results are presented in Table~\ref{tab: swin_convnext}. Results on Swin-Base show that SNELLA-8 outperforms state-of-the-art PEFT methods~(\textit{i.e.}, SPT-Adapter and SNELL-8) by 0.3\%. For ConvNeXt-Base, SNELLA achieves a performance gain of 0.4\% over the best-reported result of SNELL-8. These results across different architectures further confirm the generality and effectiveness of SNELLA.

\noindent\textbf{Memory Usage Comparison.}
We illustrate the effectiveness of SNELLA in terms of memory usage by comparing it with various PEFT methods. Note that compared to SNELL, SNELLA achieves consistent performance improvements with different ranks without additional memory usage. Therefore, we only compare existing methods with SNELLA for brevity.
Figure~\ref{fig: memory_analysis}(a) shows the accuracy and memory usage of different methods on ViT-B/16. Although some methods achieve satisfactory performance, their memory usage is excessively large, even surpassing that of full fine-tuning~(\textit{e.g.} SPT-Adapter and VPT-Deep).
In comparison, SNELLA achieves superior performance on downstream tasks with memory usage comparable to memory-efficient methods, including LoRA and Adapter. 

Additionally, we present the memory usage evolutions during the fine-tuning process in Figure~\ref{fig: memory_analysis}(b) to provide a detailed explanation of how SNELLA can save memory. 
In the model initialization stage, SNELLA exhibits a significantly smaller memory usage compared to full fine-tuning. This is because full fine-tuning stores all weight matrices as learnable parameters in the optimizer, whereas SNELLA only stores low-rank matrices with smaller parameter volumes.
In the feed-forward stage, the memory usage increases with the storage of intermediate variables for backpropagation. 
Unlike other intermediate variables, the adaptation matrix $\Delta \mathbf{W}$ in SNELLA solely relies on the low-rank matrices, which are already stored in the optimizer. Therefore, it can be dumped in the feed-forward phase and recovered in backpropagation immediately, saving a large amount of memory usage~(SNELLA \textit{v.s.} SNELLA storing $\Delta \mathbf{W}$).

\begin{figure*}
  \centering
  \includegraphics[width=1.0\linewidth]{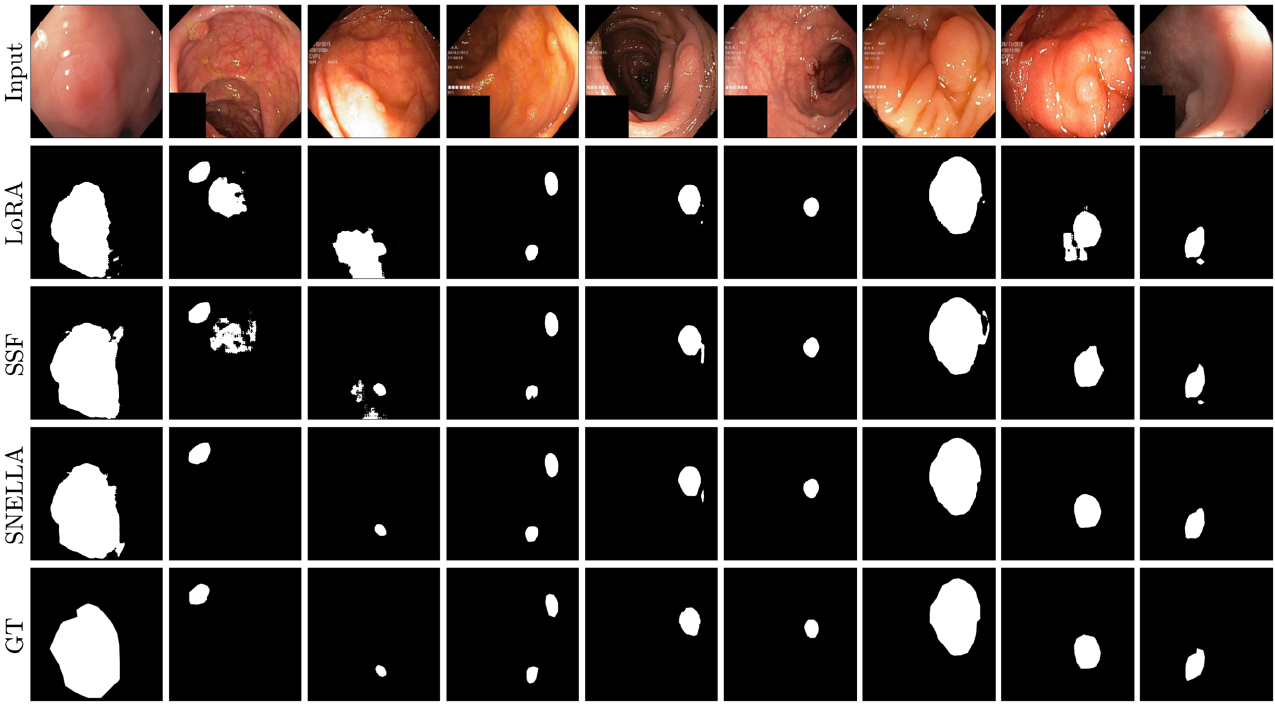}
  \vspace{-0.2cm}
  \caption{Visualizations of the polyp segmentation task. We provide the input image~(Input), ground-truth~(GT), and segmentation results of different PEFT methods~(LoRA, SSF, and SNELLA). Our SNELLA can effectively handle challenging segmentation cases compared to other methods.}
  \label{fig: SAM_vis}
  \vspace{-0.3cm}
\end{figure*}

\input{tab/tab_polyp}

\noindent\textbf{Scaling to Larger Models.} 
We apply SNELLA to ViT models of varying sizes~(ViT-B/16, ViT-L/16, and ViT-H/16 pre-trained on ImageNet-21K).
See Figure~\ref{fig: memory_analysis}(c), the memory usage of full fine-tuning increases rapidly as the model size grows. 
This observation highlights that existing PEFT methods~(\textit{e.g.}, VPT, SPT, and GPS), despite their advanced performances, incur substantial memory costs when applied to large-scale models, showing even higher memory usage than full fine-tuning.
In contrast, SNELLA demonstrates a significant advantage in memory efficiency on larger models~(comparable to LoRA-8). When applied to the large-scale model ViT-H/14, the memory usage of SNELLA is only approximately 50\% of that required for full fine-tuning.

Regarding the performance, as shown in Table~\ref{tab: vit_l_h}, SNELLA-8 outperforms LoRA-8, SSF, and SNELL-8 in the mean accuracy for both ViT-L and ViT-H on the VTAB-1k benchmark.
This demonstrates the effectiveness of SNELLA for adapting large pre-trained models to downstream tasks.

\input{tab/tab_generation}

\subsection{Performance on Dense Prediction Tasks}

\noindent\textbf{Polyp Segmentation Task}.
We apply SNELLA to the semantic segmentation task. Segment Anything Model (SAM)~\cite{kirillov2023segment} is a strong foundation model that enables powerful generalization capabilities. However, its performance is suboptimal on medical segmentation tasks, such as polyp segmentation~\cite{jha2020medico}. To address this limitation, previous studies~\cite{chen2023sam} proposed employing Adapter modules to fine-tune SAM for downstream medical segmentation tasks. Following their experimental setup~\cite{DBLP:conf/cvpr/ZhangZGZSZZ24}, we apply SNELLA to SAM and conduct a comparative analysis with other PEFT approaches.
Quantitative results are shown in Table~\ref{tab: polyp}. Compared to current methods, SNELLA demonstrates superior mDice~(0.9688 v.s. 0.9522) and mIoU~(0.9395 v.s. 0.9088) scores with low memory usage~(10874M v.s. 12488M). Figure~\ref{fig: SAM_vis} also presents a qualitative comparison of different methods. The results of SNELLA are most consistent with the ground truth. SNELLA achieves more precise segmentation of target objects, while the predictions of other approaches often include falsely-alarmed and irregularly shaped regions with indistinct boundaries.

\input{tab/tab_abl_kernel}
\input{tab/tab_abl_kernel_func}

\begin{figure*}
  \centering
  \includegraphics[width=1.0\linewidth]{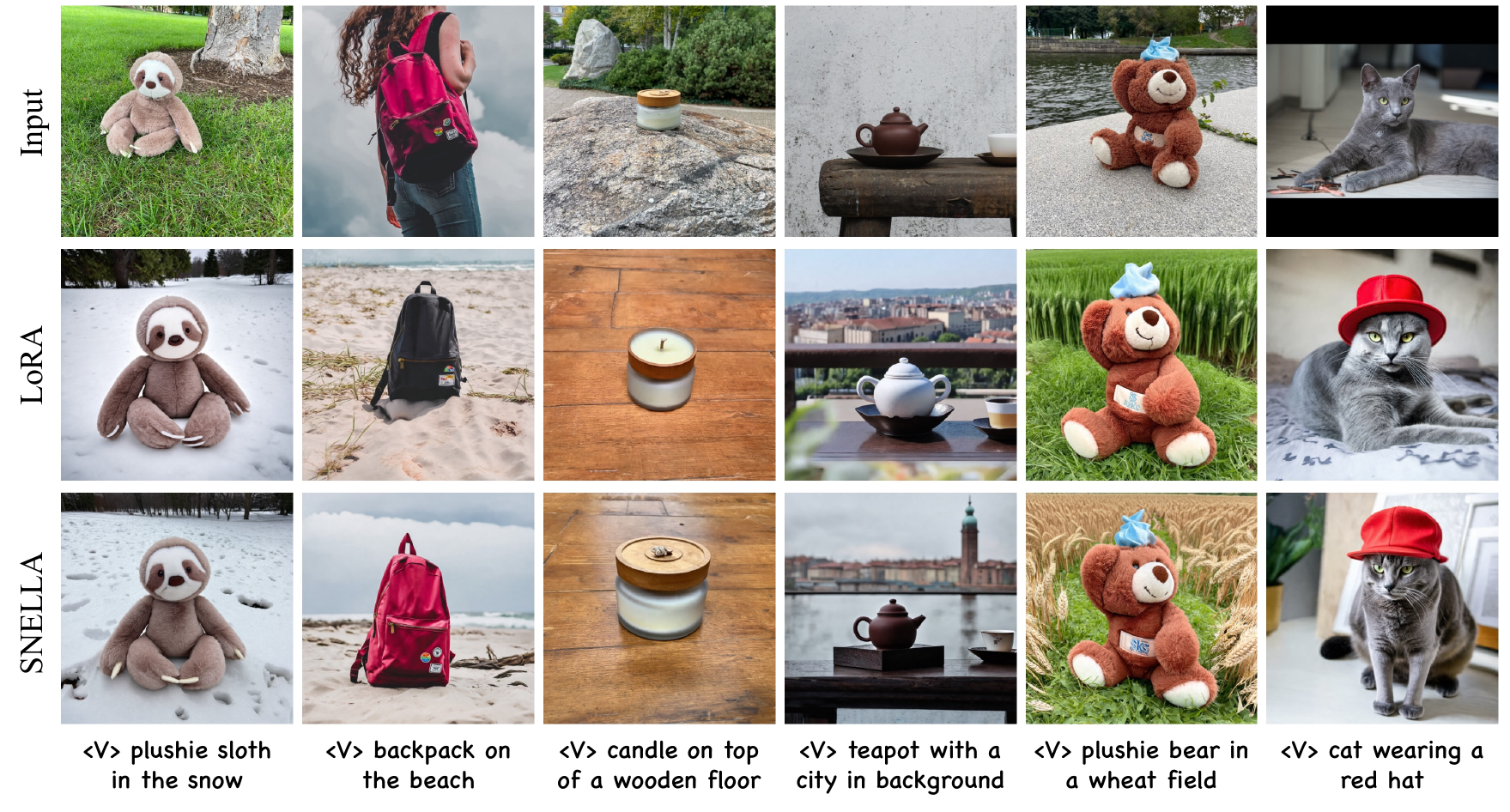}
  \vspace{-0.5cm}
  \caption{Qualitative results on concept customization for personalized text-to-image generation. Compared to the baseline method LoRA~(middle row), SNELLA~(bottom row) achieves higher concept fidelity and text alignment of the generated images. }
  \label{fig: generation}
  \vspace{-0.3cm}
\end{figure*}

\noindent\textbf{Personalized Text-to-Image Generation Task}. 
We provide the quantitative comparisons between SNELLA and other methods in (1) the concept alignment with the average cosine similarity between CLIP embeddings~\cite{radford2021learning} of the generated images and the concept images, (2) the diversity of generated images measured by the Vendi score~\cite{friedman2022vendi} calculated with the DINOv2 embeddings~\cite{oquab2023dinov2}, and (3) the text alignment of the generated images measured via average cosine similarity in the CLIP embeddings. 
As Table~\ref{tab: generation} shows, SNELLA achieves a higher alignment score and diversity than existing methods.
The qualitative comparison with LoRA is presented in Figure~\ref{fig: generation}. The images generated using SNELLA exhibit superior concept fidelity, preserving more distinctive features from the input images. For example, consider the concept ``$<$V$>$ backpack" as shown in the 2nd column in Figure~\ref{fig: generation}, the image generated by our method retains the backpack's original color from the input, whereas the LoRA-generated image alters the color. Moreover, the sand surrounding the backpack in the image of SNELLA exhibits a more realistic shape. This is achieved by SNELLA's ability to selectively adjust task-relevant weights while leaving irrelevant weights unaffected, thereby preserving the model's original image generation ability.

\subsection{Ablation Studies}
\noindent\textbf{Effects of Kernelized LoRA}.
We provide ablation studies on the nonlinear kernel function in Table~\ref{tab: abl_kernel}. First, compared with the linear kernel, the non-linear kernel~(\textit{e.g.}, P-Linear) achieves significant performance improvement~(75.2\% \textit{v.s.} 74.2\%). Particularly, when combined with a sparsification mechanism, the linear kernel suffers from performance degradation~(74.2\% \textit{v.s.} 65.7\%), whereas non-linear kernels exhibit enhanced performance~(75.2\% \textit{v.s.} 76.9\%). This underscores the importance of integrating non-linear kernels in sparse fine-tuning. Second, the proposed Mix-K function demonstrates superior performance compared to the P-Linear kernel, both with~(77.2\% \textit{v.s.} 76.9\%) and without~(75.7\% \textit{v.s.} 75.2\%) sparsification. This advantage arises from the enhanced expressive capability of Mix-K, which combines the P-Linear and exponential kernels.

\input{tab/tab_abl_sparse}
\input{tab/tab_abl_importance}

\noindent\textbf{Effects of Mix-K}.
In Table~\ref{tab: abl_kernel_func}, we present the performance of kernelized LoRA using different exponential kernel functions. First, existing kernels with exponential operators (\textit{e.g.}, Sigmoid and RBF) achieve poor performance due to gradient vanishing, as illustrated in Figure~\ref{fig: kernel_design}(c). Second, normalizing the kernel function values can significantly improve the performance of RBF kernels~(75.2\% \textit{v.s.} 58.7\%). Mix-K achieves better performance than the normalized RBF kernel~(75.7\% \textit{v.s.} 75.2\%), indicating its enhanced expressivity.

\noindent\textbf{Effects of Sparsification Mechanism}.
In Table~\ref{tab: abl_saprse}, we compare the performance of SNELLA with and without sparsification mechanisms. First, incorporating sparsity through competitions within layers significantly improves the performance on downstream tasks compared to relying solely on kernelized LoRA~(77.0\% \textit{v.s.} 75.7\%). This indicates that fine-tuning only the task-relevant weights improves the generalization capability of models across downstream tasks. 
Second, competitions across layers further improve model performance~(77.2\% \textit{v.s.} 77.0\%). This is because different layers demonstrate varying degrees of relevance to the downstream task. By allocating layer-specific sparsity based on the task, task-relevant weights are located more accurately, thereby enhancing fine-tuning performance. 

\input{tab/tab_abl_budget}
\input{tab/tab_abl_schedule}

\noindent\textbf{Effects of Importance Score}.
We investigate the impact of layer importance scores on model performance by substituting sensitivity in Equation~\ref{eq: sensitivity} with alternative metrics, including the magnitude of low-rank matrices $\mathbf{A}$ and $\mathbf{B}$~(Magnitude) and the magnitude of the merged sparse matrix $\Delta \mathbf{W}$~(W-Magnitude). As shown in Table~\ref{tab: abl_importance}, sensitivity achieves the best performance under different allocation strategies. Moreover, reallocating sparsity across layers per epoch achieves better performance than reallocating at each training step, because the importance score calculated using the entire dataset offers a more accurate estimation of the relevance between layers and downstream tasks.

\noindent\textbf{Effects of the Number of Tunable Weights}.
Table~\ref{tab: abl_budget} presents the performance of our method across four downstream datasets (Sun397, Camelyon, Clevr-count, and Snorb-azi) under varying $b_T$. 
First, no single value can maximize performance across all datasets, since each downstream task demonstrates different preferences for volume of nonzero weights. For example, the best result on the Sun397 dataset is achieved with $b_T/b_0 = 0.10$, whereas the best performance on Clevr-count is obtained with $b_T/b_0 = 0.90$. Second, the mean accuracy across datasets follows a monotonic trend, with the highest average of 63.92\% achieved at $b_T/b_0=0.30$. Beyond $b_T/b_0=0.30$, mean accuracy gradually decreases, with a significant drop at $b_T/b_0=0.99$~(62.88\%), highlighting the importance of determining the number of tunable weights.

\noindent\textbf{Effects of Scheduling Strategy}.
We investigate the effectiveness of our scheduling strategy in Equation~\ref{eq: budget_scheduling}. Specifically, we conduct experiments by setting the number of tunable weights $b_t$ to be a constant $b$, and decaying from $b_0$ to $b$ in linear, quadratic, and cubic manners in Table~\ref{tab: abl_schedule}. 
The accuracy is relatively low without budget scheduling~(61.07\% \textit{v.s.} 63.50\%). This is because a large number of weights are zeroed out at the early stage of fine-tuning, preventing these weights from being updated. The insufficient training hinders the end-to-end locating of task-relevant weights.
Second, the cubic strategy achieves the best performance~(63.88\%), followed by the quadratic strategy~(63.68\%), while the linear strategy yields the worst result~(63.50\%). 
As the power of the decay function increases, the number of tunable weights decreases in a pattern that is initially steep and subsequently smoother, ensuring greater optimization stability during the later stages of fine-tuning.


\begin{figure*}
  \centering
  \includegraphics[width=1.0\linewidth]{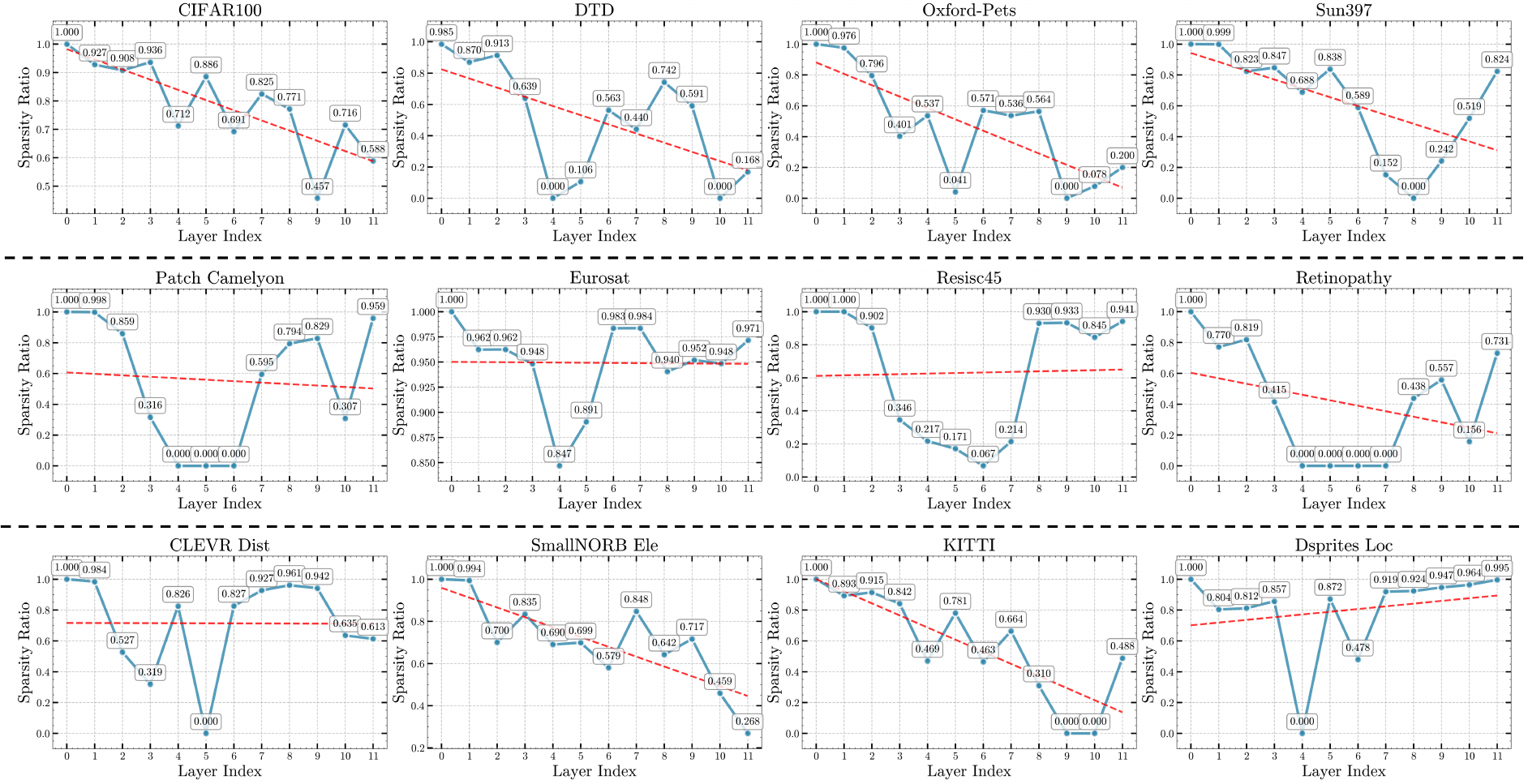}
  \vspace{-0.5cm}
  \caption{Visualization of the allocated sparsity ratios~(the ratio of zero-valued weight updates relative to the total number of weights) across different layers. We present the results of a ViT-B/16 model evaluated on multiple datasets of different groups in the VTAB-1k benchmark: (\textit{top}: Natural, \textit{middle}: Specialized, \textit{bottom}: Structured). The red line represents the linear approximation of the sparsity ratio distribution.}
  \label{fig: sparsity_pattern}
  \vspace{-0.3cm}
\end{figure*}

\begin{figure*}
  \centering
  \includegraphics[width=0.97\linewidth]{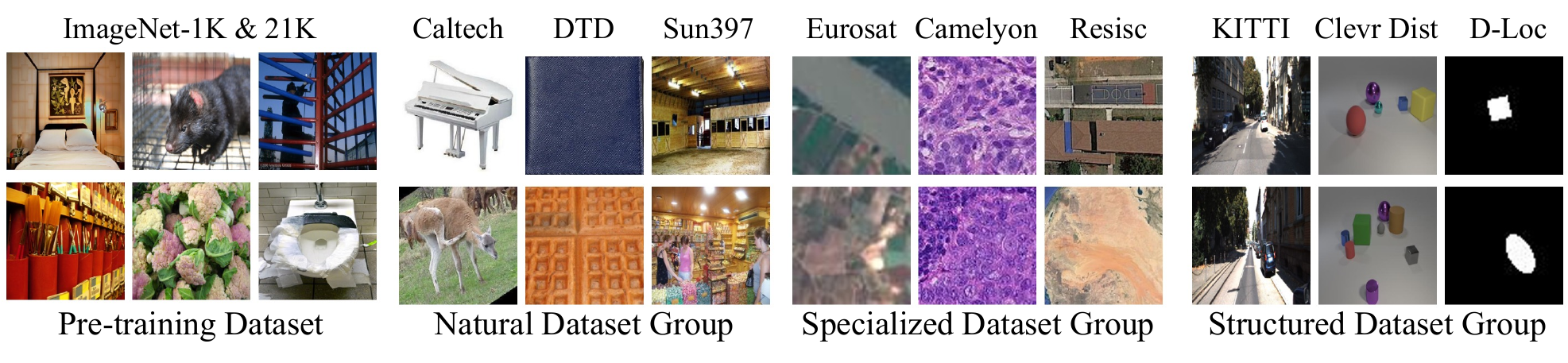}
  \vspace{-0.5cm}
  \caption{Visualization of pre-training dataset of ViT-B/16 model and the dataset groups~(Natural, Specialized, and Sturctured) in VTAB-1k benchmark. }
  \label{fig: dataset_visualization}
  \vspace{-0.3cm}
\end{figure*}

\subsection{Additional Analysis}

\noindent\textbf{Layer-level Competition Results Across Tasks}. We visualize the layer-level competition results across tasks in VTAB-1k in Figure~\ref{fig: sparsity_pattern}. For better analysis, examples of the model's pre-training dataset and different dataset groups from VTAB-1k are presented in Figure~\ref{fig: dataset_visualization}. Different types of datasets lead to distinct competition results. 

The Natural datasets, comprising natural objects and scenes, exhibit distributions similar to those of the model's pre-training dataset. Our method tends to allocate lower sparsity ratios ({\it i.e.}, more tunable weights) to the top layers rather than the bottom layers. The allocation result indicates that only the model's high-level reasoning mechanisms need to be adjusted, while its low-level image processing capabilities remain largely unchanged.
The Specialized datasets exhibit distributional differences compared to the pre-training dataset, \textit{e.g.}, cell slices and remote sensing images. Adapting pre-trained models to these datasets necessitates more fundamental fine-tuning of the model. Therefore, the middle layers rather than the top layers are allocated with more nonzero weight updates. 
For structured datasets, their allocated sparsity patterns exhibit great diversity. For example, in KITTI with similar distributions to pre-training datasets, our method tends to allocate more weight updates to top layers, whereas in CLEVR Dist and Dsprites Loc, more weight updates are allocated to middle layers. 

Based on the above analysis, we can conclude that the sparsity allocation is influenced by {\it the distribution difference between the downstream and pre-training datasets}. When the distribution difference is small, the model tends to adapt the top layers. In contrast, when the distribution difference is large, the model preferentially adjusts the middle layers. 
Bottom layers are typically allocated with a high sparsity ratio, indicating that only a small number of weights need to be adjusted in these layers. The majority of weight adjustments are concentrated in the middle and top layers for better adaptation to downstream tasks.

\begin{figure}
  \centering
  \includegraphics[width=0.96\linewidth]{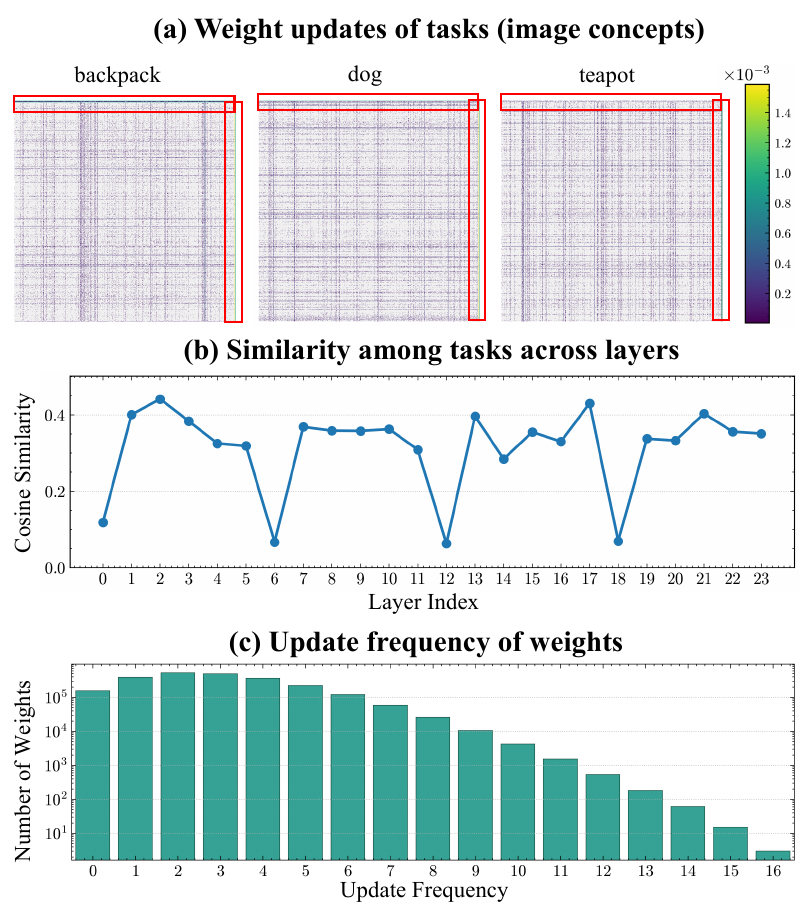}
  \vspace{-0.4cm}
  \caption{(a) Visualization examples of weight updates in the 23-th layer's Q-matrix on 3 personalized text-to-image generation tasks out of 30. Red box regions indicate the weights significantly updated by multiple tasks. (b) Similarity of weight updates among tasks at different layers. (c) Update frequency in the 23-th layer's Q-matrix across all tasks. Experiments are conducted with SD3 on the DreamBooth dataset.}
  \label{fig: sparsity_pattern_generation}
  \vspace{-0.5cm}
\end{figure}

\noindent\textbf{Weight-level Competition Results Across Tasks}. For the weight-level competition, we analyze the locating results from 30 tasks in Dreambooth~\cite{ruiz2023dreambooth} based on SD3. Each task corresponds to a specific concept that exhibits minimal correlation with other tasks, enabling a clearer evaluation of whether the locating results are task-relevant.
In Figure~\ref{fig: sparsity_pattern_generation}(a), we visualize the weight update $\mathbf{\Delta W}$ of 3 tasks as an example. 
The model encodes information of distinct tasks into separate weight locations as shown by the distinguished patterns in the rows and columns of the weight matrix, exhibiting task-specific behaviors clearly. Beyond this observation, we find that certain weights are unanimously selected across all three tasks, such as the first row and last column of the weight matrix marked by red boxes. This indicates that there exist some weights that deliver the shared information among different downstream tasks. 
We further quantitatively examine the above observations.
In Figure~\ref{fig: sparsity_pattern_generation}(b), we compute the average cosine similarity of weight update $\mathbf{\Delta W}$ across all tasks to quantify the task relevance\footnote{According to~\cite{ding2025stable,ding2025dis2booth}, the task relevance can be interpreted as the similarity between concept-specific features, {\it i.e.}, their common features.} of the locating/updating results at different layers~(\textit{i.e.}, layer with higher similarity indicates lower task relevance). The similarity remains consistently low~($<$0.5) for all layers, indicating the high task relevance of the locating results. Furthermore, the task relevance is particularly pronounced in certain layers~(\textit{e.g.}, the 6-th, 12-th, and 18-th layers), indicating minimal overlap between locating results of different tasks. These layers tend to employ distinct weights to learn new concepts and serve as critical layers for storing concept-specific information.
In Figure~\ref{fig: sparsity_pattern_generation}(c), we show the update frequency of weights across 30 tasks. Most weights are selected by a limited number of tasks~(0–3), while a small subset of weights is frequently selected by up to 16 tasks. The frequently selected weights are correlated with the shared information across tasks, confirming our observation in Figure~\ref{fig: sparsity_pattern_generation}(a).

\begin{figure}
  \centering
  \includegraphics[width=0.95\linewidth]{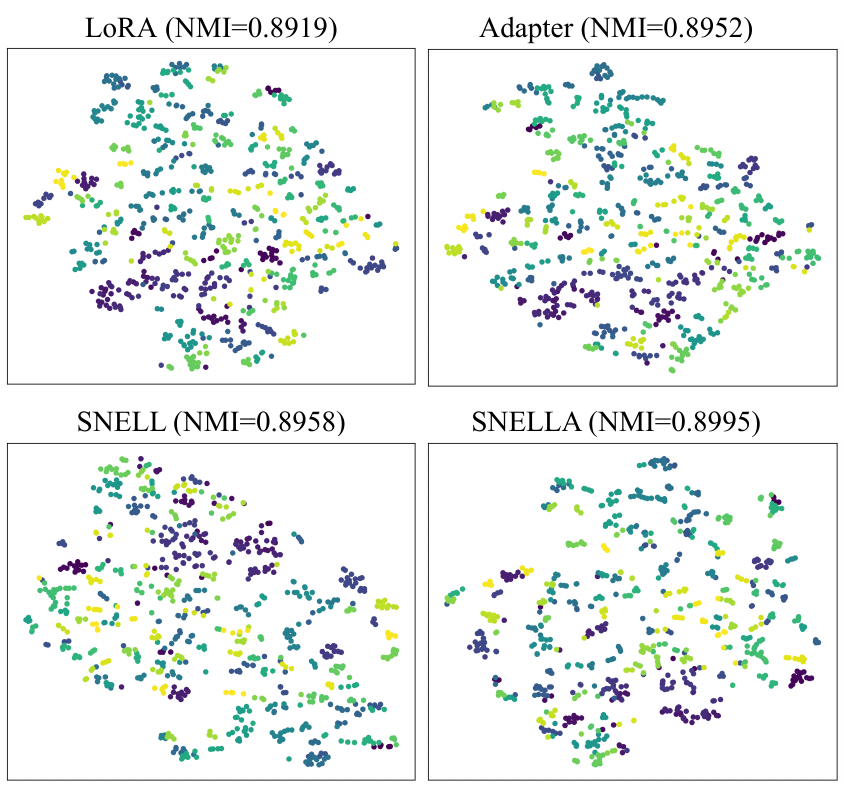}
  \vspace{-0.2cm}
  \caption{T-SNE visualizations of different fine-tuning methods. Experiments are conducted on the Stanford-Car dataset from the FGVC benchmark based on the ViT-B/16 pre-trained on ImageNet-21k.}
  \label{fig: tsne}
  \vspace{-0.3cm}
\end{figure}

\noindent\textbf{T-SNE Visualizations}.
On the Stanford-Car dataset, we use T-SNE to visualize the feature distribution of different fine-tuning methods. The visualization results are illustrated in Figure~\ref{fig: tsne}. Feature clustering results using the proposed SNELLA are superior to those with LoRA, Adapter, and SNELL. Specifically, SNELLA features within the same class exhibit a more compact clustering pattern, while the separation between different classes becomes more pronounced. This phenomenon can be quantitatively demonstrated through the higher Normalized Mutual Information~(NMI) score~\cite{estevez2009normalized} achieved by SNELLA.

\subsection{Discussion}

\noindent\textbf{Potential to Approximate Full Fine-tuning}.
While SNELLA outperforms existing methods on commonly used fine-tuning evaluation tasks, it shows potential for extension to large-scale tasks such as instruction tuning~\cite{NEURIPS2023_6dcf277e} and VLM construction~\cite{li2024llavanextinterleavetacklingmultiimagevideo}. In such scenarios, full fine-tuning is widely adopted due to the limited expressivity of current PEFT methods. Our proposed kernelized LoRA can enhance the expressivity of learnable modules using nonlinear kernel functions while maintaining memory efficiency, which has the potential to overcome the performance limitations of existing PEFT methods in large-scale tasks. However, realizing this potential presents a significant challenge, as it requires balancing the time cost introduced by nonlinear operations against the gains in expressivity. We leave this challenge for future investigation, with the ultimate goal of enabling PEFT to achieve performance comparable to full fine-tuning.

\noindent\textbf{Evaluation Metrics}.
We clarify the reason for not comparing SNELLA with other methods in terms of the volume of learnable parameters.
First, computing the volume of learnable parameters in SNELLA is difficult.
In the case of LoRA, the volume corresponds to the size of the learnable low-rank matrices. For sparse tuning, the volume is determined by the number of updated weights.
However, SNELLA employs low-rank matrices as learnable parameters and achieves additional updated weight reduction by sparsifying the merged matrices.
When using the parameter volume computation method of LoRA, calculating the reduction in parameters due to sparsification becomes challenging.
Conversely, applying the computation method of sparse tuning would be inherently unfair, given that SNELLA is specifically optimized using low-rank learnable matrices.
Second, the parameter efficiency is a pathway to achieve high performance and low memory usage rather than an objective for model improvement, because performance improvement and memory usage reduction hold practical value.
In experiments, SNELLA has demonstrated its advantages in high performance and low memory usage, which we consider more valuable than the pursuit of fewer learnable parameters in practice.

\noindent\textbf{Limitation}.
Although SNELLA achieves state-of-the-art performance with low memory consumption, the incorporation of non-linear kernel functions necessitates the recomputation of the merged weight matrix $\Delta \mathbf{W}$ during backpropagation, thereby increasing training time compared to linear methods such as LoRA. 
Furthermore, the frequency of sparsity allocation is determined by simple rules (e.g., fixed intervals such as every epoch or step). Excessively frequent sparsity allocation may result in insufficient training of model parameters. 
In future work, we aim to address these two limitations concurrently. By integrating a hybrid mechanism that combines expressive nonlinear methods for end-to-end sparse tuning with efficient linear methods for stable parameter updates, we can reduce training time while ensuring more thorough parameter training.


%% file: tab/tab_mae_moco.tex
\begin{table*}[]
\centering
\setlength\tabcolsep{8pt}
\caption{Top-1 accuracy (\%) on VTAB-1k using ViT-B/16 backbone pre-trained on ImageNet using MAE and MoCo v3 strategies. The best result is in \textbf{bold}.}
\scalebox{1.0}{
\begin{tabular}{@{}l|cccc|cccc@{}}
\toprule
\multirow{2}{*}{Methods} & \multicolumn{4}{c|}{VTAB-1k MAE}                & \multicolumn{4}{c}{VTAB-1k MoCo v3}            \\ \cmidrule(l){2-9} 
                         & Natural & Specialized & Structured & Mean Acc. & Natural & Specialized & Structured & Mean Acc. \\ \midrule
Full                     & 59.3    & 79.7        & 53.8       & 64.3      & 72.0    & 84.7        & 42.0       & 69.6      \\ \midrule
\multicolumn{9}{c}{Additional-based methods}                                                                               \\ \midrule
Adapter-8~\cite{houlsby2019parameter}                & 57.2    & 78.4        & 54.7       & 63.4      & 27.6    & 70.9        & 48.4       & 49.0      \\
Adapter-32~\cite{houlsby2019parameter}               & 55.3    & 78.8        & 53.3       & 62.5      & 74.2    & 82.7        & 47.7       & 68.2      \\
VPT-Shallow~\cite{jia2022visual}              & 40.0    & 69.7        & 27.5       & 45.7      & 67.3    & 82.3        & 37.6       & 62.4      \\
VPT-Deep~\cite{jia2022visual}                 & 36.0    & 60.6        & 26.6       & 41.1      & 70.3    & 83.0        & 42.4       & 65.2      \\
SPT-Adapter~\cite{he2023sensitivity}              & 65.6    & 82.7        & 60.7       & 69.7      & 76.6    & 85.0        & 61.7       & 74.4      \\ \midrule
\multicolumn{9}{c}{Reparameterization-based methods}                                                                       \\ \midrule
Linear~\cite{jia2022visual}                   & 18.9    & 52.7        & 23.7       & 32.1      & 67.5    & 81.1        & 30.3       & 59.6      \\
Partial-1~\cite{jia2022visual}                & 58.4    & 78.3        & 47.6       & 61.5      & 72.3    & 84.6        & 47.9       & 68.3      \\
Bias~\cite{zaken2021bitfit}                     & 54.6    & 75.7        & 47.7       & 59.3      & 72.9    & 81.1        & 53.4       & 69.2      \\
LoRA-8~\cite{hu2021lora}                   & 57.5    & 77.7        & 57.7       & 64.3      & 21.2    & 66.7        & 45.1       & 44.3      \\
LoRA-16~\cite{hu2021lora}                  & 57.3    & 77.1        & 59.9       & 64.8      & 16.0    & 64.0        & 48.7       & 42.9      \\
SSF~\cite{lian2022scaling}                 & 65.1    & 81.9        & 61.8       & 69.6      & 75.9    & 84.5        & 60.9       & 73.8      \\
SPT-LoRA~\cite{he2023sensitivity}                 & 65.4    & 82.4        & 61.5       & 69.8      & 76.5    & 86.0        & 63.6       & 75.3      \\ \midrule
SNELL-8~\cite{SNELL}           & 68.3       & 83.8           & 63.5          & 71.8         & 76.8       & 86.0            & 63.7          & 75.5         \\
SNELLA-8~(\textbf{ours})           & \textbf{69.2}       & \textbf{84.4}           & \textbf{64.8}          & \textbf{72.8}         & \textbf{77.0}       & \textbf{86.3}            & \textbf{64.4}          & \textbf{75.9}         \\ \bottomrule
\end{tabular}}
\label{tab: mae_moco}
\vspace{-0.2cm}
\end{table*}

%% file: tab/tab_swin_convnext.tex
\begin{table*}[]
\centering
\setlength\tabcolsep{8pt}
\caption{Comparisons on VTAB-1k with supervised pre-trained Swin-B and ConvNeXt-B. Top-1 accuracy (\%) is reported. The best result is in \textbf{bold}.}
\scalebox{1.0}{
\begin{tabular}{@{}ccccccccc@{}}
\toprule
\multicolumn{1}{l|}{\multirow{2}{*}{Methods}} & \multicolumn{4}{c|}{VTAB-1k Swin-B}                                                & \multicolumn{4}{c}{VTAB-1k ConvNeXt-B}                          \\ \cmidrule(l){2-9} 
\multicolumn{1}{c|}{}                         & Natural       & Specialized   & Structured    & \multicolumn{1}{c|}{Mean Acc.}     & Natural       & Specialized   & Structured    & Mean Acc.     \\ \midrule
\multicolumn{1}{l|}{Full}                     & 79.1          & 86.2          & 59.7          & \multicolumn{1}{c|}{75.0}          & 78.0          & 83.7          & 60.4          & 74.0          \\ \midrule
\multicolumn{9}{c}{Additional-based methods}                                                                                                                                                       \\ \midrule
\multicolumn{1}{l|}{MLP-3~\cite{jia2022visual}}                    & 73.6          & 75.2          & 35.7          & \multicolumn{1}{c|}{61.5}          & 73.8          & 81.4          & 35.7          & 63.6          \\
\multicolumn{1}{l|}{VPT-Deep~\cite{jia2022visual}}                 & 76.8          & 84.5          & 53.4          & \multicolumn{1}{c|}{71.6}          & 78.5          & 83.0          & 44.6          & 68.7          \\
\multicolumn{1}{l|}{Adapter-8~\cite{houlsby2019parameter}}                & 81.7          & 87.3          & 61.2          & \multicolumn{1}{c|}{76.7}          & 83.1          & 84.9          & 64.6          & 77.5          \\
\multicolumn{1}{l|}{SPT-Adapter~\cite{he2023sensitivity}}              & 83.0          & 87.3          & 62.1 & \multicolumn{1}{c|}{77.5} & 83.7          & 86.2          & 65.3          & 78.4          \\ \midrule
\multicolumn{9}{c}{Reparameterization-based methods}                                                                                                                                               \\ \midrule
\multicolumn{1}{l|}{Linear~\cite{jia2022visual}}                   & 73.5          & 80.8          & 33.5          & \multicolumn{1}{c|}{62.6}          & 74.5          & 81.5          & 34.8          & 63.6          \\
\multicolumn{1}{l|}{Partial-1~\cite{jia2022visual}}                & 73.1          & 81.7          & 35.0          & \multicolumn{1}{c|}{63.3}          & 73.8          & 81.6          & 39.6          & 65.0          \\
\multicolumn{1}{l|}{LoRA-8~\cite{hu2021lora}}                   & 81.7          & 87.2          & 60.1          & \multicolumn{1}{c|}{76.3}          & 82.2          & 84.7          & 64.1          & 77.0          \\
\multicolumn{1}{l|}{SSF~\cite{lian2022scaling}}                 & 81.8          & 86.8          & 61.9          & \multicolumn{1}{c|}{76.8}          & 83.8          & 86.8          & 65.5 & 78.7          \\
\multicolumn{1}{l|}{SPT-LoRA~\cite{he2023sensitivity}}                 & 83.1          & 87.4          & 60.4          & \multicolumn{1}{c|}{77.2}          & 83.4          & 86.7          & 65.9 & 78.7          \\ \midrule
\multicolumn{1}{l|}{SNELL-8~\cite{SNELL}}                  & 83.3 & \textbf{87.7} & 61.4          & \multicolumn{1}{c|}{77.5} & 84.5 & \textbf{87.4} & 65.6          & 79.1 \\
\multicolumn{1}{l|}{SNELLA-8~(\textbf{ours})}                  & \textbf{83.8} & 86.8 & \textbf{62.8}          & \multicolumn{1}{c|}{\textbf{77.8}} & \textbf{85.1} & 87.2 & \textbf{66.1}          & \textbf{79.5} \\ \bottomrule
\end{tabular}}
\label{tab: swin_convnext}
\vspace{-0.3cm}
\end{table*}

%% file: tab/tab_vit_l_h.tex
\begin{table*}[]
\centering
\caption{Comparisons on VTAB-1k benchmark with supervised pre-trained ViT-L/16 and ViT-H/16. Top-1 accuracy is reported. The best result is in \textbf{bold}.}
\setlength\tabcolsep{8pt}
\scalebox{1.0}{\begin{tabular}{@{}l|cccc|cccc@{}}
\toprule
\multirow{2}{*}{Methods} & \multicolumn{4}{c|}{VTAB-1k ViT-L/16}                  & \multicolumn{4}{c}{VTAB-1k ViT-H/14}                   \\ \cmidrule(l){2-9} 
                         & Natural & Specialized & Structured & Mean Acc. & Natural & Specialized & Structured & Mean Acc. \\ \midrule
LoRA-8~\cite{hu2021lora}                   & 81.2    & 86.6        & 53.4       & 73.7      & 77.9    & 84.8        & 55.9       & 72.9      \\
SSF~\cite{lian2022scaling}                   & 81.0    & 86.4        & 58.6       & 75.3      & 80.3    & 84.6        & 55.9       & 73.6      \\
SNELL-8~\cite{SNELL}                  & 82.3    & 86.9        & 56.6       & 75.3      & 79.5    & 85.1           & 56.9          & 73.8         \\
SNELLA-8~\textbf{(ours)}                & 82.8    & 86.7        & 58.2       & \textbf{75.9}      & 79.8    & 85.4           & 58.9          & \textbf{74.7}      \\ \bottomrule
\end{tabular}}
\vspace{-0.4cm}
\label{tab: vit_l_h}
\end{table*}

%% file: tab/tab_polyp.tex
\begin{table}[]
\centering
\setlength\tabcolsep{4pt}
\caption{Quantitative Result for Polyp Segmentation.}
\begin{tabular}{@{}l|cc|cc@{}}
\toprule
Method  & mDice~($\uparrow$)           & mIoU~($\uparrow$)            & Memory~($\downarrow$)    & Params.~($\downarrow$) \\ \midrule
Full    & 0.9033          & 0.8237          & 12480M          & 93.8M   \\
Adapter-8 & 0.9259          & 0.8620          & 11642M          & 4.72M   \\
LoRA-8    & 0.9373          & 0.8820          & \textbf{10870M}          & 4.72M   \\
SSF     & 0.9420          & 0.8903          & 11740M          & 4.26M   \\
GPS     & 0.9522          & 0.9088          & 12488M          & \textbf{4.22M}   \\
SNELL-8 & 0.9536          & 0.9113          & 10874M          & 4.72M   \\
SNELLA-8~(\textbf{ours})  & \textbf{0.9688} & \textbf{0.9395} & {10874M} & 4.72M   \\ \bottomrule
\end{tabular}
\label{tab: polyp}
\vspace{-0.4cm}
\end{table}

%% file: tab/tab_generation.tex
\begin{table}[]
\centering
\caption{Qualitative results on concept customization for personalized text-to-image generation. }
\begin{tabular}{@{}c|ccc@{}}
\toprule
Method       & Concept Align. & Diversity     & Text Align.   \\ \midrule
LoRA   & 0.8325$\pm$0.0710  & 4.4687$\pm$2.1526 & 0.2612$\pm$0.0285 \\
SSF   & 0.8453$\pm$0.0913  & 5.1416$\pm$2.2070 & 0.2634$\pm$0.0224 \\
SNELL  & 0.8517$\pm$0.0546 & 4.4682$\pm$2.3871 & 0.2671$\pm$0.0265 \\
SNELLA & \textbf{0.8599}$\pm$0.0492  & \textbf{5.8715}$\pm$2.7406 & \textbf{0.2702}$\pm$0.0252 \\ \bottomrule
\end{tabular}
\label{tab: generation}
\end{table}

%% file: tab/tab_abl_kernel.tex
\begin{table}[]
\centering
\caption{Ablation studies on kernelized LoRA. We utilize LoRA with linear and non-linear kernels for adapting the pre-trained ViT-B/16 model to the VTAB-1k benchmark. Performances with and without the sparsification mechanism are provided.}
\setlength\tabcolsep{2pt}
\begin{tabular}{@{}c|c|ccc|c@{}}
\toprule
                              & Method   & Natural & Specialized & Sturctured & Mean Acc. \\ \midrule
\multirow{3}{*}{w/o sparsify} & Linear   & 79.4    & 85.4        & 57.8       & 74.2      \\
                              & P-Linear & 80.8    & 85.4        & 59.4       & 75.2      \\
                              & Mix-K    & 81.4    & 85.5        & 60.3       & 75.7      \\ \midrule
\multirow{3}{*}{w/ sparsify}  & Linear   & 61.1    & 81.2        & 54.7       & 65.7      \\
                              & P-Linear & 82.7    & 86.1        & 61.8       & 76.9      \\
                              & Mix-K    & \textbf{82.8}    & \textbf{86.5}        & \textbf{62.3}       & \textbf{77.2}      \\ \bottomrule
\end{tabular}
\label{tab: abl_kernel}
\end{table}

%% file: tab/tab_abl_kernel_func.tex
\begin{table}[]
\centering
\caption{Ablation studies on the kernel function. We apply various kernel functions to evaluate their effectiveness in adapting the pre-trained ViT-B/16 model to the VTAB-1k benchmark. }
\begin{tabular}{@{}c|ccc|c@{}}
\toprule
Method     & Natural & Specialized & Sturctured & Mean Acc. \\ \midrule
Sigmoid    & 60.9    & 77.9        & 50.5       & 63.1      \\
RBF        & 55.0    & 81.5        & 39.6       & 58.7      \\
RBF-Normal & 81.0    & 85.2        & 59.3       & 75.2      \\
Mix-K      & \textbf{81.4}    & \textbf{85.5}        & \textbf{60.3}       & \textbf{75.7}      \\ \bottomrule
\end{tabular}
\label{tab: abl_kernel_func}
\end{table}

%% file: tab/tab_abl_sparse.tex
\begin{table}[]
\centering
\caption{Ablation studies on the sparsification mechanism. We provide comparisons of using kernelized LoRA~(KLoRA), sparsifying by competition within layers~(Within L.) and competition across layers~(Across L.). All experiments are conducted on VTAB-1k with pre-trained ViT-B/16.}
\setlength\tabcolsep{1pt}
\begin{tabular}{@{}ccc|ccc|c@{}}
\toprule
KLoRA & Within L. & Across L. & Natural & Specialized & Sturctured & Mean Acc. \\ \midrule
\checkmark     &             &             & 81.4    & 85.5        & 60.3       & 75.7      \\
\checkmark     & \checkmark           &             & \textbf{82.8}    & 86.2        & 62.0       & 77.0      \\
\checkmark     & \checkmark           & \checkmark           & \textbf{82.8}    & \textbf{86.5}        & \textbf{62.3}       & \textbf{77.2}      \\ \bottomrule
\end{tabular}
\label{tab: abl_saprse}
\end{table}

%% file: tab/tab_abl_importance.tex
\begin{table}[]
\centering
\caption{Ablation studies on the layer importance score. We provide comparisons between different metrics and sparsity allocation strategies. All experiments are conducted on VTAB-1k using a pre-trained ViT-B/16 with 0.01\% tunable parameters.}
\setlength\tabcolsep{2pt}
\begin{tabular}{@{}c|c|ccc|c@{}}
\toprule
 & Method      & Natural & Specialized & Sturctured & Mean Acc. \\ \midrule
\multirow{3}{*}{\begin{tabular}[c]{@{}c@{}}Alloc \\ per step\end{tabular}}  & Magnitude   & 81.3    & 84.6        & 58.7       & 74.9      \\

& W-Magnitude & 81.4    & 84.5        & 58.8       & 74.9      \\
                                                                            & Sensitivity & 81.4    & 84.7        & 59.0       & 75.0      \\ \midrule
\multirow{3}{*}{\begin{tabular}[c]{@{}c@{}}Alloc \\ per epoch\end{tabular}} & Magnitude   & 81.3    & 84.5        & 59.4       & 75.1      \\
                                                                            & W-Magnitude & 81.4    & 84.6        & 59.3       & 75.1      \\
                                                                            & Sensitivity & \textbf{81.5}    & \textbf{85.0}        & \textbf{59.5}       & \textbf{75.3}      \\ \bottomrule
\end{tabular}
\label{tab: abl_importance}
\end{table}

%% file: tab/tab_abl_budget.tex
\begin{table}[]
\centering
\caption{Ablation studies on the parameter budget $b$. All experiments are conducted on four datasets of VTAB-1k benchmark using a pre-trained ViT-B/16 with SNELLA-32.}
\setlength\tabcolsep{2pt}
\begin{tabular}{@{}c|cccc|c@{}}
\toprule
Budget       & Sun397         & Camelyon       & Clevr-count   & Snorb-azi     & Mean Acc.      \\ \midrule
$b_T/b_0=0.01$ & 53.74          & 83.91          & 83.57          & 33.51          & 63.68          \\
$b_T/b_0=0.10$  & \textbf{53.98} & 84.52          & 83.56          & 33.43          & 63.87          \\
$b_T/b_0=0.30$  & 53.91          & \textbf{84.99} & 83.43          & 33.34          & \textbf{63.92} \\
$b_T/b_0=0.50$  & 53.87          & 84.84          & 82.97          & \textbf{33.68} & 63.84          \\
$b_T/b_0=0.70$  & 53.70          & 84.45          & 83.28          & 33.20          & 63.66          \\
$b_T/b_0=0.90$  & 53.76          & 84.44          & \textbf{83.69} & 32.98          & 63.72          \\
$b_T/b_0=0.99$ & 53.73          & 83.46          & 82.50          & 31.83          & 62.88          \\ \bottomrule
\end{tabular}
\label{tab: abl_budget}
\end{table}

%% file: tab/tab_abl_schedule.tex
\begin{table}[htbp]
\centering
\caption{Ablation studies on the budget scheduling strategy. All experiments are conducted on four datasets of VTAB-1k benchmark using a pre-trained ViT-B/16 with SNELLA-32.}
\setlength\tabcolsep{3pt}
\begin{tabular}{@{}l|cccc|c@{}}
\toprule
Strategy  & Sun397         & Camelyon       & Clevr-count   & Snorb-azi     & Mean Acc.      \\ \midrule
Constant  & 53.35          & 82.62          & 82.07          & 26.22          & 61.07          \\
Linear    & \textbf{54.26} & 84.80          & 82.20          & 32.72          & 63.50          \\
Quadratic & 54.07          & 84.63          & 83.09          & 32.94          & 63.68          \\
Cubic     & 54.00          & \textbf{84.84} & \textbf{83.69} & \textbf{32.98} & \textbf{63.88} \\ \bottomrule
\end{tabular}
\label{tab: abl_schedule}
\end{table}

%% file: sec/5_conclusion.tex
\section{Conclusion}
We propose SNELLA, a PEFT method for high-performance sparse tuning with low memory usage. We adjust the pre-trained weight matrix by adding it to another matrix. This matrix is merged by learnable low-rank matrices via non-linear kernel functions, thereby reducing the volume of learnable parameters stored in the optimizer while preserving the expressivity of the merged matrix. We design an adaptive bi-level sparsity allocation mechanism that encourages the competitions within and across layers to locate the task-relevant weights in an end-to-end manner. Extensive experiments on image classification, medical segmentation, and text-to-image generation demonstrated the ability of SNELLA to leverage the high performance of sparse tuning and the low memory usage of LoRA. For future work, we will investigate the hybrid mechanism that combines non-linear and linear methods, and apply it to large-scale downstream tasks, such as the instruction fine-tuning of vision-language models and vision generation tasks.

%% file: sec/6_appendix.tex
\section{More Details of Experimental Setup}
\subsection{Contenders}
\label{sec: app_contenders}
\begin{itemize}[leftmargin=*]
    \item Full: fully tunes all the model parameters.
    \item Linear: freezes all the backbone parameters and only tunes the linear classification head.
    \item Bias~\cite{zaken2021bitfit}: freezes all model parameters except for the bias term and the linear classification head.
    \item Partial-1: freezes the backbone except for the last 1 layer and also tunes the classification head as described in \cite{jia2022visual}.
    \item MLP-3: freezes the backbone and tunes the classification head implemented by a trainable 3-layer perceptron as described in \cite{jia2022visual}.
    \item VPT-Shallow~\cite{jia2022visual}: freezes all the backbone parameters while introducing additional trainable prompts to the input space of the pretrained ViT.
    \item VPT-Deep~\cite{jia2022visual}: freezes the backbone while appending additional trainable prompts to the sequence in the multi-head self-attention layer of each ViT block.
    \item Adapter-$r$~\cite{houlsby2019parameter}: freezes all the backbone parameters while adding a down projection, a ReLU non-linearity, and an up projection layer sequentially in the feed-forward network (FFN) of each visual Transformer block. We report the performance implemented by \cite{he2023sensitivity} for comparison.
    \item LoRA-$r$~\cite{hu2021lora}: freezes all the backbone parameters while adding a parallel branch including two low-rank matrices to the weight matrices in the linear layers to efficiently update them. The low-rank matrices can be merged into the backbone weights after fine-tuning. We report the performance implemented by \cite{he2023sensitivity} for comparison.
    \item SPT~\cite{he2023sensitivity}: identifies the tunable parameters for a given task in a data-dependent way, and utilizes LoRA~(SPT-LoRA) or Adapter~(SPT-Adapter) for weight matrices with a large number of tunable parameters and sparse tuning for weight matrices with a small number of tunable parameters. 
    \item VQT~\cite{tu2023visual}: introduces a handful of learnable query tokens to each layer for adaptation.
    \item DoRA~\cite{liu2024dora}: decomposes the pre-trained weight into two components, {\it i.e.}, magnitude and direction, for fine-tuning. It specifically employs LoRA for directional updates to efficiently minimize the number of trainable parameters.
    \item GPS~\cite{DBLP:conf/cvpr/ZhangZGZSZZ24}: identifies task-dependent tunable weights and applies sparse tuning to these weights.
\end{itemize}
\subsection{Details of Matrix Fitting}
Given a random matrix $\mathbf{W}^{(gt)}\in\mathbb{R}^{m\times n}$, we fit this matrix by merging two low-rank learnable matrices $\mathbf{B}\in\mathbb{R}^{m\times r},\mathbf{A}\in\mathbb{R}^{n\times r}$ with different kernel functions $\kappa$,
\begin{equation}
    \min\limits_{\mathbf{A},\mathbf{B}} \frac{1}{mn}\sum\limits_{i=1}^m\sum\limits_{j=1}^n(\mathbf{W}^{(gt)}_{ij}-\kappa(\mathbf{B}_{i,\cdot},\mathbf{A}_{j,\cdot}))^2.
\end{equation}
We use gradient descent for $1e5$ optimization steps, employing the Adam optimizer with a learning rate of $1e-4$.
We fit 10 randomly generated matrices for each kernel and report the average MSE Loss in Figure 4(b) of the main paper.

\input{tab/tab_vtab_details}

\subsection{Kernel Function Definition}
\label{sec: app_kernel}
Consider a vector space $\mathbb{R}^r$, a kernel function $\kappa: \mathbb{R}^r\times \mathbb{R}^r \rightarrow \mathbb{R}$ is called a positive semi-definite kernel on $\mathbb{R}^r$ if 
\begin{equation}
    \sum\limits_{i=1}^n\sum\limits_{j=1}^n c_ic_j\kappa(\mathbf{x}_i, 
    \mathbf{x}_j)\ge 0
\end{equation}
holds for all $\mathbf{x}_1, ..., \mathbf{x}_n \in \mathbb{R}^r, c_1, ..., c_n \in \mathbb{R}, n\in\mathbb{N}$.

Given two vectors $\mathbf{x}, \mathbf{x'}\in\mathbb{R}^{r}$, we show the widely-used kernel functions in Table~\ref{tab: app_kernel}.
We introduce additional learnable parameters~(the \textit{e.g.} $\alpha$ for Sigmoid and RBF kernel, $\alpha_p$ for piecewise linear kernel) that enable the merged adaptation matrix $\Delta \mathbf{W}$ to accommodate both positive and negative values. The additional parameters select certain elements in the matrix and assign them negative values, without compromising the high-rank property of the merged adaptation matrix $\Delta \mathbf{W}$.
We set $P=2$ for the piecewise linear kernel.

\subsection{Positive semi-definiteness of Mix-K} 
First, the positive semi-definite property of Mix-K is preserved because linear combinations and compositions of multiple kernel functions still yield valid kernel functions. This preservation is ensured as Mix-K is constructed based on both RBF and piecewise linear kernels. 
Second, mere positive semi-definiteness is insufficient, as low-rank matrices must also be capable of constructing matrices that include negative values.
Therefore, we introduce additional learnable coefficients ($\alpha$, $\beta$, and $\alpha_p$) that can take negative values. Although kernels with these coefficients are no longer strictly positive semi-definite, the set of matrices they can generate includes those corresponding to positive semi-definite kernels~(\textit{i.e.}, when all coefficients are set to 1, the formulation degenerates to the positive semi-definite case).
\begin{table}[]
\centering
\caption{Expression of kernel function utilized in the main text.}
\begin{tabular}{@{}ll@{}}
\toprule
Function  & Expression \\ \midrule
Linear           & $\kappa(\mathbf{x}, \mathbf{x'})=\mathbf{x}^\top \mathbf{x'}$           \\
P-Linear & $\kappa(\mathbf{x}, \mathbf{x'})=\sum\limits_{p=1}^P\alpha_p\Vert \mathbf{x}_{\lceil \frac{rp}{P}\rceil :\lceil \frac{r(p+1)}{P}\rceil}-\mathbf{x '}_{\lceil \frac{rp}{P}\rceil :\lceil \frac{r(p+1)}{P}\rceil} \Vert_2$           \\
Sigmoid          & $\kappa(\mathbf{x}, \mathbf{x'})=\alpha (1+exp(-\beta\mathbf{x}^\top \mathbf{x'}))^{-1} + \gamma$          \\
RBF              & $\kappa(\mathbf{x}, \mathbf{x'})=\alpha (exp(-\beta \Vert \mathbf{x}-\mathbf{x'}\Vert_2^2))) + \gamma$           \\ \bottomrule
\end{tabular}
\label{tab: app_kernel}
\vspace{-0.4cm}
\end{table}

\subsection{Discussions with Multiple Kernel Learning}
Although Mix-K is designed to address the end-to-end optimization challenges of the exponential kernel, the mixture of different kernels may not only facilitate optimization but also enhance the model's generalization performance. This perspective is supported by existing research in multi-kernel learning~\cite{gonen2011multiple}. In the context of multi-kernel learning, it has been established that multi-kernel models offer greater flexibility compared to models based on a single kernel function. Specifically, the high-dimensional space generated by combining multiple kernel functions constitutes a composite feature space, formed through the integration of individual feature subspaces. This composite space combines the distinct feature mapping capabilities of each subspace, enabling heterogeneous features in the data to be processed using different kernel functions. Consequently, the downstream dataset can be more accurately and appropriately represented in this new composite space, which ultimately enhances the classification accuracy of the sample data.
Furthermore, the contribution of our proposed Mix-K to multi-kernel learning methods lies in the design of a mixture of kernels that can be effectively integrated into end-to-end optimization of deep learning models, offering both strong expressivity and enhanced training stability.

\subsection{Discussions with AdaLoRA}
In our sparsity allocation mechanism, we adopt the sensitivity metric from AdaLoRA as a measure of layer importance. However, our approach differs in the following aspects.
\textit{(i) Allocation targets}. AdaLoRA allocates the ranks of learnable matrices across layers, whereas SNELLA allocates the number of tunable weights within the original weight matrix, requiring the design of Algorithm 1 in the main paper.
\textit{(ii) Importance scores}. AdaLoRA determines layer importance using singular value decomposition, whereas SNELLA computes layer importance by aggregating the contributions of two learnable low-rank matrices. 
\textit{(iii) Weight-level allocation mechanism}. After performing layer-wise rank allocation, AdaLoRA prunes ranks within each layer according to the magnitude of singular values, while SNELLA employs a weight-level competition strategy and identifies tunable weights based on the magnitude of their parameter updates.

\begin{table*}[]
\centering
\caption{Training time cost of different PEFT methods on ViT-B/16 pre-trained on ImageNet-21k.}
\begin{tabular}{@{}c|c|cc|cc|cc@{}}
\toprule
Method                & LoRA-8 & SPT-LoRA & SPT-Adapter & KLoRA-8 & \begin{tabular}[c]{@{}c@{}}KLoRA-8\\ (saving $\Delta \mathbf{W}$)\end{tabular} & SNELLA-8 & \begin{tabular}[c]{@{}c@{}}SNELLA-8\\ (saving $\Delta \mathbf{W}$)\end{tabular} \\ \midrule
Training time (s/img) & 0.2009 & 0.3693   & 0.3751      & 0.2860  & 0.2050                                                                         & 0.3574  & 0.2735                                                                         \\ \bottomrule
\end{tabular}
\label{tab: app_time}
\vspace{-0.4cm}
\end{table*}

\begin{table}
	\centering
 \caption{Memory usage comparison between SNELL and LoRA. $\Delta$ Mem. denotes the incremental memory usage of SNELL in comparison to LoRA.}
	\setlength{\tabcolsep}{3pt}
	\scalebox{1.0}{
		\begin{tabular}{@{}lccc@{}}
\toprule
\begin{tabular}[c]{@{}c@{}}{Pre-trained}\\Model\end{tabular} & \begin{tabular}[c]{@{}c@{}}LoRA-8 \\ Mem.~(MB)\end{tabular} & \begin{tabular}[c]{@{}c@{}}SNELLA-8 \\ Mem.~(MB)\end{tabular} & \begin{tabular}[c]{@{}c@{}}$\Delta$ Mem. / \\ SNELLA-8 Mem. \end{tabular} \\ \midrule
ViT-B/16          & 1546                                                     & 1673                                                      & 0.076                                                              \\
ViT-L/16          & 4325                                                     & 4519                                                      & 0.043                                                              \\
ViT-H/16          & 9325                                                     & 9692                                                      & 0.038                                                              \\ \bottomrule
\end{tabular}}
	
	\label{tab: memory_comp_lora} 
	\vspace{-0.2cm}
\end{table}
\input{tab/tab_abl_adaptive}

\subsection{Details of Polyp Segmentation}
When adapting the pre-trained SAM ViT-B to the polyp segmentation task, we train the model for 20 epochs with a batch size of 1, employ an initial learning rate of 1e-3 along with a cosine learning rate decay schedule that reduces the learning rate to a minimum of 1e-7. The model parameters are optimized using the AdamW optimizer.

\subsection{Details of Image Generation}
For the experiments on the DreamBooth dataset, we fine-tune the SD3 model based on the workflow of the diffusers repository~\cite{von-platen-etal-2022-diffusers}. Specifically, we utilized the Accelerate~\cite{accelerate} for training, along with an 8-bit Adam optimizer and gradient checkpoint technique. The resolution of the output images is set to 512×512, the training batch size is 1, and the gradient accumulation steps are 4. A learning rate of 1e-4 is adopted without any additional learning rate scheduling strategy, and the model is fine-tuned for 500 steps.

\begin{figure*}
  \centering
  \includegraphics[width=0.95\linewidth]{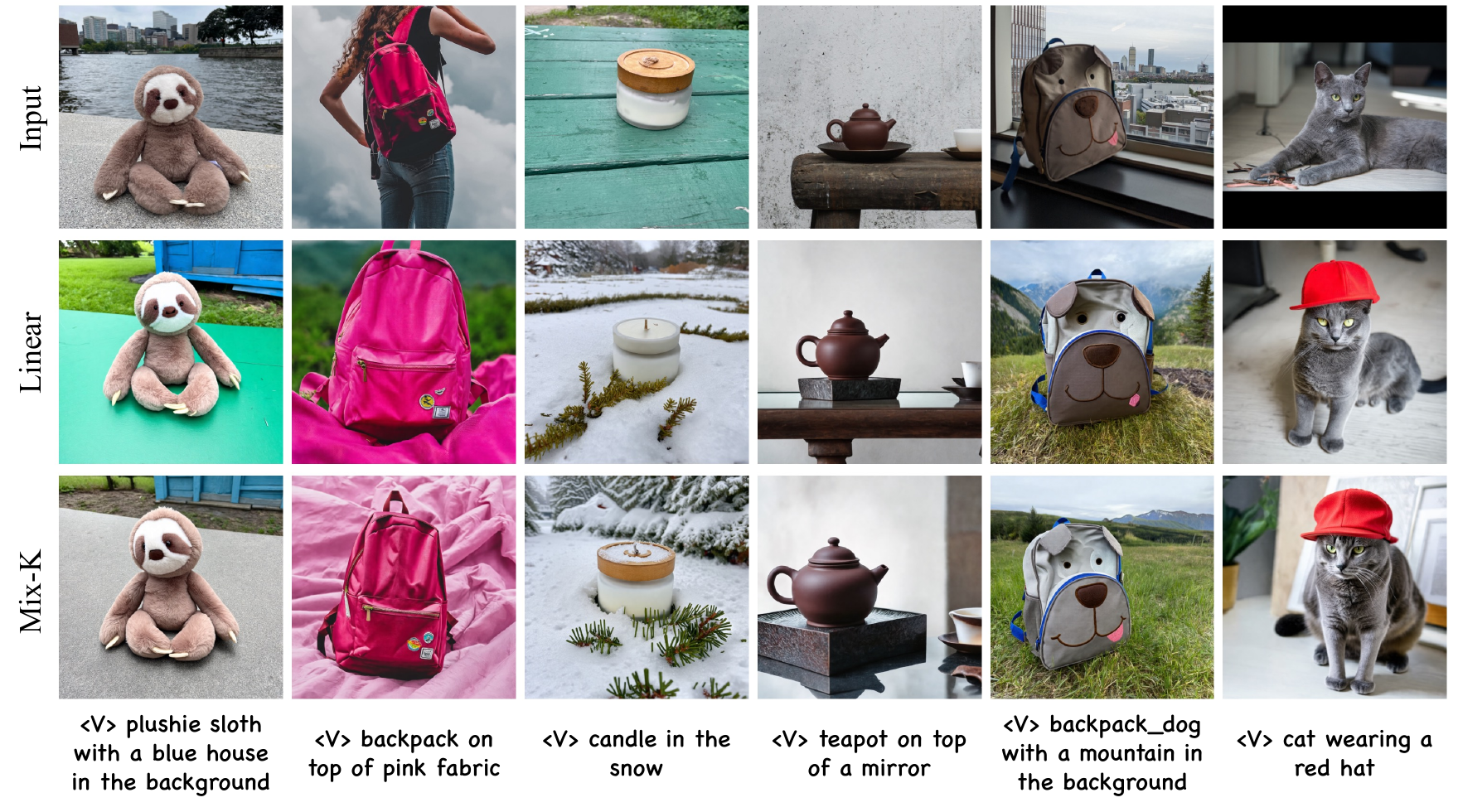}
  \vspace{-0.5cm}
  \caption{Qualitative results on concept customization for personalized text-to-image generation. Compared to SNELLA with a linear kernel (middle row), utilizing a nonlinear kernel Mix-K (bottom row) achieves higher concept fidelity and text alignment of the generated images.}
  \label{fig: generation_linear}
  \vspace{-0.3cm}
\end{figure*}

\section{Additional Experiments}

\subsection{Per-task Results on VTAB-1k Benchmark}
We provide the per-task results on the VTAB-1k benchmark using ViT-B/16 supervised pre-trained on ImageNet21K in Table~\ref{tab: vtab1k_performance_detail}.
Our SNELLA has demonstrated superior performance by achieving SOTA performance on 13 downstream tasks. Additionally, SNELLA achieves SOTA performance on the mean accuracy across all tasks~(74.6\% \textit{v.s.} 74.9\%), indicating its effectiveness in various domains.

\subsection{Memory Usage of Nonlinear Kernel Functions}
In Figure 7(a) of the main paper, we observe that SNELLA requires additional memory usage compared to LoRA due to the incorporation of nonlinear kernel functions. To explore whether the impact of this additional usage hinders the usability of SNELLA on large models, 
we compare the memory usage between SNELLA and LoRA as the model size grows~(in Table~\ref{tab: memory_comp_lora}). As the model size expands, the incremental memory usage of SNELLA becomes negligible.

\subsection{Training Time Analysis}
\label{sec: app_time}
Table~\ref{tab: app_time} provides a comparison of training time costs between SNELLA and other PEFT methods using a single NVIDIA GeForce RTX 4090 GPU.
The training time of SNELLA-8 is higher than LoRA-8~(0.3574 \textit{v.s.} 0.2009) while lower than SPT-LoRA~(0.3574 \textit{v.s.} 0.3693) and SPT-Adapter~(0.3574 \textit{v.s.} 0.3751). We can observe that the additional time cost of SNELLA mainly stems from two aspects.
First, the recomputation of the merged adaptation matrix $\Delta \mathbf{W}$ incurs additional time overhead. This is evident from the performance comparison between SNELLA-8 and SNELLA-8~(with $\Delta \mathbf{W}$ saved), as well as between KLoRA-8 and KLoRA-8~(with $\Delta \mathbf{W}$ saved).
Second, sparsification with our competition-based mechanism also requires time for the calculation of importance scores and the allocation of sparsity, which is reflected by the comparison between KLoRA-8 and SNELLA-8. 
Indeed, although SNELLA incurs a higher time cost compared to LoRA, it achieves a significant performance improvement with comparable memory usage, as shown in Figure 7 of the main paper. Moreover, compared to the existing sparse tuning method SPT, SNELLA consistently outperforms SPT in terms of accuracy, memory efficiency, and training time.

\input{tab/tab_abl_llm}

\subsection{Ablations of Adaptive Sparsification}

To verify the effectiveness of the proposed adaptive sparsification mechanism, we compare the performance on FGVC datasets between pre-defined fixed masks and our adaptive sparsification with different kernel functions in Table~\ref{tab: abl_adaptive}. The pre-defined masks are generated by SPT~\cite{he2023sensitivity}. 
First, we observe that, compared to piecewise linear functions, the use of Mix-K leads to improved performance on pre-defined masks~(89.4\% \textit{v.s.} 90.6\%). This improvement can be attributed to the greater expressive capacity of Mix-K, which allows the model to learn a wider range of parameter update patterns under fixed mask constraints.
Second, compared to our adaptive sparsification strategy, pre-defined fixed masking can hardly identify and adjust the most task-relevant weights in an end-to-end fashion, which leads to performance degradation (90.6\% \textit{v.s.} 91.9\%).

\begin{figure*}
  \centering
  \includegraphics[width=0.92\linewidth]{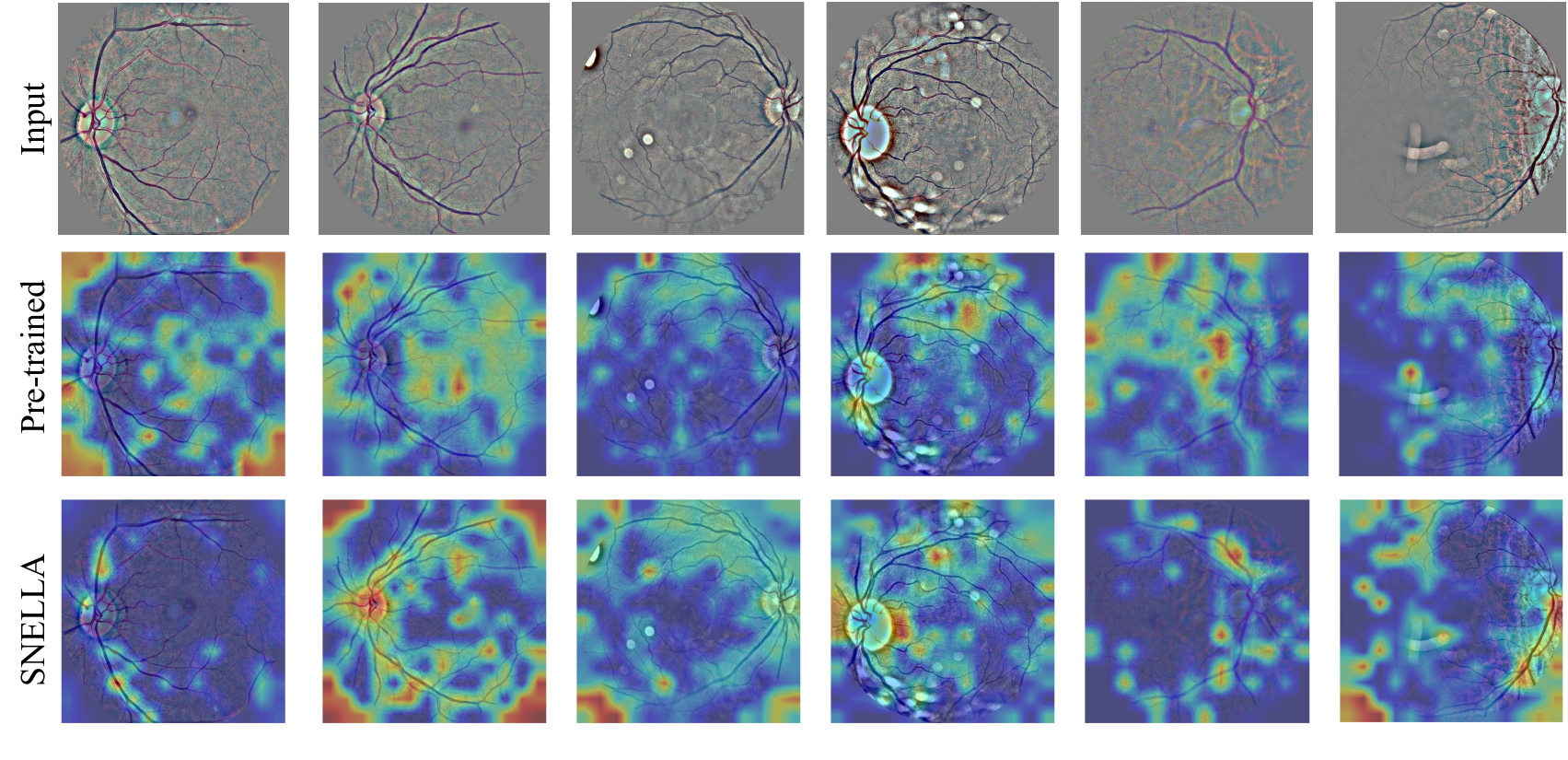}
  \vspace{-0.8cm}
  \caption{Visualization of heatmaps with Grad-CAM. We provide the input image~(\textit{top}), results of the pre-trained model~(\textit{middle}), and results of the model fine-tuned by SNELLA~(\textit{bottom}). The experiments are conducted with a pre-trained ViT-B/16 model on the RetinoPathy dataset.}
  \label{fig: gradcam}
  \vspace{-0.3cm}
\end{figure*}

\begin{figure*}
  \centering
  \includegraphics[width=1.0\linewidth]{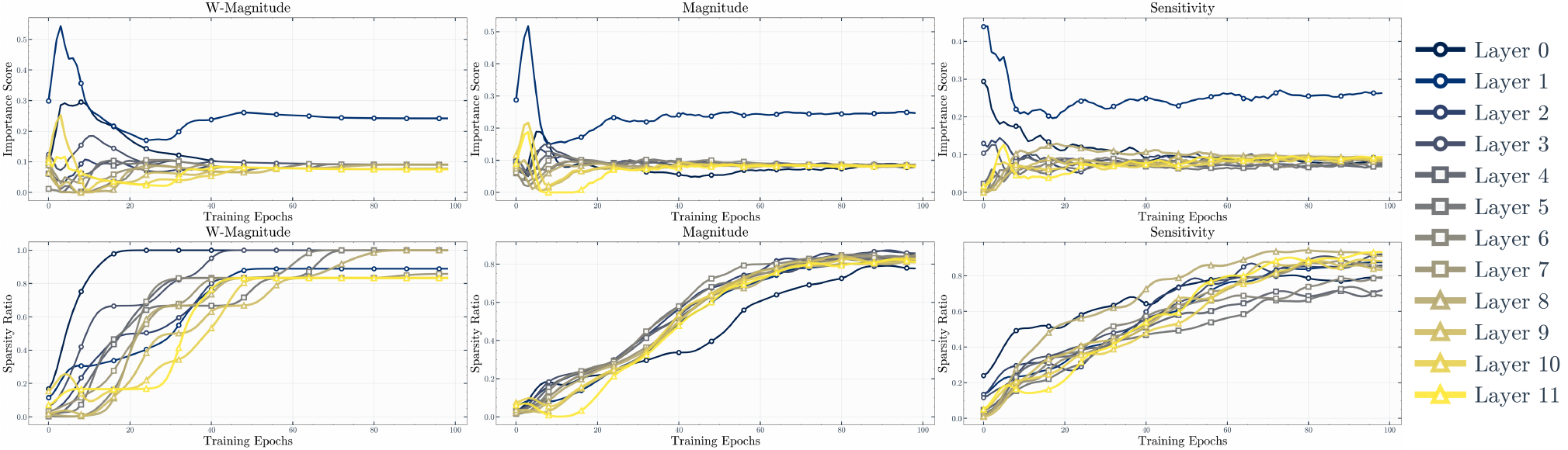}
  \vspace{-0.2cm}
  \caption{Evolution of the importance score~(\textit{top}) and the sparsity ratio~(\textit{bottom}) across layers during the fine-tuning process. We provide the results for three types of importance score~(W-Magnitude, Magnitude, and Sensitivity) using a pre-trained ViT-B/16 on the Clevr-count dataset from VTAB-1k benchmark with SNELLA-32. The importance score of each layer is calculated by averaging the importance scores of all its weight matrices.}
  \label{fig: sparsity_evolution}
  \vspace{-0.3cm}
\end{figure*}

\subsection{Ablations of Non-linear Kernels}
To further validate the necessity of nonlinear kernels, we conducted additional ablation studies on generation tasks. As illustrated in Figure~\ref{fig: generation_linear}, SNELLA with nonlinear kernels outperforms the linear kernel-based approach in both generation quality and concept fidelity. 
For example, given the text "a backpack on top of pink fabric," SNELLA with linear kernels tends to render the target backpack in pink as well, thereby reducing concept fidelity. In contrast, SNELLA with nonlinear kernels tends to retain the backpack's original color. For text ``a teapot on top of a mirror", SNELLA with linear kernels incorrectly interprets "mirror" as "glass", reducing the alignment between text and images. In contrast, SNELLA with nonlinear kernels successfully generates the reflection of the target teapot in the mirror.

\subsection{GradCAM Visualizations}
In Figure~\ref{fig: gradcam}, we visualize the GradCAM heatmap of the pre-trained model and the model fine-tuned by SNELLA. The base model is a ViT-B/16 pre-trained on ImageNet-21K. We observe that pre-trained models tend to overlook the target object regions when applied to tasks that differ substantially from the pre-training dataset. In contrast, models fine-tuned using SNELLA demonstrate a more accurate focus on the target object region in downstream tasks.

\subsection{Evolution of Sparsity Ratios during Fine-tuning}
To better understand the sparsity allocation process of SNELLA during fine-tuning, we present the evolution of importance scores and sparsity ratios~(the ratio of zero-valued weight updates relative to the total number of weights) within each layer at different training steps in Figure~\ref{fig: sparsity_evolution}. 
First, the value of W-Magnitude remains stable during fine-tuning. As $b_t$ decreases over steps, layers with low importance score may be fully frozen~(\textit{i.e.}, with a sparsity ratio of 1). This excessive penalization hinders the model of achieving advanced performance on downstream tasks~\cite{DBLP:conf/cvpr/ZhangZGZSZZ24}.
Second, although Magnitude values adjust dynamically during fine-tuning, the differences between layers are minimal, which leads to similar sparsity ratios being assigned across different layers.
Finally, utilizing sensitivity allows for a dynamically evolving assessment of layer importance and enables the assignment of different sparsity ratios across layers.

\subsection{Comparisons with PEFT Methods for LLMs}
We apply SNELLA to LLaMA2-7B~\cite{touvron2023llama} to adapt it for the commonsense reasoning benchmark. As shown in Table~\ref{tab: abl_llm}, SNELLA outperforms LoRA but performs worse than DoRA~\cite{liu2024dora}, the current state-of-the-art PEFT method designed specifically for NLP tasks. We then compare the performance of SNELLA and DoRA on visual tasks, and find that SNELLA significantly outperforms DoRA. 
The experimental results demonstrate that differences in data modalities and downstream tasks necessitate distinct fine-tuning strategies for LLMs and vision models to achieve optimal performance. No single fine-tuning approach can consistently yield optimal results across all tasks and modalities, a finding that corroborates the conclusions presented in \cite{zhang2024neural}. While this study focuses on vision tasks, extending sparse tuning to NLP tasks presents a promising direction, which we intend to explore in future work.

%% file: tab/tab_vtab_details.tex
\begin{table*}[]
\centering
\caption{Top-1 accuracy (\%) on VTAB-1k benchmarks using ViT-B/16 backbone pre-trained on ImageNet-21k supervisedly. The best result is in \textbf{bold}, and the second-best result is {\ul underlined}.}
\vspace{-0.3cm}
\setlength\tabcolsep{4pt}
\scalebox{1.0}{
\begin{tabular}{@{}ccccccccccccccccccccc@{}}
\toprule
\multicolumn{1}{c|}{\multirow{2}{*}{}} & \multicolumn{7}{c|}{Natural}                                                                                                                                                                                                                                                                                                             & \multicolumn{4}{c|}{Specialized}                                                                                                                                                                             & \multicolumn{8}{c|}{Structured}                                                                                                                                                                                                                                                                                                                                                                           & -                                            \\ \cline{2-21} 
\multicolumn{1}{c|}{}                  & \rotatebox{90}{Cifar100} & \rotatebox{90}{Caltech101} & \rotatebox{90}{DTD} & \rotatebox{90}{Flower102} & \rotatebox{90}{SVHN} & \rotatebox{90}{Sun397} & \multicolumn{1}{c|}{\rotatebox{90}{Pets}} & \rotatebox{90}{Camelyon} & \rotatebox{90}{EuroSAT} & \rotatebox{90}{Resisc45} & \multicolumn{1}{c|}{\rotatebox{90}{Retinopathy}} & \rotatebox{90}{Clevr-Count} & \rotatebox{90}{Clevr-Dist} & \rotatebox{90}{DMLab} & \rotatebox{90}{KITTI-Dist} & \rotatebox{90}{dSpr-Loc} & \rotatebox{90}{dSpr-Ori} & \rotatebox{90}{sNORB-Azim} & \multicolumn{1}{c|}{\rotatebox{90}{sNORB-Ele}} & \rotatebox{90}{Mean Acc.} \\ \midrule
\multicolumn{1}{c|}{Full}              & 68.9                                        & 87.7                                          & 64.3                                   & 97.2                                         & 87.4                                    & 38.8                                      & \multicolumn{1}{c|}{86.9}                                    & 79.7                                        & 95.7                                       & 84.2                                        & \multicolumn{1}{c|}{73.9}                                           & 56.3                                           & 58.6                                          & 41.7                                     & 65.5                                          & 57.5                                        & 46.7                                        & 25.7                                          & \multicolumn{1}{c|}{29.1}                                         & 65.6                                         \\ \midrule
\multicolumn{21}{c}{Additional-based methods}                                                                                                                                                                                                                                                                                                                                                                                                                                                                                                                                                                                                                                                                                                                                                                                                                                                                                                                                                                                                               \\ \midrule
\multicolumn{1}{c|}{MLP-3}             & 63.8                                        & 84.7                                          & 62.3                                   & 97.4                                         & 32.5                                    & 49.2                                      & \multicolumn{1}{c|}{84.7}                                    & 77.0                                        & 88.0                                       & 70.2                                        & \multicolumn{1}{c|}{56.1}                                           & 47.8                                           & 32.8                                          & 32.3                                     & 58.1                                          & 12.9                                        & 21.2                                        & 15.2                                          & \multicolumn{1}{c|}{24.8}                                         & 53.2                                         \\
\multicolumn{1}{c|}{VPT-Shallow}       & {\ul 77.7}                                  & 86.9                                          & 62.6                                   & 97.5                                         & 74.5                                    & 51.2                                      & \multicolumn{1}{c|}{87.3}                                    & 78.2                                        & 92.0                                       & 75.6                                        & \multicolumn{1}{c|}{72.9}                                           & 50.5                                           & 58.6                                          & 40.5                                     & 67.1                                          & 68.7                                        & 36.1                                        & 20.2                                          & \multicolumn{1}{c|}{34.1}                                         & 64.9                                         \\
\multicolumn{1}{c|}{VPT-Deep}          & \textbf{78.8}                               & 90.8                                          & 65.8                                   & 98.0                                         & 78.1                                    & 49.6                                      & \multicolumn{1}{c|}{88.3}                                    & 81.8                                        & 96.1                                       & 83.4                                        & \multicolumn{1}{c|}{68.4}                                           & 68.5                                           & 60.0                                          & 46.5                                     & 72.8                                          & 73.6                                        & 47.9                                        & 32.9                                          & \multicolumn{1}{c|}{37.8}                                         & 69.4                                         \\
\multicolumn{1}{c|}{Adapter-8}         & 69.2                                        & 90.1                                          & 68.0                                   & 98.8                                         & 82.8                                    & 54.3                                      & \multicolumn{1}{c|}{89.9}                                    & 84.0                                        & 94.9                                       & 81.9                                        & \multicolumn{1}{c|}{75.5}                                           & 80.9                                           & 65.3                                          & 48.6                                     & 78.3                                          & 74.8                                        & 48.5                                        & 29.9                                          & \multicolumn{1}{c|}{41.6}                                         & 71.4                                         \\
\multicolumn{1}{c|}{Adapter-32}        & 68.7                                        & 92.2                                          & 69.8                                   & 98.9                                         & 84.2                                    & 53.0                                      & \multicolumn{1}{c|}{90.3}                                    & 83.2                                        & 95.4                                       & 83.2                                        & \multicolumn{1}{c|}{74.3}                                           & 81.9                                           & 63.9                                          & 48.7                                     & 80.6                                          & 76.2                                        & 47.6                                        & 30.8                                          & \multicolumn{1}{c|}{36.4}                                         & 71.5                                         \\
\multicolumn{1}{c|}{NOAH}              & 69.6                                        & 92.7                                          & 70.2                                   & 99.1                                         & 86.1                                    & 53.7                                      & \multicolumn{1}{c|}{90.4}                                    & 84.4                                        & 95.4                                       & 83.9                                        & \multicolumn{1}{c|}{75.8}                                           & 82.8                                           & {\ul 68.9}                                    & 49.9                                     & 81.7                                          & 81.8                                        & 48.3                                        & 32.8                                          & \multicolumn{1}{c|}{\textbf{44.2}}                                & 73.2                                         \\
\multicolumn{1}{c|}{SPT-Adapter}       & 72.9                                        & 93.2                                          & {\ul 72.5}                             & {\ul 99.3}                                   & 88.8                                    & {\ul 55.8}                                & \multicolumn{1}{c|}{91.4}                                    & \textbf{86.2}                               & 96.1                                       & 85.5                                        & \multicolumn{1}{c|}{75.5}                                           & 83.0                                           & 68.0                                          & 51.9                                     & 81.2                                          & 82.4                                        & \textbf{51.9}                               & 31.7                                          & \multicolumn{1}{c|}{41.2}                                         & 74.1                                         \\ \midrule
\multicolumn{21}{c}{Reparameterized-based methods}                                                                                                                                                                                                                                                                                                                                                                                                                                                                                                                                                                                                                                                                                                                                                                                                                                                                                                                                                                                                          \\ \midrule
\multicolumn{1}{c|}{Linear}            & 63.4                                        & 85.0                                          & 63.2                                   & 97.0                                         & 36.6                                    & 51.0                                      & \multicolumn{1}{c|}{86.3}                                    & 78.5                                        & 87.5                                       & 68.6                                        & \multicolumn{1}{c|}{74.0}                                           & 34.3                                           & 30.6                                          & 33.2                                     & 55.4                                          & 12.5                                        & 20.0                                        & 9.6                                           & \multicolumn{1}{c|}{19.2}                                         & 52.9                                         \\
\multicolumn{1}{c|}{Partial-1}         & 66.8                                        & 85.9                                          & 62.5                                   & 97.3                                         & 37.6                                    & 50.6                                      & \multicolumn{1}{c|}{85.5}                                    & 78.6                                        & 89.8                                       & 72.5                                        & \multicolumn{1}{c|}{73.3}                                           & 41.5                                           & 34.3                                          & 33.9                                     & 61.0                                          & 31.3                                        & 32.8                                        & 16.3                                          & \multicolumn{1}{c|}{22.4}                                         & 56.5                                         \\
\multicolumn{1}{c|}{Bias}              & 72.8                                        & 87.0                                          & 59.2                                   & 97.5                                         & 59.9                                    & 51.4                                      & \multicolumn{1}{c|}{85.3}                                    & 78.7                                        & 91.6                                       & 72.9                                        & \multicolumn{1}{c|}{69.8}                                           & 61.5                                           & 55.6                                          & 32.4                                     & 55.9                                          & 66.6                                        & 40.0                                        & 15.7                                          & \multicolumn{1}{c|}{25.1}                                         & 62.0                                         \\
\multicolumn{1}{c|}{LoRA-8}            & 67.1                                        & 91.4                                          & 69.4                                   & 98.8                                         & 85.3                                    & 54.0                                      & \multicolumn{1}{c|}{90.4}                                    & 84.9                                        & 95.3                                       & 84.4                                        & \multicolumn{1}{c|}{73.6}                                           & 82.9                                           & \textbf{69.2}                                 & 49.8                                     & 78.5                                          & 75.7                                        & 47.1                                        & 31.0                                          & \multicolumn{1}{c|}{44.0}                                         & 72.3                                         \\
\multicolumn{1}{c|}{LoRA-16}           & 68.1                                        & 91.4                                          & 69.8                                   & 99.0                                         & 86.4                                    & 53.1                                      & \multicolumn{1}{c|}{90.5}                                    & 85.1                                        & 95.8                                       & 84.7                                        & \multicolumn{1}{c|}{74.2}                                           & 83.0                                           & 66.9                                          & 50.4                                     & 81.4                                          & 80.2                                        & 46.6                                        & 32.2                                          & \multicolumn{1}{c|}{41.1}                                         & 72.6                                         \\
\multicolumn{1}{c|}{SPT-LoRA}          & 73.5                                        & 93.3                                          & {\ul 72.5}                             & {\ul 99.3}                                   & 87.9                                    & 55.5                                      & \multicolumn{1}{c|}{91.5}                                    & 85.7                                        & {\ul 96.2}                                 & 85.9                                        & \multicolumn{1}{c|}{75.9}                                           & {\ul 84.4}                                     & 67.6                                          & 52.5                                     & 82.0                                          & 81.0                                        & 51.1                                        & 30.2                                          & \multicolumn{1}{c|}{41.3}                                         & 74.1                                         \\ \midrule
\multicolumn{1}{c|}{SNELL-8}           & 73.7                                        & 92.7                                          & 72.4                                   & 99.2                                         & 89.2                                    & 55.4                                      & \multicolumn{1}{c|}{91.4}                                    & 84.9                                        & 96.1                                       & 86.4                                        & \multicolumn{1}{c|}{75.2}                                           & 84.0                                           & 68.5                                          & {\ul 53.5}                               & 81.0                                          & {\ul 82.7}                                  & 49.9                                        & 33.9                                          & \multicolumn{1}{c|}{39.2}                                         & 74.2                                         \\
\multicolumn{1}{c|}{SNELL-16}          & 74.2                                        & 93.4                                          & {\ul 72.5}                             & {\ul 99.3}                                   & 90.2                                    & 55.7                                      & \multicolumn{1}{c|}{91.4}                                    & 85.7                                        & 95.8                                       & 86.5                                        & \multicolumn{1}{c|}{{\ul 76.3}}                                     & {\ul 84.4}                                     & 68.2                                          & 53.0                                     & 82.0                                          & 82.2                                        & 49.6                                        & 33.3                                          & \multicolumn{1}{c|}{40.6}                                         & 74.4                                         \\
\multicolumn{1}{c|}{SNELL-32}          & 74.5                                        & 93.4                                          & \textbf{73.1}                          & {\ul 99.3}                                   & 91.1                                    & \textbf{55.9}                             & \multicolumn{1}{c|}{91.5}                                    & 85.5                                        & 96.1                                       & 86.5                                        & \multicolumn{1}{c|}{76.2}                                           & 83.4                                           & 68.6                                          & 52.2                                     & 81.3                                          & \textbf{83.2}                               & 50.7                                        & 35.9                                          & \multicolumn{1}{c|}{39.0}                                         & {\ul 74.6}                                   \\ \midrule
\multicolumn{1}{c|}{SNELLA-8}          & 73.8                                        & 93.5                                          & 72.7                                   & \textbf{99.4}                                & 89.5                                    & 55.7                                      & \multicolumn{1}{c|}{{\ul 91.6}}                              & 85.2                                        & 96.0                                       & 86.9                                        & \multicolumn{1}{c|}{75.6}                                           & \textbf{85.0}                                  & 68.0                                          & {\ul 53.5}                               & 82.3                                          & 80.8                                        & 50.6                                        & 35.9                                          & \multicolumn{1}{c|}{41.1}                                         & {\ul 74.6}                                   \\
\multicolumn{1}{c|}{SNELLA-16}         & 73.4                                        & \textbf{93.8}                                 & {\ul 72.5}                             & 99.3                                         & {\ul 91.2}                              & 55.5                                      & \multicolumn{1}{c|}{\textbf{91.7}}                           & {\ul 86.0}                                  & 96.0                                       & {\ul 87.0}                                  & \multicolumn{1}{c|}{\textbf{76.4}}                                  & 84.2                                           & 68.4                                          & \textbf{54.7}                            & {\ul 82.6}                                    & 80.8                                        & 49.9                                        & {\ul 36.8}                                    & \multicolumn{1}{c|}{{\ul 42.3}}                                   & \textbf{74.9}                                \\
\multicolumn{1}{c|}{SNELLA-32}         & 74.3                                        & {\ul 93.7}                                    & 72.3                                   & \textbf{99.4}                                & \textbf{92.0}                           & \textbf{55.9}                             & \multicolumn{1}{c|}{\textbf{91.7}}                           & 85.7                                        & \textbf{96.5}                              & \textbf{87.7}                               & \multicolumn{1}{c|}{76.1}                                           & 83.9                                           & 68.6                                          & 53.0                                     & \textbf{82.7}                                 & 81.9                                        & {\ul 51.3}                                  & \textbf{37.0}                                 & \multicolumn{1}{c|}{39.9}                                         & \textbf{74.9}                                \\ \bottomrule
\end{tabular}
}
\label{tab: vtab1k_performance_detail}
\vspace{-0.4cm}
\end{table*}

%% file: tab/tab_abl_adaptive.tex
\begin{table*}[]
\centering
\caption{We provide the performance on FGVC benchmark of different sparsification strategies~(Pre-defined and Adaptive) with two kernel functions~(P-Linear and Mix-K). A ViT-B/16 pre-trained on ImageNet-21k is selected as the base model.}
\begin{tabular}{@{}c|ccccc|c@{}}
\toprule
Method                 & CUB-200 & NABirds & Oxford Flowers & Stanford Dogs & Stanford Cars & Mean \\ \midrule
Pre-defined (P-Linear) & 88.0    & 82.1    & 99.0           & 89.4          & 88.4          & 89.4 \\
Pre-defined (Mix-K)    & 89.1    & 86.1    & 99.2           & 89.4          & 89.4          & 90.6 \\
Adaptive (Mix-K)       & 90.1    & 87.2    & 99.5           & 92.1          & 90.8          & 91.9 \\ \bottomrule
\end{tabular}
\label{tab: abl_adaptive}
\end{table*}

%% file: tab/tab_abl_llm.tex
\begin{table*}[t]
\caption{Performance on commonsense reasoning benchmark with LLaMA-2-7B and FGVC benchmark with ViT-B/16.}
\setlength\tabcolsep{2pt}
\begin{tabular}{@{}c|ccccccccc|cccccc@{}}
\toprule
\multirow{2}{*}{Model} & \multicolumn{9}{c|}{Commonsense Reasoning}                                          & \multicolumn{6}{c}{FGVC}                                                                                                                                                                                         \\ \cmidrule(l){2-16} 
                       & BoolQ & PIQA & SIQA & HellaSwag & WinoGrande & ARC-e & ARC-c & OBQA & Mean          & CUB-200 & NABirds & \begin{tabular}[c]{@{}c@{}}Oxford\\ Flowers\end{tabular} & \begin{tabular}[c]{@{}c@{}}Stanford\\ Dogs\end{tabular} & \begin{tabular}[c]{@{}c@{}}Stanford\\ Cars\end{tabular} & Mean          \\ \midrule
LoRA                   & 69.8  & 79.9 & 79.5 & 83.6      & 82.6       & 79.8  & 64.7  & 81.0 & 77.6          & 85.6    & 79.8    & 98.9                                                     & 87.6                                                    & 72.0                                                    & 84.8          \\
DoRA                   & 71.8  & 83.7 & 76.0 & 89.1      & 82.6       & 83.7  & 68.2  & 82.4 & \textbf{79.7} & 88.9    & 83.7    & 99.3                                                     & 90.4                                                    & 87.5                                                    & 90.0          \\
SNELLA                 & 71.5  & 83.0 & 81.0 & 82.0      & 80.7       & 82.7  & 68.1  & 81.0 & 78.8          & 90.1    & 87.2    & 99.5                                                     & 92.1                                                    & 90.8                                                    & \textbf{91.9} \\ \bottomrule
\end{tabular}
\label{tab: abl_llm}
\vspace{-0.2cm}
\end{table*}

%% file: main.bbl
\begin{thebibliography}{100}
\providecommand{\url}[1]{#1}
\csname url@samestyle\endcsname
\providecommand{\newblock}{\relax}
\providecommand{\bibinfo}[2]{#2}
\providecommand{\BIBentrySTDinterwordspacing}{\spaceskip=0pt\relax}
\providecommand{\BIBentryALTinterwordstretchfactor}{4}
\providecommand{\BIBentryALTinterwordspacing}{\spaceskip=\fontdimen2\font plus
\BIBentryALTinterwordstretchfactor\fontdimen3\font minus \fontdimen4\font\relax}
\providecommand{\BIBforeignlanguage}[2]{{%
\expandafter\ifx\csname l@#1\endcsname\relax
\typeout{** WARNING: IEEEtran.bst: No hyphenation pattern has been}%
\typeout{** loaded for the language `#1'. Using the pattern for}%
\typeout{** the default language instead.}%
\else
\language=\csname l@#1\endcsname
\fi
#2}}
\providecommand{\BIBdecl}{\relax}
\BIBdecl

\bibitem{chen2020simple}
T.~Chen, S.~Kornblith, M.~Norouzi, and G.~E. Hinton, ``A simple framework for contrastive learning of visual representations,'' in \emph{ICML}, 2020.

\bibitem{he2020momentum}
K.~He, H.~Fan, Y.~Wu, S.~Xie, and R.~B. Girshick, ``Momentum contrast for unsupervised visual representation learning,'' in \emph{CVPR}, 2020.

\bibitem{he2022masked}
K.~He, X.~Chen, S.~Xie, Y.~Li, P.~Doll{\'{a}}r, and R.~B. Girshick, ``Masked autoencoders are scalable vision learners,'' in \emph{CVPR}, 2022.

\bibitem{liang2025parameter}
D.~Liang, T.~Feng, X.~Zhou, Y.~Zhang, Z.~Zou, and X.~Bai, ``Parameter-efficient fine-tuning in spectral domain for point cloud learning,'' \emph{IEEE Transactions on Pattern Analysis and Machine Intelligence}, 2025.

\bibitem{kirillov2023segment}
A.~Kirillov, E.~Mintun, N.~Ravi, H.~Mao, C.~Rolland, L.~Gustafson, T.~Xiao, S.~Whitehead, A.~C. Berg, W.-Y. Lo \emph{et~al.}, ``Segment anything,'' in \emph{ICCV}, 2023.

\bibitem{esser2024scaling}
P.~Esser, S.~Kulal, A.~Blattmann, R.~Entezari, J.~M{\"u}ller, H.~Saini, Y.~Levi, D.~Lorenz, A.~Sauer, F.~Boesel \emph{et~al.}, ``Scaling rectified flow transformers for high-resolution image synthesis,'' in \emph{ICML}, 2024.

\bibitem{zhai2022scaling}
X.~Zhai, A.~Kolesnikov, N.~Houlsby, and L.~Beyer, ``Scaling vision transformers,'' in \emph{CVPR}, 2022.

\bibitem{bai2023sequential}
Y.~Bai, X.~Geng, K.~Mangalam, A.~Bar, A.~Yuille, T.~Darrell, J.~Malik, and A.~A. Efros, ``Sequential modeling enables scalable learning for large vision models,'' \emph{ArXiv preprint}, vol. abs/2312.00785, 2023.

\bibitem{dai2021coatnet}
Z.~Dai, H.~Liu, Q.~V. Le, and M.~Tan, ``Coatnet: Marrying convolution and attention for all data sizes,'' in \emph{NeurIPS}, 2021.

\bibitem{zhao2020masking}
M.~Zhao, T.~Lin, F.~Mi, M.~Jaggi, and H.~Sch{\"u}tze, ``Masking as an efficient alternative to finetuning for pretrained language models,'' in \emph{EMNLP}, 2020.

\bibitem{hu2021lora}
E.~J. Hu, Y.~Shen, P.~Wallis, Z.~Allen{-}Zhu, Y.~Li, S.~Wang, L.~Wang, and W.~Chen, ``Lora: Low-rank adaptation of large language models,'' in \emph{ICLR}, 2022.

\bibitem{zhang2024neural}
Y.~Zhang, K.~Zhou, and Z.~Liu, ``Neural prompt search,'' \emph{IEEE Transactions on Pattern Analysis and Machine Intelligence}, 2024.

\bibitem{jia2022visual}
M.~Jia, L.~Tang, B.-C. Chen, C.~Cardie, S.~Belongie, B.~Hariharan, and S.-N. Lim, ``Visual prompt tuning,'' in \emph{ECCV}, 2022.

\bibitem{chen2022adaptformer}
S.~Chen, C.~Ge, Z.~Tong, J.~Wang, Y.~Song, J.~Wang, and P.~Luo, ``Adaptformer: Adapting vision transformers for scalable visual recognition,'' \emph{NeurIPS}, 2022.

\bibitem{he2023sensitivity}
H.~He, J.~Cai, J.~Zhang, D.~Tao, and B.~Zhuang, ``Sensitivity-aware visual parameter-efficient fine-tuning,'' in \emph{ICCV}, 2023.

\bibitem{tu2023visual}
C.-H. Tu, Z.~Mai, and W.-L. Chao, ``Visual query tuning: Towards effective usage of intermediate representations for parameter and memory efficient transfer learning,'' in \emph{CVPR}, 2023.

\bibitem{zaken2021bitfit}
E.~Ben~Zaken, Y.~Goldberg, and S.~Ravfogel, ``{B}it{F}it: Simple parameter-efficient fine-tuning for transformer-based masked language-models,'' in \emph{Proceedings of the 60th Annual Meeting of the Association for Computational Linguistics (Volume 2: Short Papers)}, 2022.

\bibitem{caelles2017one}
S.~Caelles, K.~Maninis, J.~Pont{-}Tuset, L.~Leal{-}Taix{\'{e}}, D.~Cremers, and L.~V. Gool, ``One-shot video object segmentation,'' in \emph{CVPR}, 2017.

\bibitem{zhai2019large}
X.~Zhai, J.~Puigcerver, A.~Kolesnikov, P.~Ruyssen, C.~Riquelme, M.~Lucic, J.~Djolonga, A.~S. Pinto, M.~Neumann, A.~Dosovitskiy \emph{et~al.}, ``A large-scale study of representation learning with the visual task adaptation benchmark,'' \emph{ArXiv preprint}, vol. abs/1910.04867, 2019.

\bibitem{jha2020medico}
D.~Jha, S.~A. Hicks, K.~Emanuelsen, H.~Johansen, D.~Johansen, T.~de~Lange, M.~A. Riegler, and P.~Halvorsen, ``Medico multimedia task at mediaeval 2020: Automatic polyp segmentation,'' \emph{arXiv preprint arXiv:2012.15244}, 2020.

\bibitem{ruiz2023dreambooth}
N.~Ruiz, Y.~Li, V.~Jampani, Y.~Pritch, M.~Rubinstein, and K.~Aberman, ``Dreambooth: Fine tuning text-to-image diffusion models for subject-driven generation,'' in \emph{CVPR}, 2023.

\bibitem{fu2023effectiveness}
Z.~Fu, H.~Yang, A.~M.-C. So, W.~Lam, L.~Bing, and N.~Collier, ``On the effectiveness of parameter-efficient fine-tuning,'' in \emph{AAAI}, 2023.

\bibitem{lian2022scaling}
D.~Lian, D.~Zhou, J.~Feng, and X.~Wang, ``Scaling \& shifting your features: A new baseline for efficient model tuning,'' \emph{NeurIPS}, pp. 109--123, 2022.

\bibitem{SNELL}
S.~Shen, J.~Sun, X.~Ji, Q.~Huang, and S.~Wang, ``Expanding sparse tuning for low memory usage,'' in \emph{NeurIPS}, 2024.

\bibitem{gonen2011multiple}
M.~G{\"o}nen and E.~Alpayd{\i}n, ``Multiple kernel learning algorithms,'' \emph{The Journal of Machine Learning Research}, 2011.

\bibitem{zhang2023adalora}
Q.~Zhang, M.~Chen, A.~Bukharin, N.~Karampatziakis, P.~He, Y.~Cheng, W.~Chen, and T.~Zhao, ``Adalora: Adaptive budget allocation for parameter-efficient fine-tuning,'' \emph{arXiv preprint arXiv:2303.10512}, 2023.

\bibitem{louizos2017learning}
C.~Louizos, M.~Welling, and D.~P. Kingma, ``Learning sparse neural networks through l{\_}0 regularization,'' in \emph{ICLR}, 2018.

\bibitem{dosovitskiy2020image}
A.~Dosovitskiy, L.~Beyer, A.~Kolesnikov, D.~Weissenborn, X.~Zhai, T.~Unterthiner, M.~Dehghani, M.~Minderer, G.~Heigold, S.~Gelly, J.~Uszkoreit, and N.~Houlsby, ``An image is worth 16x16 words: Transformers for image recognition at scale,'' in \emph{ICLR}, 2021.

\bibitem{liu2021swin}
Z.~Liu, Y.~Lin, Y.~Cao, H.~Hu, Y.~Wei, Z.~Zhang, S.~Lin, and B.~Guo, ``Swin transformer: Hierarchical vision transformer using shifted windows,'' in \emph{ICCV}, 2021.

\bibitem{liu2022convnet}
Z.~Liu, H.~Mao, C.~Wu, C.~Feichtenhofer, T.~Darrell, and S.~Xie, ``A convnet for the 2020s,'' in \emph{CVPR}, 2022.

\bibitem{yao2025bi}
H.~Yao, R.~Zhang, H.~Lyu, Y.~Zhang, and C.~Xu, ``Bi-modality individual-aware prompt tuning for visual-language model,'' \emph{IEEE Transactions on Pattern Analysis and Machine Intelligence}, 2025.

\bibitem{qiao2025gradient}
J.~Qiao, Z.~Zhang, X.~Tan, Y.~Qu, W.~Zhang, Z.~Han, and Y.~Xie, ``Gradient projection for continual parameter-efficient tuning,'' \emph{IEEE Transactions on Pattern Analysis and Machine Intelligence}, 2025.

\bibitem{kim2024prompt}
M.~Kim, H.-I. Kim, and Y.~M. Ro, ``Prompt tuning of deep neural networks for speaker-adaptive visual speech recognition,'' \emph{IEEE Transactions on Pattern Analysis and Machine Intelligence}, 2024.

\bibitem{chen2025graph}
M.-S. Chen, P.-Y. Lai, D.-Z. Liao, C.-D. Wang, and J.-H. Lai, ``Graph prompt clustering,'' \emph{IEEE Transactions on Pattern Analysis and Machine Intelligence}, 2025.

\bibitem{bapna2019simple}
A.~Bapna and O.~Firat, ``Simple, scalable adaptation for neural machine translation,'' in \emph{EMNLP-IJCNLP}, 2019.

\bibitem{houlsby2019parameter}
N.~Houlsby, A.~Giurgiu, S.~Jastrzebski, B.~Morrone, Q.~de~Laroussilhe, A.~Gesmundo, M.~Attariyan, and S.~Gelly, ``Parameter-efficient transfer learning for {NLP},'' in \emph{ICML}, 2019.

\bibitem{pfeiffer2020adapterfusion}
J.~Pfeiffer, A.~Kamath, A.~R{\"u}ckl{\'e}, K.~Cho, and I.~Gurevych, ``{A}dapter{F}usion: Non-destructive task composition for transfer learning,'' in \emph{Proceedings of the 16th Conference of the European Chapter of the Association for Computational Linguistics: Main Volume}, 2021.

\bibitem{sung2022vl}
Y.-L. Sung, J.~Cho, and M.~Bansal, ``Vl-adapter: Parameter-efficient transfer learning for vision-and-language tasks,'' in \emph{CVPR}, 2022.

\bibitem{ding2021openprompt}
N.~Ding, S.~Hu, W.~Zhao, Y.~Chen, Z.~Liu, H.~Zheng, and M.~Sun, ``{O}pen{P}rompt: An open-source framework for prompt-learning,'' in \emph{Proceedings of the 60th Annual Meeting of the Association for Computational Linguistics: System Demonstrations}, 2022.

\bibitem{ju2022prompting}
C.~Ju, T.~Han, K.~Zheng, Y.~Zhang, and W.~Xie, ``Prompting visual-language models for efficient video understanding,'' in \emph{ECCV}, 2022.

\bibitem{liu2021p}
X.~Liu, K.~Ji, Y.~Fu, W.~L. Tam, Z.~Du, Z.~Yang, and J.~Tang, ``P-tuning v2: Prompt tuning can be comparable to fine-tuning universally across scales and tasks,'' \emph{ArXiv preprint}, vol. abs/2110.07602, 2021.

\bibitem{guo2020parameter}
D.~Guo, A.~Rush, and Y.~Kim, ``Parameter-efficient transfer learning with diff pruning,'' in \emph{Proceedings of the 59th Annual Meeting of the Association for Computational Linguistics and the 11th International Joint Conference on Natural Language Processing (Volume 1: Long Papers)}, 2021.

\bibitem{zhang2021tip}
R.~Zhang, R.~Fang, W.~Zhang, P.~Gao, K.~Li, J.~Dai, Y.~Qiao, and H.~Li, ``Tip-adapter: Training-free clip-adapter for better vision-language modeling,'' \emph{ArXiv preprint}, vol. abs/2111.03930, 2021.

\bibitem{li2021prefix}
X.~L. Li and P.~Liang, ``Prefix-tuning: Optimizing continuous prompts for generation,'' in \emph{Proceedings of the 59th Annual Meeting of the Association for Computational Linguistics and the 11th International Joint Conference on Natural Language Processing (Volume 1: Long Papers)}, 2021.

\bibitem{li2022cross}
W.~Li, X.~Liu, and H.~Bilen, ``Cross-domain few-shot learning with task-specific adapters,'' in \emph{CVPR}, 2022.

\bibitem{DBLP:conf/cvpr/ZhangZGZSZZ24}
Z.~Zhang, Q.~Zhang, Z.~Gao, R.~Zhang, E.~Shutova, S.~Zhou, and S.~Zhang, ``Gradient-based parameter selection for efficient fine-tuning,'' in \emph{CVPR}, 2024.

\bibitem{davenport2016overview}
M.~A. Davenport and J.~Romberg, ``An overview of low-rank matrix recovery from incomplete observations,'' \emph{IEEE Journal of Selected Topics in Signal Processing}, vol.~10, no.~4, pp. 608--622, 2016.

\bibitem{wang2024svd}
X.~Wang, Y.~Zheng, Z.~Wan, and M.~Zhang, ``Svd-llm: Truncation-aware singular value decomposition for large language model compression,'' \emph{arXiv preprint arXiv:2403.07378}, 2024.

\bibitem{zhao2024galore}
J.~Zhao, Z.~Zhang, B.~Chen, Z.~Wang, A.~Anandkumar, and Y.~Tian, ``Galore: Memory-efficient llm training by gradient low-rank projection,'' in \emph{ICML}.\hskip 1em plus 0.5em minus 0.4em\relax PMLR, 2024, pp. 61\,121--61\,143.

\bibitem{meng2024pissa}
F.~Meng, Z.~Wang, and M.~Zhang, ``Pissa: Principal singular values and singular vectors adaptation of large language models,'' \emph{NeurIPS}, 2024.

\bibitem{lialin2023relora}
V.~Lialin, N.~Shivagunde, S.~Muckatira, and A.~Rumshisky, ``Relora: High-rank training through low-rank updates,'' \emph{arXiv preprint arXiv:2307.05695}, 2023.

\bibitem{gonen2013kernelized}
M.~G{\"o}nen, S.~Khan, and S.~Kaski, ``Kernelized bayesian matrix factorization,'' in \emph{ICML}.\hskip 1em plus 0.5em minus 0.4em\relax PMLR, 2013, pp. 864--872.

\bibitem{rendle2008online}
S.~Rendle and L.~Schmidt-Thieme, ``Online-updating regularized kernel matrix factorization models for large-scale recommender systems,'' in \emph{Proceedings of the 2008 ACM conference on Recommender systems}, 2008, pp. 251--258.

\bibitem{gao2024parameter}
Z.~Gao, Q.~Wang, A.~Chen, Z.~Liu, B.~Wu, L.~Chen, and J.~Li, ``Parameter-efficient fine-tuning with discrete fourier transform,'' in \emph{ICML}.\hskip 1em plus 0.5em minus 0.4em\relax PMLR, 2024, pp. 14\,884--14\,901.

\bibitem{tian2024hydralora}
C.~Tian, Z.~Shi, Z.~Guo, L.~Li, and C.-Z. Xu, ``Hydralora: An asymmetric lora architecture for efficient fine-tuning,'' \emph{NeurIPS}, vol.~37, pp. 9565--9584, 2024.

\bibitem{pei2023dynamics}
Z.~Pei and S.~Wang, ``Dynamics-inspired neuromorphic visual representation learning,'' in \emph{ICML}, 2023.

\bibitem{wang2021learning}
S.~Wang, Z.~Chen, S.~Du, and Z.~Lin, ``Learning deep sparse regularizers with applications to multi-view clustering and semi-supervised classification,'' \emph{IEEE Transactions on Pattern Analysis and Machine Intelligence}, 2021.

\bibitem{peng2022exact}
J.~Peng, Y.~Wang, H.~Zhang, J.~Wang, and D.~Meng, ``Exact decomposition of joint low rankness and local smoothness plus sparse matrices,'' \emph{IEEE Transactions on Pattern Analysis and Machine Intelligence}, 2022.

\bibitem{liu2022learning}
Y.~Liu, J.~Cao, B.~Li, W.~Hu, and S.~Maybank, ``Learning to explore distillability and sparsability: a joint framework for model compression,'' \emph{IEEE Transactions on Pattern Analysis and Machine Intelligence}, 2022.

\bibitem{niu2021grim}
W.~Niu, Z.~Li, X.~Ma, P.~Dong, G.~Zhou, X.~Qian, X.~Lin, Y.~Wang, and B.~Ren, ``Grim: A general, real-time deep learning inference framework for mobile devices based on fine-grained structured weight sparsity,'' \emph{IEEE Transactions on Pattern Analysis and Machine Intelligence}, 2021.

\bibitem{zhang2023lottery}
Y.~Zhang, M.~Lin, Y.~Zhong, F.~Chao, and R.~Ji, ``Lottery jackpots exist in pre-trained models,'' \emph{IEEE Transactions on Pattern Analysis and Machine Intelligence}, 2023.

\bibitem{shen2025enhancing}
S.~Shen, Z.~Qi, J.~Sun, Q.~Huang, Q.~Tian, and S.~Wang, ``Enhancing pre-trained representation classifiability can boost its interpretability,'' in \emph{ICLR}, 2025.

\bibitem{shen2025vlsae}
S.~Shen, J.~Sun, Q.~Huang, and S.~Wang, ``Vl-sae: Interpreting and enhancing vision-language alignment with a unified concept set,'' 2025.

\bibitem{liu2025edit}
J.~Liu, J.~Sun, S.~Shen, C.~Yang, and S.~Wang, ``Edit less, achieve more: Dynamic sparse neuron masking for lifelong knowledge editing in llms,'' 2025.

\bibitem{han2016eie}
S.~Han, X.~Liu, H.~Mao, J.~Pu, A.~Pedram, M.~A. Horowitz, and W.~J. Dally, ``Eie: Efficient inference engine on compressed deep neural network,'' \emph{ACM SIGARCH Computer Architecture News}, 2016.

\bibitem{molchanov2017variational}
D.~Molchanov, A.~Ashukha, and D.~P. Vetrov, ``Variational dropout sparsifies deep neural networks,'' in \emph{ICML}, 2017.

\bibitem{liu2021discrimination}
J.~Liu, B.~Zhuang, Z.~Zhuang, Y.~Guo, J.~Huang, J.~Zhu, and M.~Tan, ``Discrimination-aware network pruning for deep model compression,'' \emph{IEEE Transactions on Pattern Analysis and Machine Intelligence}, 2021.

\bibitem{dong2024prompt}
H.~Dong, B.~Chen, and Y.~Chi, ``Prompt-prompted adaptive structured pruning for efficient llm generation,'' \emph{arXiv preprint arXiv:2404.01365}, 2024.

\bibitem{an2024fluctuation}
Y.~An, X.~Zhao, T.~Yu, M.~Tang, and J.~Wang, ``Fluctuation-based adaptive structured pruning for large language models,'' in \emph{Proceedings of the AAAI Conference on Artificial Intelligence}, 2024.

\bibitem{wang2024rl}
B.~Wang and V.~Kindratenko, ``Rl-pruner: Structured pruning using reinforcement learning for cnn compression and acceleration,'' \emph{arXiv preprint arXiv:2411.06463}, 2024.

\bibitem{hu2016network}
H.~Hu, R.~Peng, Y.-W. Tai, and C.-K. Tang, ``Network trimming: A data-driven neuron pruning approach towards efficient deep architectures,'' \emph{ArXiv preprint}, vol. abs/1607.03250, 2016.

\bibitem{srinivas2015data}
S.~Srinivas and R.~V. Babu, ``Data-free parameter pruning for deep neural networks,'' in \emph{BMVC}, 2015.

\bibitem{dong2017learning}
X.~Dong, S.~Chen, and S.~J. Pan, ``Learning to prune deep neural networks via layer-wise optimal brain surgeon,'' in \emph{NeurIPS}, 2017.

\bibitem{yang2017designing}
T.~Yang, Y.~Chen, and V.~Sze, ``Designing energy-efficient convolutional neural networks using energy-aware pruning,'' in \emph{CVPR}, 2017.

\bibitem{bellec2017deep}
G.~Bellec, D.~Kappel, W.~Maass, and R.~A. Legenstein, ``Deep rewiring: Training very sparse deep networks,'' in \emph{ICLR}, 2018.

\bibitem{frankle2018lottery}
J.~Frankle and M.~Carbin, ``The lottery ticket hypothesis: Finding sparse, trainable neural networks,'' in \emph{ICLR}.\hskip 1em plus 0.5em minus 0.4em\relax OpenReview.net, 2019.

\bibitem{koutroumbas2008pattern}
K.~Koutroumbas and S.~Theodoridis, \emph{Pattern recognition}.\hskip 1em plus 0.5em minus 0.4em\relax Academic Press, 2008.

\bibitem{suykens1999chaos}
J.~A. Suykens and J.~Vandewalle, ``Chaos control using least-squares support vector machines,'' \emph{International journal of circuit theory and applications}, 1999.

\bibitem{berlinet2011reproducing}
A.~Berlinet and C.~Thomas-Agnan, \emph{Reproducing kernel Hilbert spaces in probability and statistics}, 2011.

\bibitem{zhang2023mosa}
Q.~Zhang, B.~Zou, R.~An, J.~Liu, and S.~Zhang, ``Mosa: Mixture of sparse adapters for visual efficient tuning,'' \emph{ArXiv preprint}, vol. abs/2312.02923, 2023.

\bibitem{wah2011caltech}
C.~Wah, S.~Branson, P.~Welinder, P.~Perona, and S.~Belongie, ``The caltech-ucsd birds-200-2011 dataset,'' 2011.

\bibitem{van2015building}
G.~V. Horn, S.~Branson, R.~Farrell, S.~Haber, J.~Barry, P.~Ipeirotis, P.~Perona, and S.~J. Belongie, ``Building a bird recognition app and large scale dataset with citizen scientists: The fine print in fine-grained dataset collection,'' in \emph{CVPR}, 2015.

\bibitem{nilsback2008automated}
M.-E. Nilsback and A.~Zisserman, ``Automated flower classification over a large number of classes,'' in \emph{2008 Sixth Indian conference on computer vision, graphics \& image processing}.\hskip 1em plus 0.5em minus 0.4em\relax IEEE, 2008.

\bibitem{gebru2017fine}
T.~Gebru, J.~Krause, Y.~Wang, D.~Chen, J.~Deng, and L.~Fei{-}Fei, ``Fine-grained car detection for visual census estimation,'' in \emph{AAAI}, 2017.

\bibitem{dataset2011novel}
E.~Dataset, ``Novel datasets for fine-grained image categorization,'' in \emph{First Workshop on Fine Grained Visual Categorization, CVPR. Citeseer. Citeseer}, 2011.

\bibitem{ridnik2021imagenet}
T.~Ridnik, E.~Ben-Baruch, A.~Noy, and L.~Zelnik-Manor, ``Imagenet-21k pretraining for the masses,'' \emph{arXiv preprint arXiv:2104.10972}, 2021.

\bibitem{chen2021empirical}
X.~Chen, S.~Xie, and K.~He, ``An empirical study of training self-supervised vision transformers,'' in \emph{ICCV}, 2021.

\bibitem{russakovsky2015imagenet}
O.~Russakovsky, J.~Deng, H.~Su, J.~Krause, S.~Satheesh, S.~Ma, Z.~Huang, A.~Karpathy, A.~Khosla, M.~Bernstein \emph{et~al.}, ``Imagenet large scale visual recognition challenge,'' \emph{International journal of computer vision}, 2015.

\bibitem{chen2023sam}
T.~Chen, L.~Zhu, C.~Ding, R.~Cao, Y.~Wang, Z.~Li, L.~Sun, P.~Mao, and Y.~Zang, ``Sam fails to segment anything?--sam-adapter: Adapting sam in underperformed scenes: Camouflage, shadow, medical image segmentation, and more,'' \emph{arXiv preprint arXiv:2304.09148}, 2023.

\bibitem{radford2021learning}
A.~Radford, J.~W. Kim, C.~Hallacy, A.~Ramesh, G.~Goh, S.~Agarwal, G.~Sastry, A.~Askell, P.~Mishkin, J.~Clark \emph{et~al.}, ``Learning transferable visual models from natural language supervision,'' in \emph{ICML}, 2021.

\bibitem{friedman2022vendi}
D.~Friedman and A.~B. Dieng, ``The vendi score: A diversity evaluation metric for machine learning,'' \emph{arXiv preprint arXiv:2210.02410}, 2022.

\bibitem{oquab2023dinov2}
M.~Oquab, T.~Darcet, T.~Moutakanni, H.~Vo, M.~Szafraniec, V.~Khalidov, P.~Fernandez, D.~Haziza, F.~Massa, A.~El-Nouby \emph{et~al.}, ``Dinov2: Learning robust visual features without supervision,'' \emph{arXiv preprint arXiv:2304.07193}, 2023.

\bibitem{ding2025stable}
G.~Ding, X.~Han, S.~Wang, X.~Jin, and Q.~Huang, ``Stable attribute group editing for reliable few-shot image generation,'' \emph{IEEE Transactions on Circuits and Systems for Video Technology}, 2025.

\bibitem{ding2025dis2booth}
G.~Ding, C.~Yang, S.~Wang, X.~Li, J.~Zhang, X.~Jin, and Q.~Huang, ``Dis$^2$booth: Learning image distribution with disentangled features for text-to-image diffusion models,'' in \emph{Proceedings of the AAAI Conference on Artificial Intelligence}, vol.~39, no.~3, 2025, pp. 2744--2752.

\bibitem{estevez2009normalized}
P.~A. Est{\'e}vez, M.~Tesmer, C.~A. Perez, and J.~M. Zurada, ``Normalized mutual information feature selection,'' \emph{IEEE Transactions on neural networks}, 2009.

\bibitem{NEURIPS2023_6dcf277e}
H.~Liu, C.~Li, Q.~Wu, and Y.~J. Lee, ``Visual instruction tuning,'' in \emph{NeurIPS}, vol.~36, 2023.

\bibitem{li2024llavanextinterleavetacklingmultiimagevideo}
F.~Li, R.~Zhang, H.~Zhang, Y.~Zhang, B.~Li, W.~Li, Z.~Ma, and C.~Li, ``Llava-next-interleave: Tackling multi-image, video, and 3d in large multimodal models,'' 2024.

\bibitem{liu2024dora}
S.-Y. Liu, C.-Y. Wang, H.~Yin, P.~Molchanov, Y.-C.~F. Wang, K.-T. Cheng, and M.-H. Chen, ``Dora: Weight-decomposed low-rank adaptation,'' \emph{ArXiv preprint}, vol. abs/2402.09353, 2024.

\bibitem{von-platen-etal-2022-diffusers}
P.~von Platen, S.~Patil, A.~Lozhkov, P.~Cuenca, N.~Lambert, K.~Rasul, M.~Davaadorj, D.~Nair, S.~Paul, W.~Berman, Y.~Xu, S.~Liu, and T.~Wolf, ``Diffusers: State-of-the-art diffusion models,'' 2022.

\bibitem{accelerate}
S.~Gugger, L.~Debut, T.~Wolf, P.~Schmid, Z.~Mueller, S.~Mangrulkar, M.~Sun, and B.~Bossan, ``Accelerate: Training and inference at scale made simple, efficient and adaptable.'' 2022.

\bibitem{touvron2023llama}
H.~Touvron, L.~Martin, K.~Stone, P.~Albert, A.~Almahairi, Y.~Babaei, N.~Bashlykov, S.~Batra, P.~Bhargava, S.~Bhosale \emph{et~al.}, ``Llama 2: Open foundation and fine-tuned chat models,'' \emph{ArXiv preprint}, vol. abs/2307.09288, 2023.

\end{thebibliography}
